\documentclass{article}

\usepackage[numbers]{natbib}

\usepackage[final]{neurips_2022}

\usepackage[utf8]{inputenc} 
\usepackage[T1]{fontenc}    
\usepackage[colorlinks=true,bookmarks=false,linkcolor=NavyBlue,urlcolor=NavyBlue,citecolor=NavyBlue,breaklinks=true]{hyperref}  
\usepackage{url}            
\usepackage{booktabs}       
\usepackage{nicefrac}       
\usepackage{microtype}      
\usepackage[dvipsnames]{xcolor}         
\usepackage{graphics}
\usepackage{amsmath}
\usepackage{mathtools}
\usepackage{amssymb}
\usepackage{wrapfig}
\usepackage{lipsum}
\usepackage{floatrow}
\usepackage{amsmath,amsthm,bm}
\usepackage{algorithm}
\usepackage{algorithmicx}

\usepackage[footnotesize]{caption}
\usepackage[algoruled,boxed,lined,noend]{algorithm2e}
\usepackage{subcaption}
\usepackage{mathptmx}
\usepackage{thmtools,thm-restate}
\usepackage{wasysym}

\let\mathcal\undefined
\DeclareMathAlphabet{\mathcal}{OMS}{cmsy}{m}{n}

\newtheorem{theorem}{Theorem}[section]
\newtheorem{lemma}[theorem]{Lemma}
\newtheorem{corollary}[theorem]{Corollary}

\newtheorem{assumption}{Assumption}
\newtheorem{definition}{Definition}

\DeclareMathOperator*{\argmax}{arg\,max}

\DeclareCaptionLabelFormat{andtable}{#1~#2  \&  \tablename~\thetable}
\title{Unpacking Reward Shaping: Understanding the Benefits of Reward Engineering on Sample Complexity}

\author{%
    Abhishek Gupta\thanks{Equal Contributions} \\
    University of Washington\\
    \texttt{abhgupta@cs.washington.edu} \\
    \And
    Aldo Pacchiano$^*$ \\
    Microsoft Research, NYC\\
    \texttt{apacchiano@microsoft.com} \\
    \AND
    Yuexiang Zhai \\
    UC Berkeley, EECS\\
    \texttt{simonzhai@berkeley.edu} \\
    \And
    Sham M. Kakade \\
    Harvard University \\
    \texttt{sham@seas.harvard.edu} \\
    \And
    Sergey Levine \\
    UC Berkeley, EECS \\
    \texttt{svlevine@eecs.berkeley.edu} \\
}

\begin{document}

\newcommand{\cmt}[1]{{\footnotesize\textcolor{red}{#1}}}
\newcommand{\todo}[1]{\cmt{(TODO: #1)}}
\newcommand{\kelvin}[1]{{\footnotesize\textcolor{red}{Kelvin: #1}}}
\newcommand{\boldme}[1]{{\footnotesize\textbf{#1}}}
\newcommand{\ourmethod}{\boldme{METHOD\_NAME}}

\newcommand{\E}[2]{\operatorname{\mathbb{E}}_{#1}\left[#2\right]}
\newcommand{\expectation}{\mathbb{E}}
\newcommand{\density}{p}
\newcommand{\proposal}{q}  
\newcommand{\target}{p}  
\newcommand{\prop}{P}
\newcommand{\kl}[2]{\mathrm{D_{KL}}\left(#1\;\middle\|\;#2\right)}
\newcommand{\fdiv}[2]{D_{f}\left(#1\;\middle\|\;#2\right)}
\newcommand{\entropy}{\mathcal{H}}
\newcommand{\information}{\mathcal{I}}
\newcommand{\ent}{\mathcal{H}}
\newcommand{\sdots}{\,\cdot\,}
\newcommand{\func}{\mathbf{f}}
\newcommand{\methodname}{RLDE }
\newcommand{\piexplore}{$\pi_{\text{explore}}$}
\newcommand{\piexploit}{$\pi_{\text{exploit}}$}
\newcommand{\rhoexploit}{$\rho_{\text{exploit}}(s,a)$ }
\newcommand{\rhoexplore}{$\rho_{\text{explore}}(s,a)$ }

\newcommand{\Vstarmax}{\Vstar_{\mathrm{max}}}


\newcommand{\ratio}{r(x)}
\newcommand{\classifier}{D(x)}
\newcommand{\dataset}{\mathcal{X}}

\newcommand{\context}{\mathbf{z}}

\newcommand{\objective}{\mathcal{J}}

\newcommand{\info}{\mathcal{I}}

\newcommand{\ones}{\boldsymbol{1}}
\newcommand{\eye}{\boldsymbol{I}}
\newcommand{\zeros}{\boldsymbol{0}}

\newcommand{\task}{\mathcal{T}}
\newcommand{\metatrain}{\mathcal{T}^\text{meta-train}}
\newcommand{\metatest}{\mathcal{T}^\text{meta-test}}

\newcommand{\metatrainset}{\{\mathcal{T}_i~;~ i=1..N\}}
\newcommand{\metatestset}{\{\mathcal{T}_j~;~ j=1..M\} }

\newcommand{\vphi}{\boldsymbol{\phi}}
\newcommand{\vtheta}{\boldsymbol{\theta}}
\newcommand{\vmu}{\boldsymbol{\mu}}

\newcommand{\vx}{\mathbf{x}}
\newcommand{\vy}{\mathbf{y}}
\newcommand{\discrim}{q_{\omega}}

\newcommand{\Scal}{\mathcal{S}}
\newcommand{\Acal}{\mathcal{A}}
\newcommand{\Ocal}{\mathcal{O}}
\newcommand{\Mcal}{\mathcal{M}}
\newcommand{\Tcal}{\mathcal{T}}
\newcommand{\Vtilde}{\widetilde{V}}
\newcommand{\Vhat}{\widehat{V}}
\newcommand{\Vstar}{V^\star}
\newcommand{\Phat}{\widehat{\mathbb{P}}}
\newcommand{\Pstar}{\mathbb{P}^\star}
\newcommand{\Vcal}{\mathcal{V}}
\newcommand{\Qhat}{\widehat{Q}}
\newcommand{\Qstar}{Q^\star}
\newcommand{\Qtilde}{\widetilde{Q}}
\newcommand{\pseudosub}{\mathrm{PseudoSub}}
\newcommand{\boundarypseudosub}{\mathrm{BoundaryPseudoSub}}

\newcommand{\pathpseudosub}{\mathrm{PathPseudoSub}}

\newcommand{\neighbor}{\mathrm{Neighbor}}
\newcommand{\Vtildemax}{\Vtilde^{\mathrm{max}}}
\newcommand{\Qtildemax}{\Qtilde^{\mathrm{max}}}

\newcommand{\sspace}{\mathcal{S}}
\newcommand{\aspace}{\mathcal{A}}
\newcommand{\state}{s}
\newcommand{\sz}{{\state_0}}
\newcommand{\stm}{{\state_{t-1}}}
\newcommand{\st}{{\state_t}}
\newcommand{\sti}{{\state_t^{(i)}}}
\newcommand{\sT}{{\state_T}}
\newcommand{\stp}{{\state_{t+1}}}
\newcommand{\stpi}{{\state_{t+1}^{(i)}}}
\newcommand{\pdyn}{\density_\state}
\newcommand{\learnedmodel}{\hat{\density}_\state}
\newcommand{\pz}{\density_0}
\newcommand{\horizon}{H_T}
\newcommand{\evalhorizon}{H_E}
\newcommand{\action}{a}
\newcommand{\az}{{\action_0}}
\newcommand{\atm}{{\action_{t-1}}}
\newcommand{\at}{{\action_t}}
\newcommand{\ati}{{\action_t^{(i)}}}
\newcommand{\atj}{{\action_t^{(j)}}}
\newcommand{\attildej}{{\tilde{\action}_t^{(j)}}}
\newcommand{\atij}{{\action_t^{(i,j)}}}
\newcommand{\atp}{{\action_{t+1}}}
\newcommand{\aT}{{\action_T}}
\newcommand{\atk}{{\action_t^{(k)}}}
\newcommand{\aTm}{\action_{\horizon-1}}
\newcommand{\opt}{^*}
\newcommand{\mdps}{\mathcal{M}}


\newcommand{\explorei}{\pi^{\text{explore}}_{i}}

\newcommand{\explorepolicy}{\pi^{\text{explore}}_{\phi}}
\newcommand{\exploitpolicy}{\pi^{\text{exploit}}_{\theta}}

\newcommand{\demos}{\mathcal{D}}
\newcommand{\traj}{\tau}
\newcommand{\ptraj}{\density_\traj}
\newcommand{\visits}{\rho}  

\newcommand{\reward}{r}
\newcommand{\rz}{\reward_0}
\newcommand{\rt}{\reward_t}
\newcommand{\rti}{\reward^{(i)}_t}
\newcommand{\rmi}{r_\mathrm{min}}
\newcommand{\rmax}{r_\mathrm{max}}

\newcommand{\return}{J}

%

\newcommand{\loss}{\mathcal{L}}

\newcommand{\policyopt}{\pi^*}

\newcommand{\V}{V}
\newcommand{\Vsoft}{V_\mathrm{soft}}
\newcommand{\Vsoftparams}{V_\mathrm{soft}^\qparams}
\newcommand{\Vhatsoftparams}{\hat V_\mathrm{soft}^\qparams}
\newcommand{\Vhatsoft}{\hat V_\mathrm{soft}}
\newcommand{\Vhard}{V^{\dagger}}
\newcommand{\Q}{Q}
\newcommand{\Qsoft}{Q_\mathrm{soft}}
\newcommand{\Qsoftparams}{Q_\mathrm{soft}^\qparams}
\newcommand{\Qhatsoft}{\hat Q_\mathrm{soft}}
\newcommand{\Qhatsoftparams}{\hat Q_\mathrm{soft}^{\bar\qparams}}
\newcommand{\Qhard}{Q^{\dagger}}
\newcommand{\A}{A}
\newcommand{\Asoft}{A_\mathrm{soft}}
\newcommand{\Asoftparams}{A_\mathrm{soft}^\qparams}
\newcommand{\Ahatsoft}{\hat A_\mathrm{soft}}

\newcommand{\energy}{\mathcal{E}}

\newcommand{\tasklosstrain}{\loss_{\task}^{\text{ tr}}}
\newcommand{\tasklosstest}{\loss_{\task}^{\text{ test}}}

\newcommand{\tasklosstraini}{\loss_{\task_i}^{\text{ tr}}}
\newcommand{\tasklosstesti}{\loss_{\task_i}^{\text{ test}}}


\newcommand{\indep}{\rotatebox[origin=c]{90}{$\models$}}

\renewcommand{\mathbf}{\boldsymbol}

\newcommand{\conv}{\circledast}
\newcommand{\mb}{\mathbf}
\newcommand{\mc}{\mathcal}
\newcommand{\mf}{\mathfrak}
\newcommand{\md}{\mathds}
\newcommand{\bb}{\mathbb}
\newcommand{\msf}{\mathsf}
\newcommand{\magnitude}[1]{ \left| #1 \right| }
\newcommand{\set}[1]{\left\{ #1 \right\}}
\newcommand{\condset}[2]{ \left\{ #1 \;\middle|\; #2 \right\} }

\newcommand{\reals}{\bb R}
\newcommand{\proj}{\mathrm{proj}}

\newcommand{\eps}{\varepsilon}
\newcommand{\R}{\reals}
\newcommand{\Cp}{\bb C}
\newcommand{\Z}{\bb Z}
\newcommand{\N}{\bb N}
\newcommand{\Sp}{\bb S}
\newcommand{\Ba}{\bb B}
\newcommand{\indicator}[1]{\mathbf{1}\left\{#1\right\}}
\renewcommand{\P}{\mathbb{P}}
\newcommand{\rvline}{\hspace*{-\arraycolsep}\vline\hspace*{-\arraycolsep}}


\newcommand{\event}{\mc E}

\newcommand{\e}{\mathrm{e}}
\newcommand{\im}{\mathrm{i}}
\newcommand{\rconcave}{r_\fgecap}
\newcommand{\Lconcave}{\mc L^\fgecap}
\newcommand{\rconvex}{r_\fgecup}
\newcommand{\Rconvex}{R_\fgecup}
\newcommand{\Lconvex}{\mc L^\fgecup}

\newcommand{\wh}{\widehat}
\newcommand{\wt}{\widetilde}
\newcommand{\ol}{\overline}

\newcommand{\betaconcave}{\beta_\fgecap}
\newcommand{\betagrad}{\beta_{\mathrm{grad}}}

\newcommand{\norm}[2]{\left\| #1 \right\|_{#2}}
\newcommand{\abs}[1]{\left| #1 \right|}
\newcommand{\row}[1]{\text{row}\left( #1 \right)}
\newcommand{\innerprod}[2]{\left\langle #1,  #2 \right\rangle}
\newcommand{\prob}[1]{\bb P\left[ #1 \right]}
\newcommand{\expect}[1]{\bb E\left[ #1 \right]}
\newcommand{\integral}[4]{\int_{#1}^{#2}\; #3\; #4}
\newcommand{\paren}[1]{\left( #1 \right)}
\newcommand{\brac}[1]{\left[ #1 \right]}
\newcommand{\Brac}[1]{\left\{ #1 \right\}}

\newcommand{\here}{{\color{purple}{\bf [Writing here]}}}
\newcommand{\highlight}[1]{{\color{red}{#1}}}
\newcommand{\simon}[1]{{\color{blue}{\bf Simon: #1}}}
\newcommand{\aldo}[1]{{\color{red}{\bf Aldo: #1}}}
\newcommand{\mr}{\mathrm}
\newcommand{\sym}{\mathrm{Sym}}
\newcommand{\sks}{\mathrm{Skew}}
\newcommand{\inprod}[2]{\langle#1,#2\rangle}
\newcommand{\parans}[1]{\left(#1\right}

\maketitle

\begin{abstract}
Reinforcement learning provides an automated framework for learning behaviors from high-level reward specifications, but in practice the choice of reward function can be crucial for good results -- while in principle the reward only needs to specify what the task is, in reality practitioners often need to design more detailed rewards that provide the agent with some hints about how the task should be completed. The idea of this type of ``reward-shaping'' has been often discussed in the literature, and is often a critical part of practical applications, but there is relatively little formal characterization of how the choice of reward shaping can yield benefits in sample complexity. In this work, we build on the framework of novelty-based exploration to provide a simple scheme for incorporating shaped rewards into RL along with an analysis tool to show that particular choices of reward shaping provably improve sample efficiency. We characterize the class of problems where these gains are expected to be significant and show how this can be connected to practical algorithms in the literature. We confirm that these results hold in practice in an experimental evaluation, providing an insight into the mechanisms through which reward shaping can significantly improve the complexity of reinforcement learning while retaining asymptotic performance. 
\end{abstract}

\section{Introduction}
\label{sec:intro}
\vspace{-0.3cm}
Reinforcement learning (RL) in its most general form presents a very difficult optimization problem: when there are no constraints on the reward function or dynamics, a learning algorithm may need to exhaustively explore the entire state space to discover high-reward regions. Na\"{i}ve algorithms that rely entirely on random exploration are known to be exponentially expensive~\citep{kakade2003sample,agarwal2020pc}, and much of the theoretical work on efficient RL algorithms has focused on smarter exploration strategies that aim to more efficiently cover state space, typically in time polynomial in the state space cardinality~\citep{auer2008near,azar2017minimax,jin2018q,zhang2020almost,li2021breaking,pacchiano2020optimism,zanette2019tighter}. Much of this work is based on upper confidence bound (UCB) principles and prescribes some kind of exploration bonus to prioritize exploration of rarely visited regions. Analogous strategies have also been employed in a number of practical RL algorithms~\citep{tang2017exploration,stadie2015incentivizing,bellemare2016unifying,burda2018exploration,houthooft2016vime,pathak2017curiosity,vezhnevets2017feudal,VanSeijen2017Hybrid,Raileanu2020RIDE:}. However, perhaps surprisingly, much of the empirical work on reinforcement learning does not make use of explicit exploration bonuses or other dedicated exploration strategies, despite numerous theoretical results showing them to be essential to attain tractable sample complexity. Instead, practitioners often incorporate prior knowledge of each task into designing or shaping the reward function~\citep{ng1999policy,silver2016mastering,silver2017mastering,vinyals2019grandmaster,andrychowicz2020learning,berner2019dota,ye2020towards,zhai2022computational}, preferring this heuristic approach over the more principled exploration strategies. At this point, one may wonder why is reward shaping often practically preferable to dedicated exploration?

A likely answer to this question lies in the fact that even the best general-purpose exploration strategies still require visiting every state in the MDP at least once in the worst case. This is of course unavoidable without further assumptions. However, in practice, sample complexity that is polynomial in the size of the entire state space might \emph{still} not be practical, and hence prior knowledge in the form of reward shaping is required to render such tasks tractable. Surprisingly, despite the widespread popularity of reward shaping in RL applications, the analysis of reward shaping has remained limited to proving policy invariance~\citep{ng1999policy} or largely empirical observations, often relegating reward shaping to folk knowledge. In this work, we take a step towards studying the effect of reward shaping on the efficiency of RL algorithms, by asking the following question:\vspace{-0.2cm}
\begin{center}
    \emph{Can we theoretically justify the sample complexity benefits that reward shaping from prior domain knowledge can provide for reinforcement learning?}\vspace{-0.2cm}
\end{center}
We aim to provide a set of tools that formally analyze how reward shaping can improve the complexity of tabula-rasa RL and better direct exploration. To perform this analysis, we first propose a simple modification to standard RL algorithms \textemdash ``UCBVI-Shaped'' that incorporates shaped rewards into optimism based exploration. We use this algorithm instantiation to then provide a regret analysis framework that studies how this introduction of shaped rewards can (in certain cases) provide much more directed optimism than uninformed exploration, while maintaining asymptotic performance. 

To approach our analysis, we specifically consider problems where the reward shaping is provided through a term $\Vtilde$, a (potentially suboptimal) approximation of the optimal value function $\Vstar$. In particular, we assume that the shaping $\Vtilde$ is a multiplicatively bounded approximation of the optimal value function $\Vstar$, i.e., $ \Vtilde(s) \leq \Vstar(s) \leq \beta \Vtilde(s), \forall s$, for a finite multiplicative factor $\beta$. This type of shaped reward function $\Vtilde$ can be incorporated into a standard RL algorithm like UCBVI~\citep{azar2017minimax} through two channels: (1) bonus scaling -- simply reweighting a standard, decaying count-based bonus $\frac{1}{\sqrt{N_h(s, a)}}$ by the per-state reward shaping and (2) value projection -- adaptively projecting learned value functions into ranges of value functions derived from the reward shaping. 


We show that this relatively simple algorithmic instantiation lends itself to an analysis that shows significant sample complexity benefits with shaping. Intuitively, the key pieces in our complexity analysis of UCBVI-Shaped are: (1) A multiplicative sandwich condition (via $\beta$) between $\Vtilde$ and $\Vstar$ allows for the bonus scaling to depend on $\beta \Vtilde$ instead of a coarse approximation of $\Vstar$ such as $H$. This allows for a reduction of complexity from a horizon $H$ dependent term to one scaling with $\beta\Vtilde \leq \beta\Vstar$ by allowing for a faster decay of the exploration bonuses while still providing enough optimism. (2) A projection of the value function prevents ``over optimism'' by hastening the convergence of the empirical $\Qhat$ functions during value iteration, thus allowing for faster detection of sub-optimal actions. This results in the ability to prune out large parts of the state space, as we also validate empirically. 

To summarize, the key contribution of this work is to characterize how reward shaping can provably improve sample efficiency by providing gains in both $|\Scal|$ and $H$ dependent terms. 
We do so by analyzing the gains from reward shaping in two different ways: bonus scaling and value projection. We show that the ``quality'' (determined by $\beta$) of the provided shaping can significantly improve the sample efficiency of the resulting reinforcement learning algorithm, provide a set of analysis techniques to understand improvements in sample complexity from shaping, and confirm our findings with numerical experiments.
\vspace{-0.2cm}

\section{Related Work}
\vspace{-0.3cm}

\paragraph{Regret Analysis in Finite Horizon Episodic Tabular MDPs.}
Recent research on regret analysis has studied both model-based and model-free RL methods. Model-based methods~\citep{jaksch2010near,agrawal2017posterior,azar2017minimax,fruit2018efficient,efroni2019tight,pacchiano2021towards} first learn a model from previous experience and use the learned model for future decision making. In contrast, model-free methods~\citep{jin2018q,bai2019provably,zhang2020almost,yang2021q,menard2021ucb,li2021breaking} aim to learn the value function without the model estimation stage and use the learned value function for decision making. Our analysis lies in the model-based framework and is similar to the setting of~\citep{azar2017minimax} but with additional assumptions on knowing $\Vtilde$ as a multiplicatively bounded approximation of $\Vstar$. The main difference between our results and the aforementioned works is that we consider the reward shaping setting with a truncated interval assumption. As a result, we reduce the state dependency $|\Scal|$ to the some smaller ``effective state space''. Our work is also closely related to~\citep{golowich2022can}. We discuss this in Section~\ref{sec:preliminary}. \vspace{-0.3cm}

\paragraph{Regret Analysis with Linear Function Approximation.}
A recent line of literature has investigated the linear function approximation setting by assuming the transition kernels or the value functions can be represented by $d$-dimensional linear features~\citep{jin2020provably,zanette2020frequentist,ayoub2020model,zhou2021provably,cai2020provably,yang2020reinforcement,yang2019sample,wang2019optimism} or even general function classes~\citep{yang2020function}. With the aforementioned assumptions, the regret analysis will only depend on the ambient dimension $d$ (or other intrinsic complexity measure), instead of $S,A$ in the tabular setting~\citep{wang2019optimism,yang2020reinforcement,cai2020provably,zhou2021provably,ayoub2020model,zanette2019tighter,jin2020provably}, which could greatly decrease the complexity of learning. Instead of posing structural assumptions on the function class representing the MDP's values or dynamics, we ask the question of whether having approximate knowledge of $\Vstar$ can improve the speed of learning. \vspace{-0.3cm}


\paragraph{Practical Reward Design and Reward Shaping.}
\citet{ng1999policy} proposed a potential-based shaping function $F$ that ensures policy invariance under transformation. However, unbiased potential based reward shaping is rarely used in practice. In most applications, heuristic reward design is carefully performed with  potentially biased reward functions. In some large-scale practical RL tasks~\citep{berner2019dota,ye2020towards, yu2019metaworld}, the reward functions are heavily handcrafted based on prior knowledge. Besides reward engineering, a distinct line of work applies uninformed exploration algorithms like count-based reward shaping or intrinsic rewards to encourage greater state visitation~\citep{tang2017exploration,stadie2015incentivizing,bellemare2016unifying,burda2018exploration,houthooft2016vime,pathak2017curiosity,vezhnevets2017feudal,VanSeijen2017Hybrid,Raileanu2020RIDE:,li2021mural}. Importantly, these methods optimize for the worst case, as they try to cover \emph{all} states since the exact location of the reward is unknown. In contrast, we look at the problem of incorporating domain knowledge via reward shaping into the exploration process. Closely related to our work is that of ~\citet{cheng21heuristic}, which studies how to incorporate shaping (heuristics) into the RL process via reducing the effective horizon. This work provides gains by reducing the horizon factor while our work provides gains by reducing the size of the effective state space that needs to be searched through. 
\vspace{-0.3cm}\vspace{-0.3cm}




\section{Overview}
\label{sec:preliminary}
\vspace{-0.3cm}
We consider an episodic Markov Decision process $\Mcal = (\Scal, \Acal, \Pstar, r, H)$ where $\Scal$ corresponds to the state space, $\Acal$ is the action space, $\Pstar$ is the transition operator, $r : \mathcal{S} \times \mathcal{A} \rightarrow[0,1]$ is the reward function and $H$ is the problem horizon. We use $|\Scal|,|\Acal|$ to denote the number of state and actions, and we use $\Pstar(\cdot|s,a)$ to denote the transition probability of state action pair $(s,a)$. The value function $V^\pi(s_0)$ of a policy $\pi$ starting at an initial state $s_0$ is defined as $V^\pi(s_0) := \bb E_\pi \brac{\sum_{h=0}^Hr(s_h,a_h)}$, where $\bb E_\pi$ denote the transition dynamic of $\Mcal$ under $\pi$. Similarly, $\Vstar(s_0)$ is the value function of the optimal policy $\pi^\star$. We consider a sequential interaction between a learner and the MDP $\mathcal{M}$ occurring in rounds indexed by $t \in \mathbb{N}$. At the start of round $t$ the learner selects a policy $\pi_t$ that is used to gather a sample trajectory from $\mathcal{M}$. As is standard in the literature, we measure the learner's performance up to round $T$ by ${\mathrm{Regret}(T) := \sum_{t=1}^T   \Vstar(s_0) - V^{\pi_t}(s_0)}$.

\begin{wrapfigure}{r}{0.6\linewidth}
    \vspace{-1em}\vspace{-0.3cm}
    \centering
    \begin{subfigure}[t]{0.3\textwidth}
         \centering
            \includegraphics[width=1\textwidth]{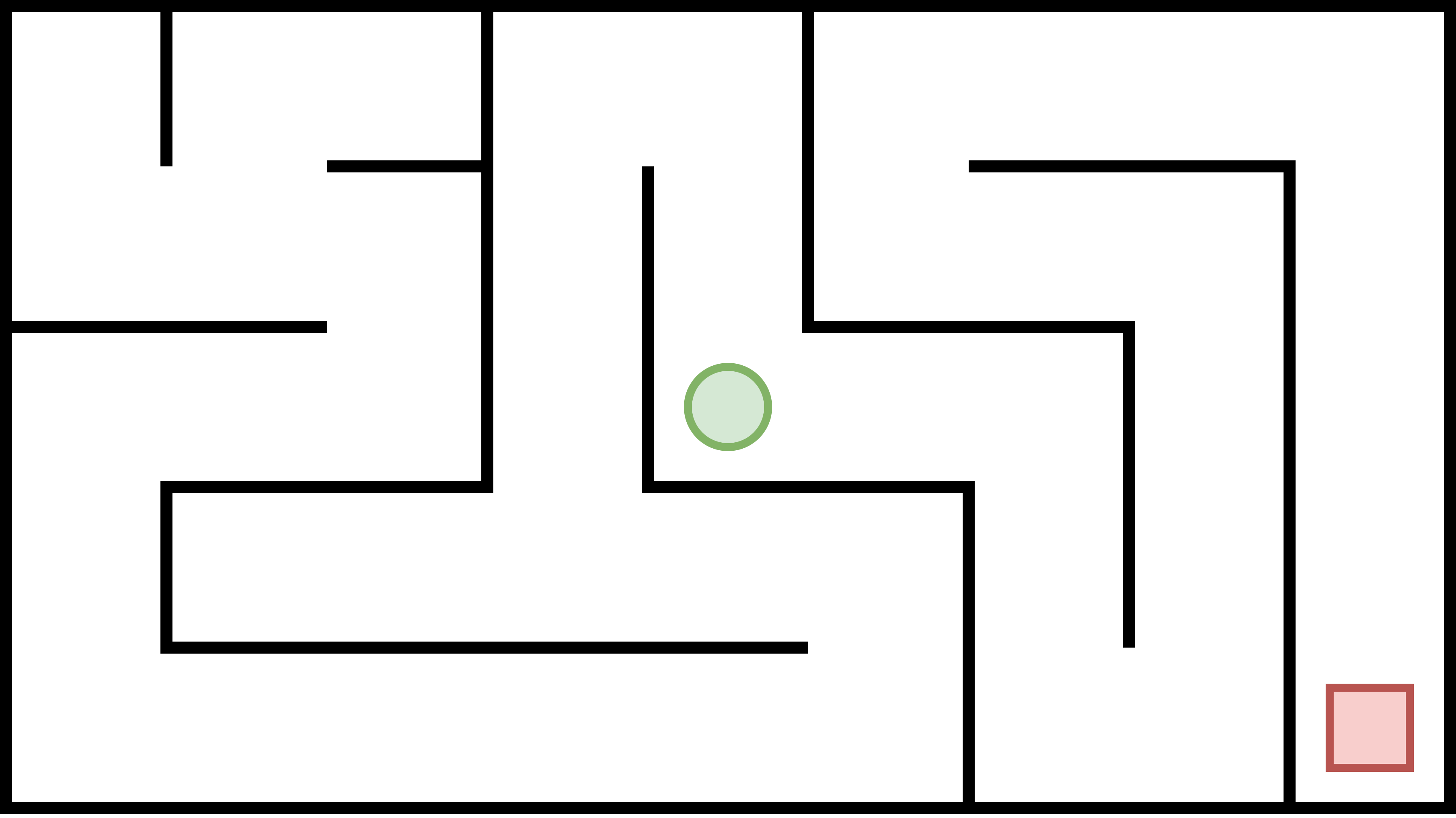}\vspace{-0.3cm}
     \end{subfigure}
     \begin{subfigure}[t]{0.3\textwidth}
         \centering
            \includegraphics[width=1\textwidth]{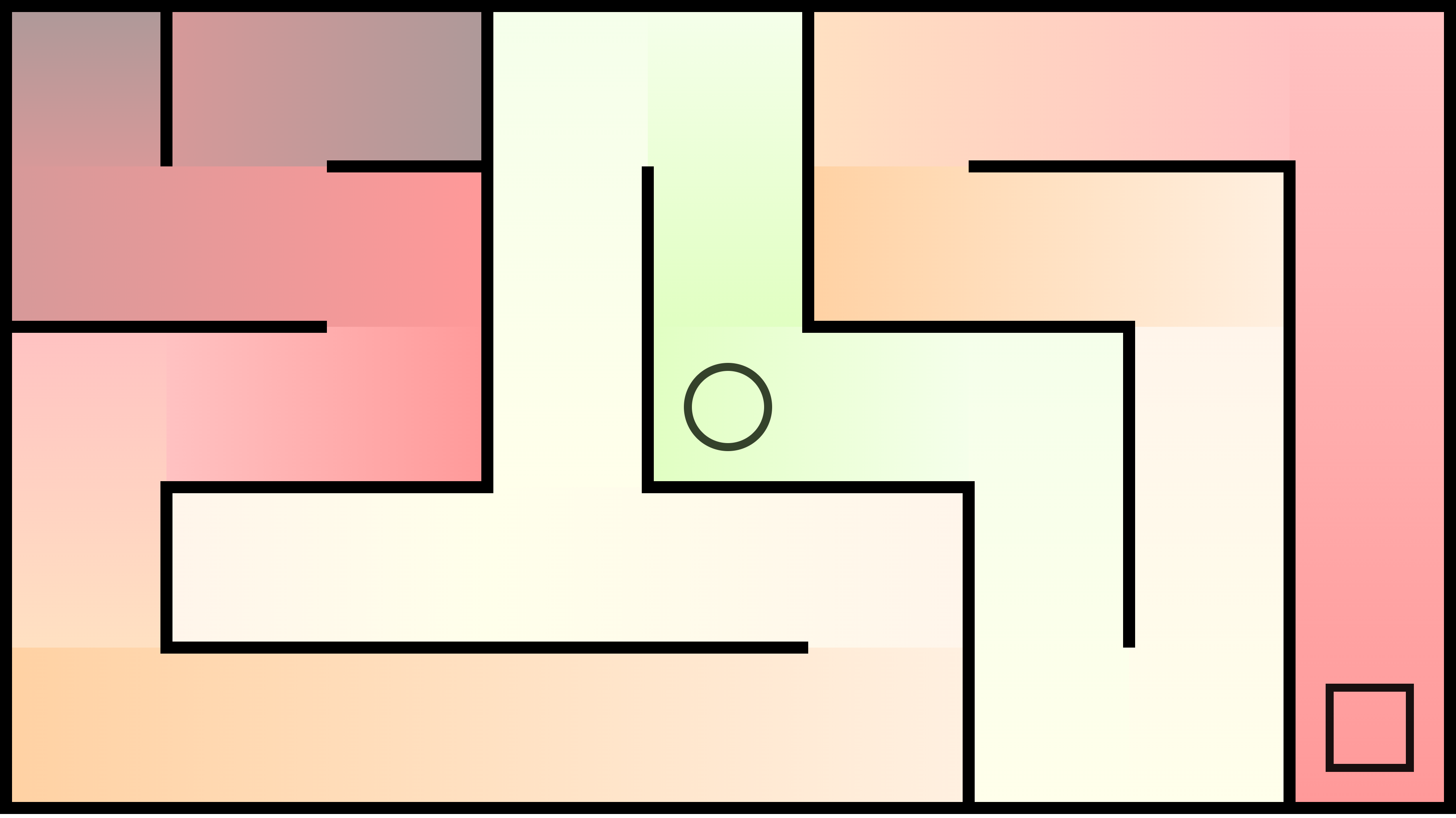}\vspace{-0.3cm}
     \end{subfigure}
      \begin{subfigure}[t]{0.3\textwidth}
         \centering
            \includegraphics[width=1\textwidth]{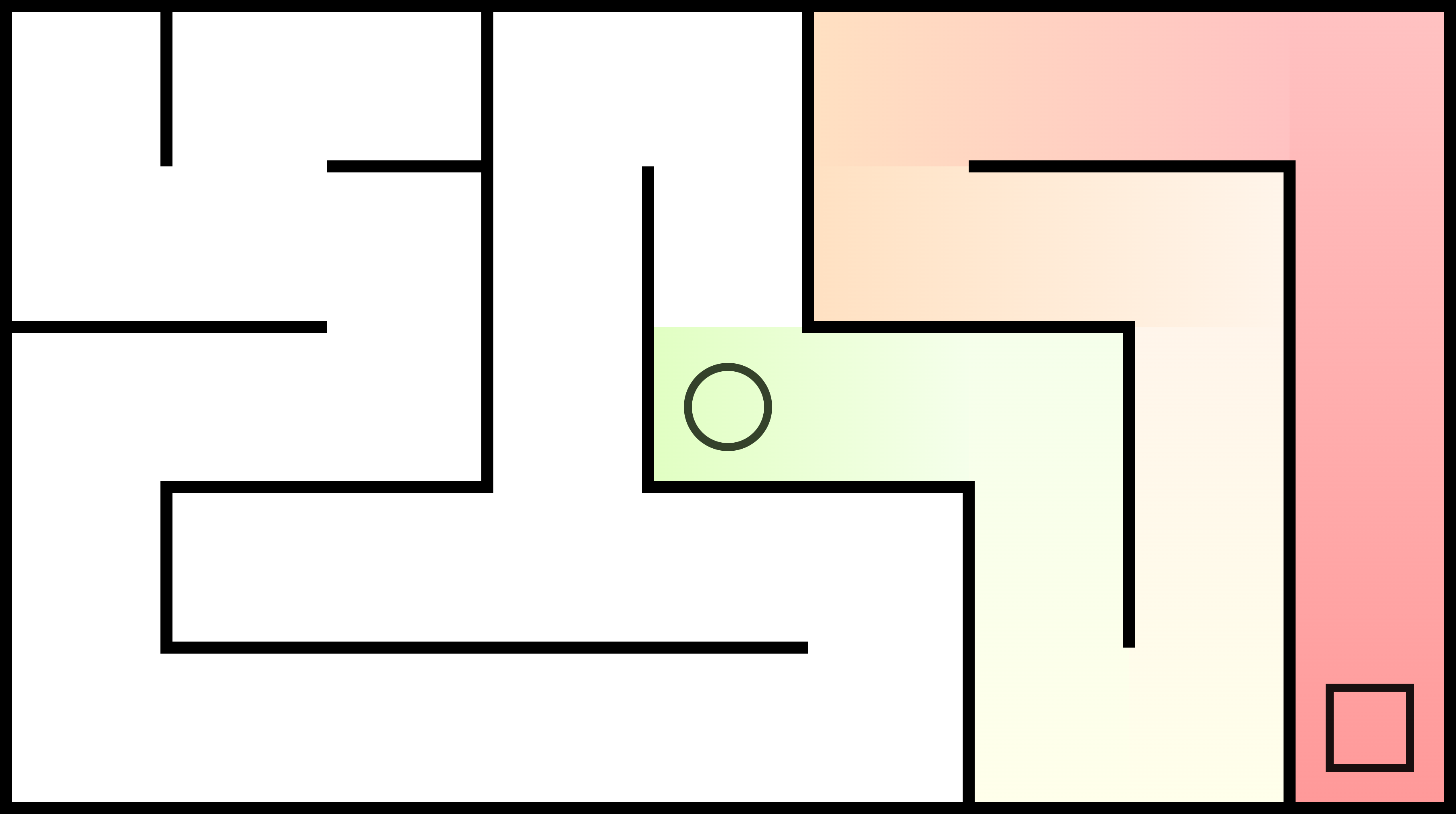}\vspace{-0.3cm}
     \end{subfigure}
    \caption{\footnotesize{Reward shaping can allow for reduction of the ``effective'' state space size. In the above maze environment, the initial state is in the middle of the maze so finding the goal only requires solving half of the maze \textbf{(Left)}. While uninformed state covering exploration \textbf{(Middle)} would go both directions since it has no knowledge of goal location, effective shaping \textbf{(Right)} would allow for halving of the effective state space. The green circle represents the starting point and the red square represents the goal. Different colors represents the landscape of the shaped value function (green indicates smaller value and red indicates larger value).}}
    \vspace{-0.3cm}
    \label{fig:shaping-illustration}
\end{wrapfigure}
Our goal is to show that knowing a reward shaping term $\Vtilde$ allows for significantly more sample efficient learning, which with high probability has a sublinear regret upper bound. This bound has a leading term that depends on an ``effective state space'' of $\mathcal{M}$, determined by the quality of a reward shaping term $\Vtilde$ and the nature of the particular MDP being solved. This pruned effective state space can be much smaller than $|\Scal|$. Additionally, we will show an improved horizon dependence as the shaping term $\Vtilde$ allows for the bonus terms to be smaller and therefore decay faster thus replacing horizon factors of $H$ with $\beta \Vtilde$ ones. 



Intuitively, reward shaping allows for the consideration of a reduced effective state space. Consider the maze environment in Fig.~\ref{fig:shaping-illustration} with agent starting in the middle. Without knowing where the goal is, an uninformed exploration algorithm needs to explore nearly the entire maze. However, an effectively incorporated reward shaping term $\Vtilde$ would allow the algorithm to prune out the entire left half of the maze, effectively halving the effective state space size. This can happen in two ways: firstly, if the shaping can directly indicate that certain actions are sub-optimal, then entire regions of the space can be eliminated from search. Secondly, even if states are not eliminated, if their corresponding bonuses are scaled according to their shaped value $\Vtilde$ from the shaping rather than uniformly with $H$, it limits unnecessary exploration of suboptimal states.

Based on this intuition, we propose modifications to a standard reinforcement algorithm that allows us to perform this analysis effectively. To aid our analysis, we introduce UCBVI-Shaped, a modification of the UCBVI algorithm~\citep{azar2017minimax} that uses an additional reward shaping term $\Vtilde$ in two ways: (1) Bonus scaling: $\Vtilde$ is used to provide an exploration bonus that combines inverse state visitation counts with reward shaping. This allows for the reduction of overoptimism and the dropping of the horizon $H$ dependence (2) Value projection: the shaping term is used to clip the empirical value function $\widehat{V}_h(s)$, which enables pruning unnecessary elements of the state space and allows for a bound that depends only on the effective state space size. 
Informally, our main result for UCBVI-Shaped can be summarized as follows:
\begin{theorem}[Main Informal] \label{theorem::informal_main}
With probability at least $1-\delta$, the regret of UCBVI - Shaped satisfies\footnote{Our main result, Theorem~\ref{theorem::main_result}, is slightly more complex than this statement. We have chosen this simplified form to aid the reader to form the right intuition. }
\begin{equation*}
\mathrm{Regret}(T)= \Ocal\left(  B(\Vtilde, \mathcal{M})\log(T/\delta) +\mathrm{Regret-UCBVI}(\Scal_{\mathrm{remain}}, \Acal, T)  \right) .
\end{equation*}
Here, $\Scal_{\mathrm{remain}}$ corresponds to the set of states in $\mathcal{M}$ where the information contained in $\Vtilde$ was not enough to rule them out, and $\mathrm{Regret-UCBVI}(\Scal_{\mathrm{remain}} \Acal, T)$ is a UCBVI regret upper bound function evaluated on $\Scal_{\mathrm{remain}}$ and $\Acal$. $B(\Vtilde, \mathcal{M})$ is a time-agnostic complexity measure that depends solely on the $\mathcal{M}$ and the quality of the shaping $\Vtilde$. 
\end{theorem}\vspace{-0.1cm}

We will define $\Scal_{\mathrm{remain}}$ and $B(\Vtilde, \mathcal{M})$ and describe their relationship to $\Vtilde$ precisely in Section~\ref{sec:analysis}. If $\Scal_{\mathrm{remain}} \ll \Scal$, the regret of UCBVI - Shaped can be substantially smaller than $\mathrm{Regret-UCBVI}(\Scal, \Acal)$, particularly since the first term of the regret grows logarithmically while the second scales with $\sqrt{T}$. We state our assumptions next:


\begin{assumption}\label{assump::sandwich}
The quality of the shaping term $\Vtilde$ is described by a parameter $\beta$. We assume access to a shaping ``value function'' estimators $\widetilde{V}_h: \mathcal{S} \rightarrow \mathbb{R}$ such that $ \Vstar_h(s)  \leq \beta \Vtilde_h(s)$, for all $s \in \Scal$, $h \in [H]$ and for some $ \beta \geq 1$.
\end{assumption}\vspace{-0.2cm}
Instead of absorbing $\beta$ into the definition of $\Vtilde_h$ we allow $\Vtilde_h(s) < \Vstar_h(s)$ for some states $s \in \Scal$. We'll show that learning a value of $\beta$ that turns this assumption true can be performed online. This assumption is intimately related to the optimistic $\Qtilde$ assumptions of~\citep{golowich2022can}. A thorough comparison with this work can be found in Appendix~\ref{section::generalizations}.  


\begin{assumption}\label{assump::rewards}
We assume the reward functions satisfy ${r(s,a) \in [0,1]},\forall (s, a) \in \Scal\times \Acal.$
\end{assumption}

\begin{assumption}\label{assump::h_indexing}
The states $\Scal$ are $h-$indexed, i.e., the states reachable at time $h \in [H]$ are disjoint from the states reachable at time $h' \in [H]$ when $h \neq h'$. 
\end{assumption}\vspace{-0.3cm}


\paragraph{Contributions.} Our main conceptual innovation is to introduce the notions of pseudosuboptimal and path-pseudosuboptimal states to quantify the ``effective'' size of the state space as a function of the quality of the shaping term $\Vtilde$ and use these notions to show how UCBVI-Shaped can attain significantly improved regret rates. In contrast with for example~\citep{golowich2022can}, where the regret rates will always scale at least with the number of states, our regret guarantees depend on an effective state size that may be orders of magnitude smaller. We believe this captures the real complexity improvement that reward shaping may yield, namely, avoid exploration of unnecessary areas of the state space. Other approaches such as~\citep{golowich2022can} may (in general) at most yield an improvement in the dependence on the effective size of the action space. We further show that incorporating shaping into the exploration bonus term improves the horizon-dependence in the bound when the shaping is good enough, allowing us to replace a leading $H$ term with $\max_s \beta \Vtilde(s)$. \vspace{-0.3cm}


\section{The UCBVI-Shaped Algorithm}
\label{sec:method}
\vspace{-0.5em}
To perform analysis of reward shaping, we build on the framework of the UCBVI algorithm~\citep{azar2017minimax}. UCBVI is an exploration algorithm based on the upper confidence bound, described in detail in Appendix D. This forms a convenient base algorithm for incorporating shaped rewards in a way that admits faster learning while maintaining asymptotic performance (as we will discuss in Section~\ref{sec:analysis}). Our algorithm, UCBVI-Shaped, is a combination of two changes to the upper confidence bound algorithm. First, we modify the bonus scaling to depend on $\Vtilde$. Second, we introduce a projection subroutine into value iteration, implemented as a standard clipping operation. 
\begin{small}
\begin{algorithm}[!h]
  \caption{UCBVI - Shaped}\label{algo:UCBVI_shaped}
  \begin{algorithmic}[1]
    \State \textbf{Input} reward function $r$ (assumed to be known), confidence parameters 
      \State \textbf{for} {$t=1,...,T$}
        \State \qquad Compute $\wh{P}_t$ using all previous empirical transition data as $\wh{\bb P}_t(s^\prime|s,a) := \frac{N_h^t(s,a,s^\prime)}{N_h^t(s,a)},\;\forall h,s,a,s^\prime$.
        \State \qquad Compute reward bonus \textcolor{NavyBlue}{$b_h^t(s, a)$ from Eqn.~\ref{eq::shaped_bonus}} (roughly of order $\frac{\Vtilde}{\sqrt{N(s,a)}}$) \Comment{\textcolor{NavyBlue}{Bonus scaling}}
        \State \qquad Run Value-Iteration with Projection (Algorithm~\ref{algo:ValueIterationProj}).  
        \State \qquad Set $\pi_t$ as the returned policy of VI.
      \State \textbf{End for }
  \end{algorithmic}
\end{algorithm}
\end{small}
As in standard UCBVI, we define $N_h^t(s,a)$ to be the visitation count for the state-action pair $(s,a)$ at iteration $t-1$ for horizon $h$: ${N_h^t(s,a) := \sum_{i=0}^{t-1}\indicator{(s^i_h,a_h^i)=(s,a)}}$. Similarly to $N_h^t(s,a)$, we use $N_h^t(s,a,s^\prime) := \sum_{i=0}^{t-1}\indicator{(s_h^i,a_h^i,s^i_{h+1}) = (s,a,s^\prime)}$ as the visitation count of state-action pair $(s,a)$ and the subsequent state $s^\prime$. We then use $\wh{\bb P}_t(s^\prime|s,a) := \frac{N_h^t(s,a,s^\prime)}{N_h^t(s,a)}$,\footnote{The $\Phat_t$ here is dependent on the horizon $h$, but since we have assumed (Assumption~\ref{assump::h_indexing}) the states $s$ are $h$-indexed, we will use $\Phat_t$ for notation simplicity.} $\;\forall h,s,a,s^\prime$ to denote the empirical transition kernels at iteration $t$. UCBVI uses value iteration with the empirical transition function $\wh{\bb P}_t$ and a reward function augmented with an exploration bonus, given by $r_h+b_h^t$. This is defined as a dynamic programming procedure that starts at $H$ and then proceeds backward in time to $h=0$, updating according to Algorithm~\ref{algo:UCBVI_shaped} and~\ref{algo:ValueIterationProj}. The key algorithmic changes between UCBVI and UCBVI-Shaped are highlighted in blue: (1) scaling bonus $b_h^t$ by the shaping term $\Vtilde$ and (2) projecting the empirical value function $\wh{V}_h^t(s)$ by an upper bound based on the sandwiched shaping, $\beta\Vtilde(s,a)$. As we discuss in Section~\ref{sec:practical}, we also show how the sandwich factor $\beta$ does need to be provided beforehand but instead can be inferred through a straightforward technique for online model selection. While the approximate order of the bouns term is $\frac{\Vtilde}{\sqrt{N(s,a)}}$, a more detailed description can be found in Section~\ref{sec:analysis}.  \vspace{-0.5em}\vspace{-0.5em}
\begin{small}
\begin{algorithm}[!h]
  \caption{Value Iteration with Projection}\label{algo:ValueIterationProj}
  \begin{algorithmic}[1]
  \State \textbf{Input} $\Brac{\wh{P}_t,r+b_h^t}_{h=0}^{H-1}$.
    \State $\wh{V}_H^t(s) =\; 0,\;\forall s,\wh{Q}_h^t(s,a)$
    \State \textbf{While} not converged
    \State \qquad $\wh{Q}_h^t(s,a) =  \min\Brac{r_h(s,a)+b_h^t(s,a)+\wh{\bb P}_t(\cdot|s,a)\cdot\wh{V}_{h+1}^t,H}$
    \State \qquad $\wh{V}^t_h(s) =\textcolor{NavyBlue}{\min \Big(}\max_a\wh{Q}_h^t(s,a)\textcolor{NavyBlue}{, \beta\Vtilde(s)\Big)}$  \Comment{\textcolor{NavyBlue}{Value projection}}
    \State $\pi_h^t(s) = \arg\max_a\wh{Q}_h^t(s,a),\;\forall h,s,a.$
  \end{algorithmic}
\end{algorithm}
\end{small}\vspace{-0.5em}\vspace{-0.5em}
\section{Analyzing UCBVI-Shaped}
\label{sec:analysis}
\vspace{-0.5em}

In this section, we will derive our main result on the sample complexity of the UCBVI-Shaped algorithm, and along the way introduce pseudosuboptimal and path-pseudosuboptimal states as a tool for deriving bounds that depend only on the ``effective'' state space as determined by the provided reward shaping. We will first introduce some notation to use in our analysis. As described in the previous section, UCBVI-Shaped proceeds in rounds. At the beginning of round $t$, the learner has access to an empirical model $\Phat_t$ built from the data collected up to iteration $t-1$. 
The bonus terms we consider are built by taking the empirical second moment of $\Vtilde$. This is related to the definition of $\mathrm{bonus_2}$ in~\citet{azar2017minimax}. Importantly, the empirical value functions $\Vhat^t_h$ are clipped above by $\beta \Vtilde_h$: \vspace{-0.4em}
\begin{equation}
    \label{eq::shaped_bonus}
    b_h^t(s,a) = \min\left( 16 \beta \sqrt{ \frac{ \widehat{\mathbb{E}}_{s' \sim \Phat_t(\cdot | s,a) }[\Vtilde^2_{h+1}(s') | s,a ]  \ln\frac{ 2 |\Scal||\Acal|}{\delta} }{ N_h^t(s,a) }}  +  \frac{12\beta \Vtildemax }{N_h^t(s,a)}\ln \frac{ 2 |\Scal||\Acal|t}{\delta}, 2\beta \Vtildemax\right),
\end{equation}
where $\Vtildemax_{h} = \max_{s'} \Vtilde_h(s')$ and $\Vtildemax = \max_{s', h'} \Vtilde_{h'}(s')$. 

Despite clipping and a modified bonus term, the $\Qhat$ and $\Vhat$ values of UCBVI-Shaped are optimistic:
\begin{restatable}{lemma}{lemmaoptimism}\label{lemma::optimism}
With probability at least $1-\delta$  we have 
\begin{equation}
    \Vhat_0^t(s_0)\geq \Vstar_0(s_0),\;\forall s_0\in \Scal; \qquad \text{and} \qquad  \Qhat_h^t(s,a) \geq \Qstar_h(s,a), \quad \forall (s,a) \in \Scal\times \Acal,
\end{equation}
for all $t,h \in \mathbb{N}\times[H]$, where $\Vhat_h^t$ is computed via Algorithm~\ref{algo:ValueIterationProj}.
\end{restatable}
The proof of Lemma~\ref{lemma::optimism} can be found in Appendix~\ref{appendix::lemma_optimism_proof}. Optimism (Lemma~\ref{lemma::optimism}) and the simulation Lemma~\citep{agarwal2019reinforcement} imply that:\vspace{-0.5em}
\begin{equation}\label{equation::simulation_optimism}
    \begin{small}
    \Vstar(s_0) - V^{\pi_t}(s_0) \leq \Vhat_1^t(s
    _0) -V^{\pi_t}(s_0)   = \mathbb{E}_{\tau \sim \pi_t }\left[ \sum_{h=1}^H b_h^t(s_h, a_h) + \left( \Phat_h^t(\cdot | s_h, a_h) - \Pstar( \cdot | s_h, a_h) \right) \cdot \Vhat_{h+1}^{\pi_t} \right].
    \end{small}
\end{equation}
We now turn our attention to characterize the information contained in $\Vtilde$. \vspace{-0.5em}


\subsection{Pruning of the State Space}\label{section::prunning_state_space}
\vspace{-0.5em}

\begin{wrapfigure}{r}{0.3\linewidth}
    \centering
    \vspace{-1em}
    \includegraphics[width=\textwidth]{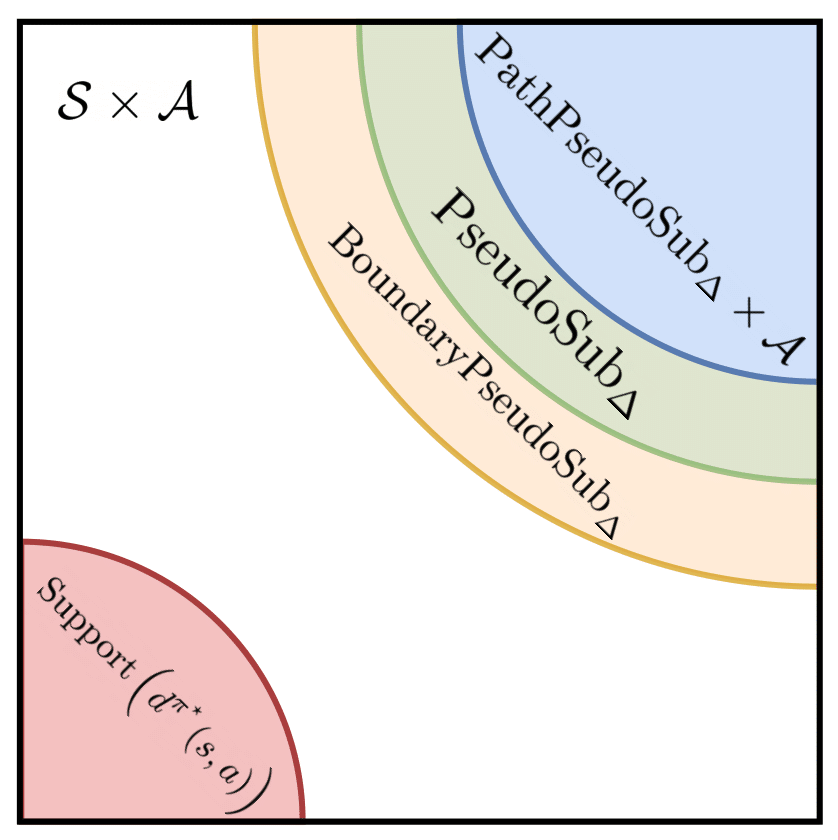}
    \caption{\footnotesize{$\pathpseudosub_\Delta \times \mathcal{A}$ is split from $\mathrm{Support}(d^{\pi^*}(s,a))$ by $\boundarypseudosub_{\Delta}$. UCBVI-Shaped can avoid exploring over $\pathpseudosub_\Delta \times \mathcal{A}$.   }}
    \label{fig:notation-explain}
\end{wrapfigure}
Consider the following ``surrogate'' Q functions, $ \Qtilde_h^u: \Scal \times \Acal \rightarrow \mathbb{R}$ induced by $\widetilde{V}$ via:
\begin{equation*}
     \widetilde Q_h^u(s,a) = \mathbb{E}_{s' \sim \mathbb{P}(\cdot | s,a)}\left[ r(s,a) + \beta \tilde{V}_{h+1}(s')\right].
\end{equation*}
By Assumption~\ref{assump::sandwich}, we can bound $Q^\star$ via $Q_h^\star(s,a) \leq \Qtilde_h^u(s,a) $. 
The basis of our main results is the following observation.
If an action $a$ satisfies $\Qtilde_h^u(s,a) < \Vstar_h(s)$ for state $s$, then $a$ is a suboptimal action for state $s$. The projection (Step 5 of Algorithm~\ref{algo:ValueIterationProj}) ensures the `empirical' $Q-$functions $\widehat{Q}_{h}^t(s,a)$ of Algorithm~\ref{algo:UCBVI_shaped} will quickly converge to values upper bounded by $\Qtilde_h^u(s,a)$. Since optimism guarantees that $\widehat{Q}_{h}^t(s,\pi_*(s)) \geq \Vstar(s)$, and the policy executed by UCBVI-Shaped is greedy w.r.t $\Qhat_h^t$, actions belonging to state-action pairs such that $\Qtilde_h^u(s,a) < \Vstar_h(s)$ will quickly stopped being played by the algorithm. Moreover, all states that are only accessible through state-action pairs of this kind will also stop being visited by the algorithm after only a few iterations. In the subsequent discussion we will call $\pseudosub$ to the set of state action pairs that are quickly `prunned out' by UCBVI-Shaped, the set of states only accessible through these state action pairs as $\pathpseudosub$ and the set of state-action pairs in $\pseudosub$ that are not in $\pathpseudosub\times \Acal$ as $\boundarypseudosub$. Once all states in $\boundarypseudosub$ have been visited enough, no states in $\pathpseudosub$ will be visited again. Provided this happens sufficiently fast, we can show a regret bound that is independent on the size of $\pathpseudosub$. The subsequent discussion is aimed at formalizing this and culminates with our main result (Theorem~\ref{theorem::main_result}).

Given a parameter $\Delta > 0$, we say that an action $a$ is $\Delta-$pseudosuboptimal\footnote{We add the qualifier pseudo to our name to distinguish between suboptimality as captured by our surrogate $Q$ values and true suboptimality between the true values of $Q_h^\star$. } for state $s$ if $\Vstar_h(s) \geq \Delta + \Qtilde_h^u(s,a)$. We denote the set of $\Delta-$pseudosuboptimal state action pairs as:
\begin{equation*}
    \pseudosub_\Delta = \{ (s,a) \in \Scal \times \Acal \text{ s.t. }   \Vstar_h(s) \geq \Delta + \Qtilde_h^u(s,a)\}.
\end{equation*}
The intuition we want to capture is that all states $\tilde s$ that can be accessed only through traversing a state action pair in $  \pseudosub_\Delta$, can be safely ignored because we can determine their suboptimality as soon as the state action pairs in $\pseudosub_\Delta$ that lead to $\tilde s$ are visited enough. 


Now we define the set of $\Delta$-path-pseudosuboptimal states as the states that can be reached only by traversing a $\Delta-$pseudosuboptimal state action pair:
\begin{equation*}
    \pathpseudosub_\Delta = \{ s \in \Scal \text{ s.t. } \text{all feasible paths from initial states to $s$ intersect $\pseudosub_\Delta$} \}.
\end{equation*}
Notice that there may be state action pairs in $\pathpseudosub_\Delta \times \Acal$ that may not be in $\pseudosub_\Delta$. In fact, there may exist states $s$ in $\pathpseudosub_\Delta$ such that no $(s,a)$
is in $\pseudosub_\Delta$ for all $ a\in \Acal$.
We will show that states in $\pathpseudosub_\Delta$ will not be explored by UCBVI-Shaped after a few iterations. Intuitively, this happens because the support of the state visitation distribution of the optimal policy does not contain any state in $\pathpseudosub_\Delta,\forall \Delta>0$ (or equivalently, ${\mathrm{Support}( d^{\pi^\star}(s,a)   ) \cap  \pathpseudosub_\Delta = \emptyset, \forall \Delta > 0}$).
Hence, once the UCBVI-shaped identifies some sub-optimal state action pairs, the algorithm will not visit these state-action pairs again.


For any state action pair $(s,a)$, define its set of neighboring states $\neighbor(s,a)$ as the set of states with nonzero probability in $\Pstar(\cdot | s,a)$. By definition of $\pathpseudosub_\Delta$, for all ${(s, a) \in (\Scal \times \Acal) \backslash ( \left(\pathpseudosub_\Delta \times \Acal \right)\cup \pseudosub_\Delta)}$, we have:
\begin{equation}\label{equation::neighbors_restricted}
    \neighbor(s,a) \subseteq \Scal  \backslash \pathpseudosub_\Delta.
\end{equation}
In other words, the neighborhood of any state action pair $(s,a)$ whose state is not in $\pathpseudosub_\Delta$ and such that $(s,a)$ is not $\Delta-$pseudo-suboptimal, are not in $\pathpseudosub_\Delta$. We also introduce the notion of ``boundary pseudosuboptimal'' state action pairs to capture the set of state action pairs that are $\Delta-$suboptimal but whose states are not in $\pathpseudosub_\Delta$. \vspace{-0.5em}
\begin{equation*}
     \boundarypseudosub_\Delta = \{ (s,a) \in \pseudosub_{\Delta} \text{ and } s \not\in \pathpseudosub_\Delta \}.  
\end{equation*}
Naturally, one of these states has to be traversed by any trajectory that contains any state in $\pathpseudosub_\Delta$.

Although we only use $\Qtilde^u$ in the definition of $\pseudosub_\Delta$, $\pathpseudosub_\Delta$, and $\boundarypseudosub_\Delta$, the size of these sets is modulated by the scale of $\beta$ and the width of the intervals $\left[\Qtilde^l(s,a),\Qtilde^u(s,a) \right]$. We will show that (in the notation of Theorem~\ref{theorem::informal_main}) $\Scal_{\mathrm{pruned}} \approx \Scal \backslash \pathpseudosub_\Delta$. With this notation, the formal version of our main results is stated as follows.

\begin{theorem}\label{theorem::main_result}
With probability at least $1-6\delta$, the regret of UCBVI-Shaped is upper bounded by
\begin{small}
\begin{align*}
\sum_{t=1}^T   \Vstar(s_0) - V^{\pi_t}(s_0) &= \Ocal\Bigg( \min_{\Delta}\Bigg(  H\beta \Vtildemax  \sqrt{| \Scal \backslash \pathpseudosub_\Delta| |\Acal| T \ln \frac{\Vtildemax |\Scal| |\Acal| T}{\delta} } +  \\
&\quad   \beta^2 \left(\Vtildemax   \right)^2 H^{1/2} |\boundarypseudosub_\Delta|^{1/2} \ln \frac{\Vtildemax |\Scal| |\Acal| T}{\delta}\times \min \left( A(\Delta), B(\Delta) \right) \Bigg)   \Bigg).
\end{align*}
\end{small}

for all $T \in \mathbb{N}$. Where $A(\Delta) =  \frac{  |\Scal|^{1/2} |\Acal|^{1/2}   }{\Delta} $ and $B(\Delta) = \frac{\beta \Vtildemax H^{1/2}|\boundarypseudosub_\Delta|^{1/2} }{\Delta^2} $.

\end{theorem}

Theorem~\ref{theorem::main_result} instantiates the desiderata of Theorem~\ref{theorem::informal_main}.
Although this regret upper bound cannot be decomposed into a sum of two term as in Theorem~\ref{theorem::informal_main}. For any fixed $\Delta$, the regret upper bound has two components, one where $\Scal_\mathrm{remain}$ can be identified with $| \Scal \backslash \pathpseudosub_\Delta|$ and a second one that scales logarithmically in $T$.  In the next section we flesh out the steps in the proof of this result. The full argument can be found in Appendix~\ref{sec:supporting_lemmas}. The bonus scaling also allows us to ameliorate the horizon dependence of the upper bound. Instead of obtaining a $H^2$ dependence as the bound of theorem 7.1 in~\cite{agarwal2019reinforcement}, the first term depends on $  H\beta\Vtildemax $.
Notice that the state dependence in the second term may be only be of order $\log(|\Scal|)$ if there is a $\Delta$ such that $|\boundarypseudosub_\Delta| \ll |\Scal|$  (albeit at the cost of a quadratic dependence on $1/\Delta^2$). Moreover notice that $|\boundarypseudosub_\Delta|$ could be much smaller than $|\pathpseudosub_\Delta|$ (the set of states that are reachable only by visiting states in $|\pseudosub_\Delta|$ ) thus showing that in some cases we can guarantee that even in the low order terms, the regret of UCBVI-Shaped that has polynomial dependence on an effective state space size that may be orders of magnitude smaller than the original one. The full version of this bound, with all the low order terms we have omitted for the sake of readability can be found in Appendix~\ref{section::full_proof_main_result}, Theorem~\ref{theorem::main_results_refined}.
Note that Theorem~\ref{theorem::main_result} is a strict generalization to the UCBVI regret upper bounds, as setting $\Delta$ to a value that is smaller than the minimum gap recovers the exact result (Theorem 7.6in~\cite{agarwal2019reinforcement}).   \vspace{-0.5em}\vspace{-0.5em}

\subsection{Proof Intuitions and Sketch for Theorem~\ref{theorem::main_result}}
\vspace{-0.5em}

\paragraph{Improved Horizon Dependence.} An empirical Bernstein bound shows that adding a bonus scaling (up to low order terms) with $\Ocal\left(\sqrt{  \widehat{\mathrm{Var}}_{s' \sim \Phat_h^t(\cdot | s,a) }(\Vstar_{h+1}(s') | s,a)  \ln\frac{ 2 |\Scal| |\Acal| t}{\delta} / N_h^t(s,a) } \right) $ is sufficient to ensure optimism. Since $\Vstar$ is not known, this variance term can be substituted by the empirical second moment of $\beta\Vtilde_{h+1}$. Finally, the scaling of these terms can be upper bounded by $\beta \Vtildemax$. Without knowledge of $\Vtilde$, achieving this scaling would be challenging, since the only proxy for $\Vstar_{h+1}$ available is $\Vhat_{h+1}^t$ or $H$ both of which may vastly overestimate it. \vspace{-0.5em}\vspace{-0.5em}\vspace{-0.5em}

\paragraph{State Pruning.} The value function clipping mechanism ensures that $\mathbb{E}_{s' \sim \Phat_h^t(\cdot | s,a) } \left[  \Vhat^t_{h+1}(s') \right] \leq \beta \mathbb{E}_{s' \sim \Phat_h^t(\cdot | s,a) } \left[  \Vtilde_{h+1}(s') \right] $
and therefore the empirical gap between $\Qhat_h^t(s,a)$ and $\Vstar(s)$ is decreasing at a rate of at most $\Ocal\left( \beta\Vtildemax/\sqrt{N_h^t(s,a)} \right)$. Since optimism ensures that $\Qhat_h^t(s,\pi^\star(a)) \geq \Vstar(s)$, once $\Qhat(s,a)< \Vstar(s)$, action $a$ will not be chosen anymore. Thus, for any $\Delta$, the number of times any state action pair in $\pseudosub_\Delta$ may be visited by UCBVI-Shaped is upper bounded by $\beta^2\left(\Vtildemax\right)^2/\Delta^2$. Since any visit to a state action pair in $\pathpseudosub_\Delta \times \mathcal{A} \cup \pseudosub_\Delta$ requires a visit to a state in $\boundarypseudosub_\Delta$, which
allows us to bound the total number of visits to a state (or a trajectory containing such a state) in $\pathpseudosub_\Delta \times \mathcal{A} \cup \pseudosub_\Delta$ by $H|\boundarypseudosub_\Delta|\beta^2\left(\Vtildemax\right)^2/\Delta^2$. 

Finally, we can split (a version of) the regret decomposition in Eqn.~\ref{equation::simulation_optimism} into two sums, one over state action pairs in  $\mathcal{U}^{\mathrm{bad}} = \pathpseudosub_\Delta \times \Acal \cup \pseudosub_\Delta$ (or trajectories intersecting $\mathcal{U}^{\mathrm{bad}}$) and a second one over state action pairs in $\mathcal{U}^{\mathrm{good}} = (\Scal \times \Acal)\backslash \left(\pathpseudosub_\Delta \times \Acal \cup \pseudosub_\Delta\right)$  (or trajectories without intersection with $\mathcal{U}^{\mathrm{bad}}$ ). We can then apply the upper bound on the visitation of state action pairs in $\pathpseudosub_\Delta \times \mathcal{A} \cup \pseudosub_\Delta$ to derive a regret upper bound over these states scaling with $1/\Delta$. Using the bound on the number of trajectories that intersect $\pathpseudosub_\Delta \times \mathcal{A} \cup \pseudosub_\Delta$,
the trajectory decomposition yields the term scaling with $1/\Delta^2$ (but having only logarithmic state dependence) in Theorem~\ref{theorem::main_result}.
Recall that Eqn.~\ref{equation::neighbors_restricted} implies that the transition operators over state action pairs in $\mathcal{U}^{\mathrm{good}}$ have support only over $\Scal \backslash \pathpseudosub_\Delta$. Our proofs use this fact to prove a polynomial dependence on $|\Scal \backslash \pathpseudosub_\Delta|$ and not $|\Scal|$  in Theorem~\ref{theorem::main_results_refined} (see Appendix ~\ref{section::full_proof_main_result}), the full version of Theorem~\ref{theorem::main_result}. Detailed proofs are in Appendix~\ref{sec:supporting_lemmas}.   
\vspace{-0.5em}\vspace{-0.5em}
\section{Practical Considerations: Online Model Selection}
\label{sec:practical}

Now one may notice that UCBVI-Shaped requires knowledge of the scaling $\beta$ is in order to actually perform the value projection. While this can be pre-provided by a user or set conservatively, in this section we discuss how this can be inferred by viewing this as an online model selection problem. In particular, given a set of $N$ different values of $\beta$ \textemdash~ $\left[ \beta_1, \beta_2, \dots, \beta_N\right]$, each of which parameterizes a different setting of the learning algorithm UCBVI-Shaped($\beta$), an online model selection algorithm such as Stochastic CORRAL~\citep{agarwal2017corralling,pacchiano2020model} or RegretBalancing~\citep{pacchiano2020regret,cutkosky2021dynamic} jointly infers the value of $\beta$ and learns the appropriate value function online. In Appendix~\ref{section::model_selection} we show how these techniques can yield meaningful regret guarantees and we provide pseudocode.  

\section{Numerical Simulations}
\label{sec:experiments}
\begin{figure}[ht]
    \centering
    \begin{subfigure}[b]{0.3\textwidth}
         \centering
         \includegraphics[height=0.4\linewidth]{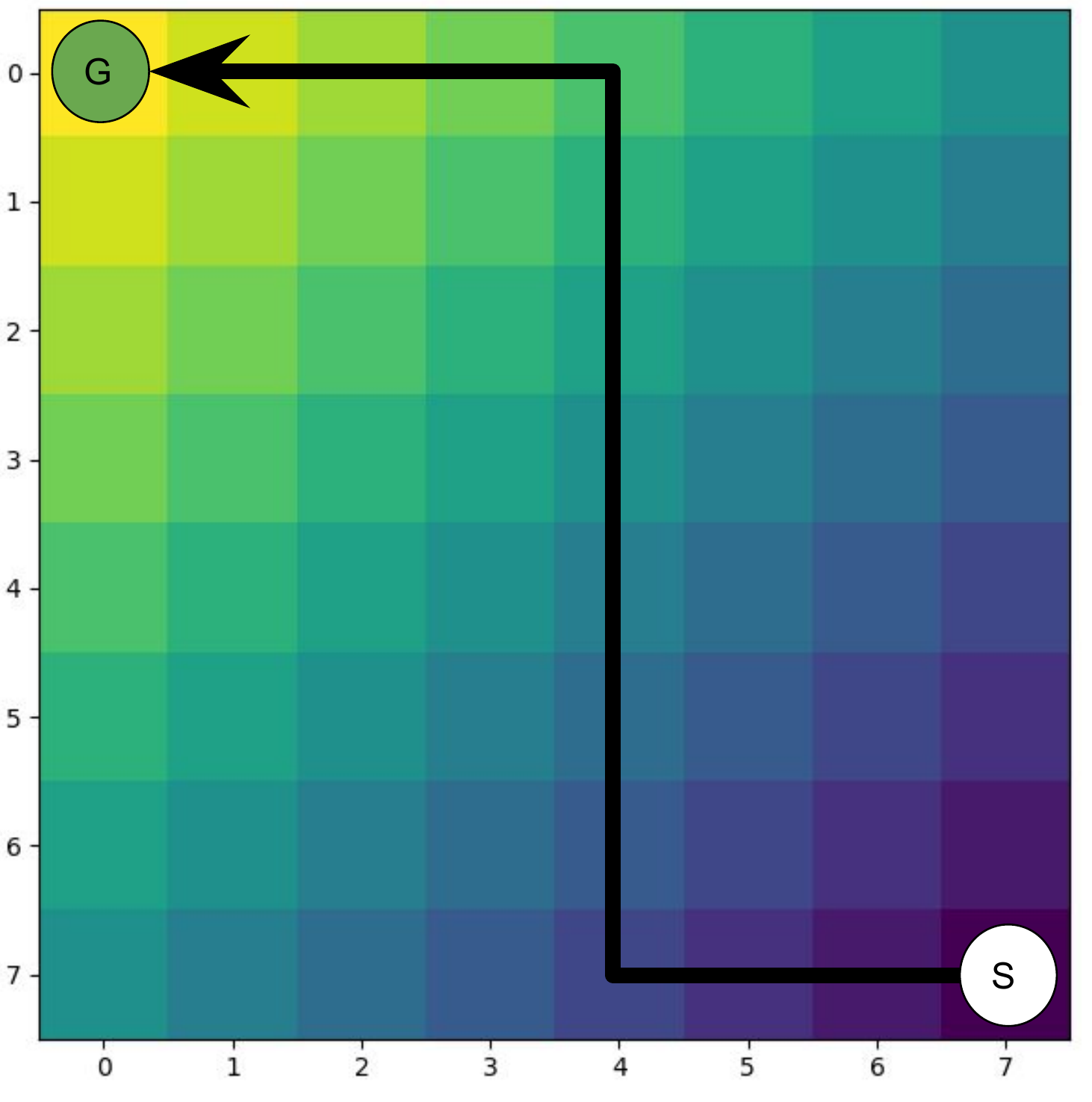}
         \caption{Value function of Gridworld}
     \end{subfigure}
     \begin{subfigure}[b]{0.3\textwidth}
         \centering
         \includegraphics[height=0.4\linewidth]{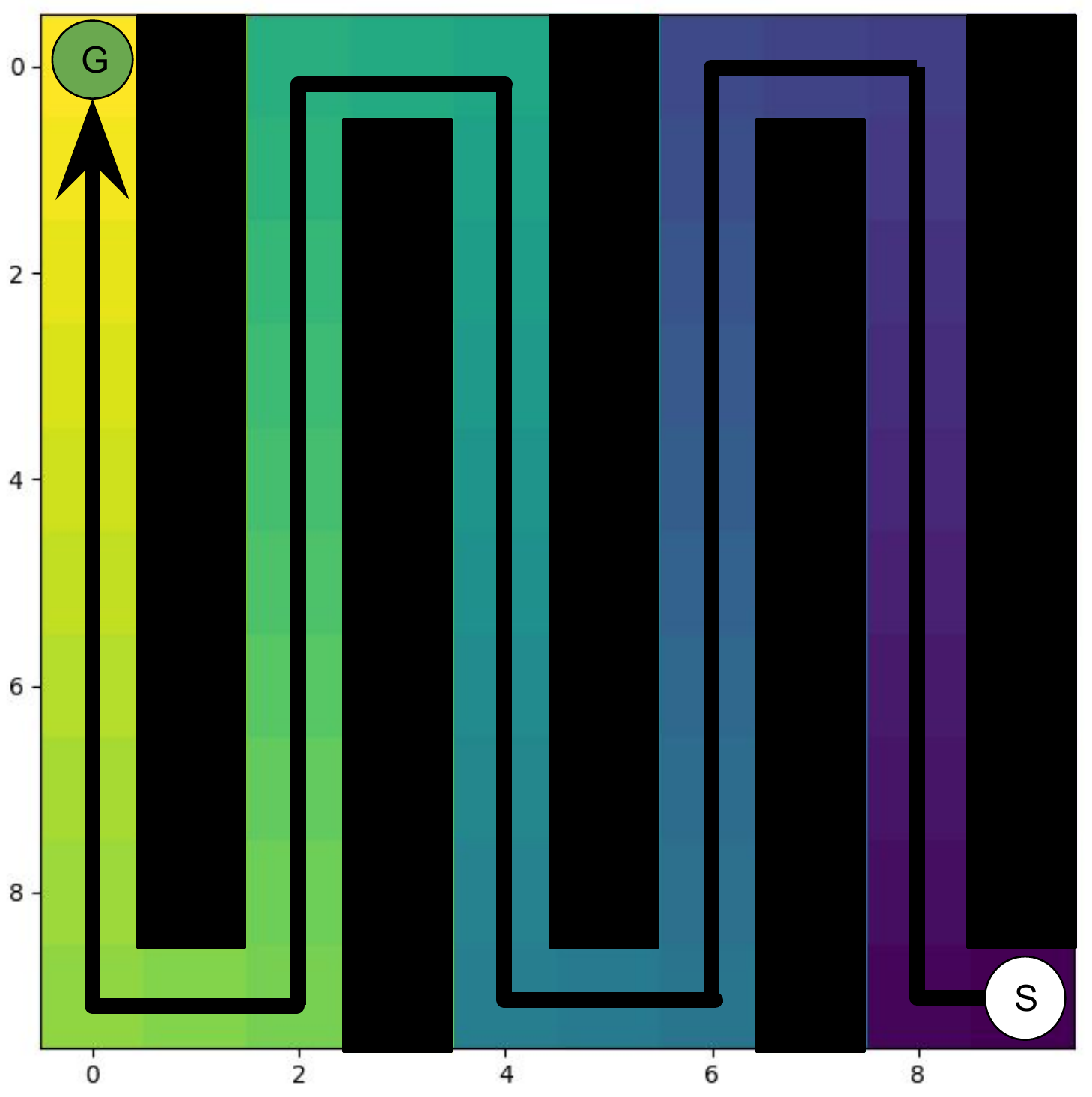}
         \caption{Value function of Single Corridor}
     \end{subfigure}
     \begin{subfigure}[b]{0.3\textwidth}
         \centering
         \includegraphics[height=0.4\linewidth]{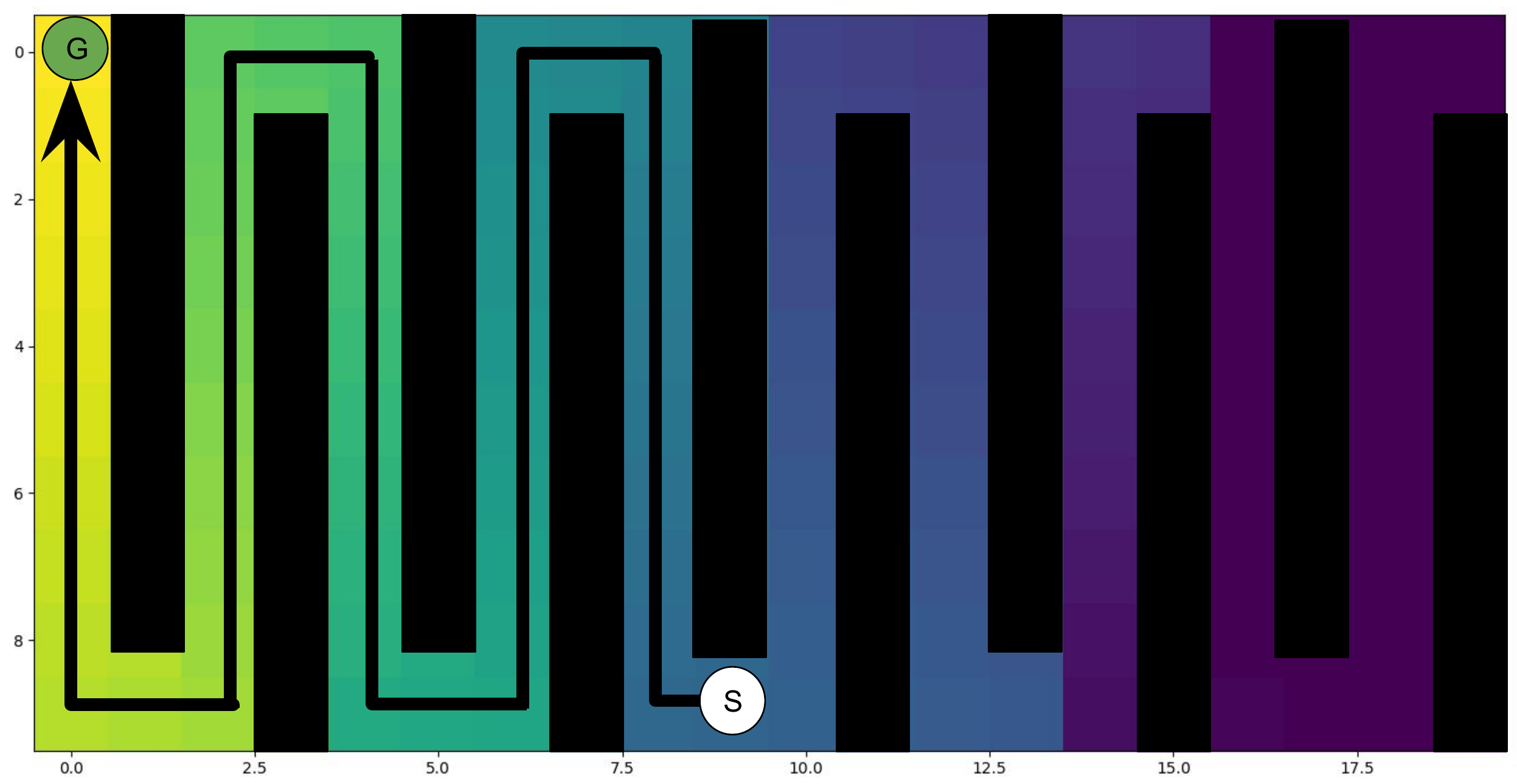}
         \caption{Value function of Double Corridor}
     \end{subfigure}
    \caption{\footnotesize{Environments used for numerical simulations. \textbf{(Left)} Open Gridworld  \textbf{(Middle)} Single corridor, agent starts bottom right has to reach a goal in top left  \textbf{(Right)} Double corridor, agent starts in the middle and has to reach a goal on the left, with many irrelevant states on the right hand side}}
    \label{fig:envs}
    \vspace{-1em}
\end{figure}

\vspace{-1em}
To show the practical relevance of our analysis on reward shaping we perform some numerical simulations on a family of maze environments with tabular state-action representations, as shown in Fig.~\ref{fig:envs}. These environments have deterministic dynamics and reward. The reward is $0$ at all states except a goal sink state, which has reward $1$. These simulations are aimed at studying the following questions: \textbf{(1)} Does reward shaping improve sample complexity in these types of maze environments over uninformed UCBVI? \textbf{(2)} What is the relative importance of the bonus reweighting and the $\Vtilde$ projection? \textbf{(3)} How does the ``suboptimality'' of $\Vtilde$ impact the resulting sample complexity? \textbf{(4)} Does introducing decayed shaping actually allow for policy convergence? In these experiments, $\Vtilde$ is constructed by scaling the optimal value function $\Vstar$ by per-state scaling factors sampled independently within the range $\beta$. 
\vspace{-0.5em}

\vspace{-1em}
\subsection{Does reward shaping help direct exploration over optimism under uncertainty?}
\begin{figure}[!ht]
    \vspace{-1em}
    \centering
     
    \begin{subfigure}[b]{0.3\textwidth}
         \centering
         \includegraphics[width=\linewidth]{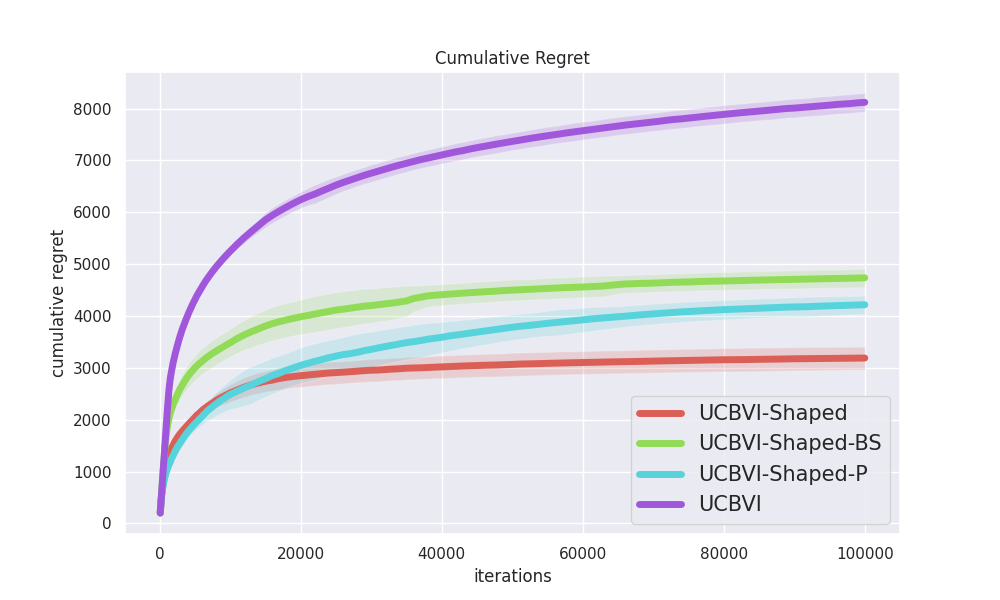}
         \caption{Open Gridworld, $\beta = 1.5$}
     \end{subfigure}
     \begin{subfigure}[b]{0.3\textwidth}
         \centering
         \includegraphics[width=\linewidth]{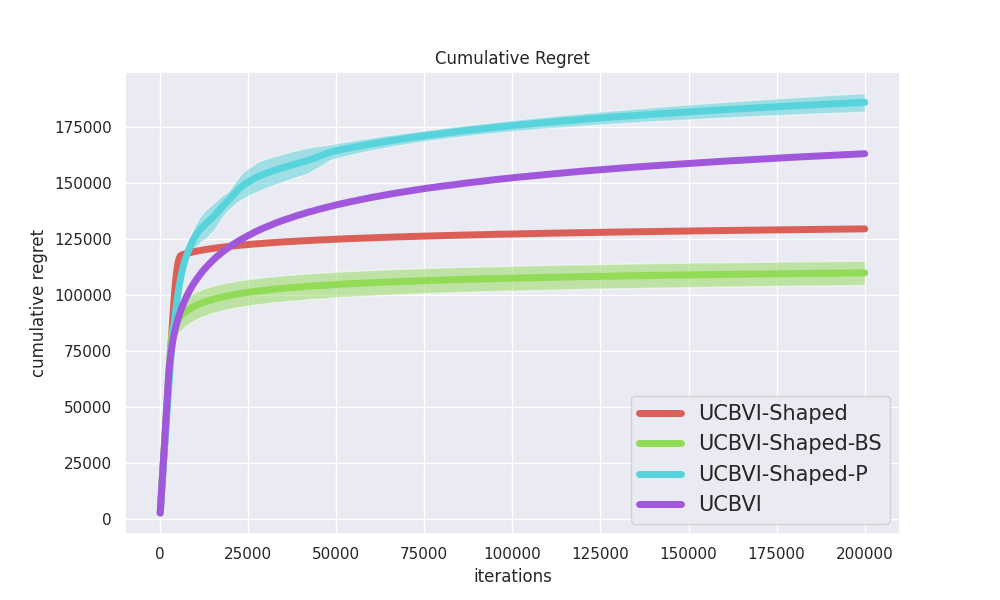}
         \caption{Single Corridor, $\beta = 1.5$}
     \end{subfigure}
     \begin{subfigure}[b]{0.3\textwidth}
         \centering
        \includegraphics[width=\linewidth]{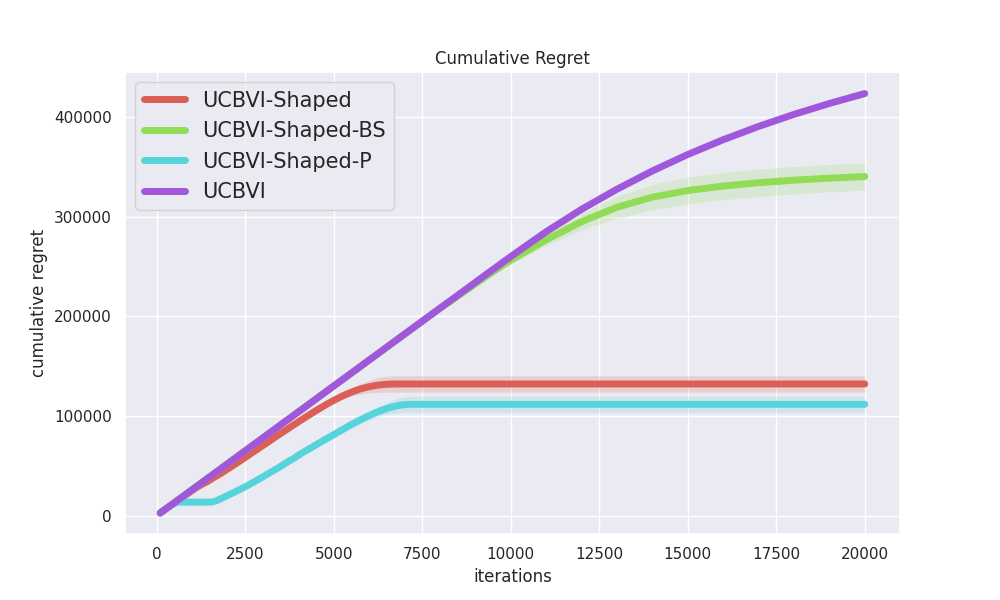}
         \caption{Double Corridor, $\beta = 1.5$}
     \end{subfigure} \\
     
     \begin{subfigure}[b]{0.3\textwidth}
         \centering
         \includegraphics[width=\linewidth]{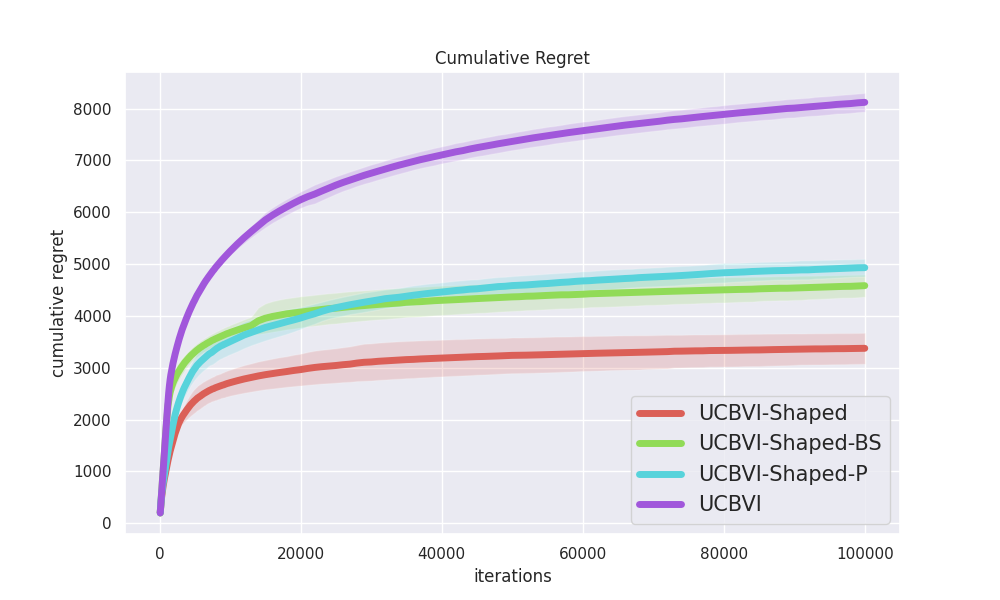}
         \caption{Open Gridworld, $\beta = 1.9$}
     \end{subfigure}
     \begin{subfigure}[b]{0.3\textwidth}
         \centering
         \includegraphics[width=\linewidth]{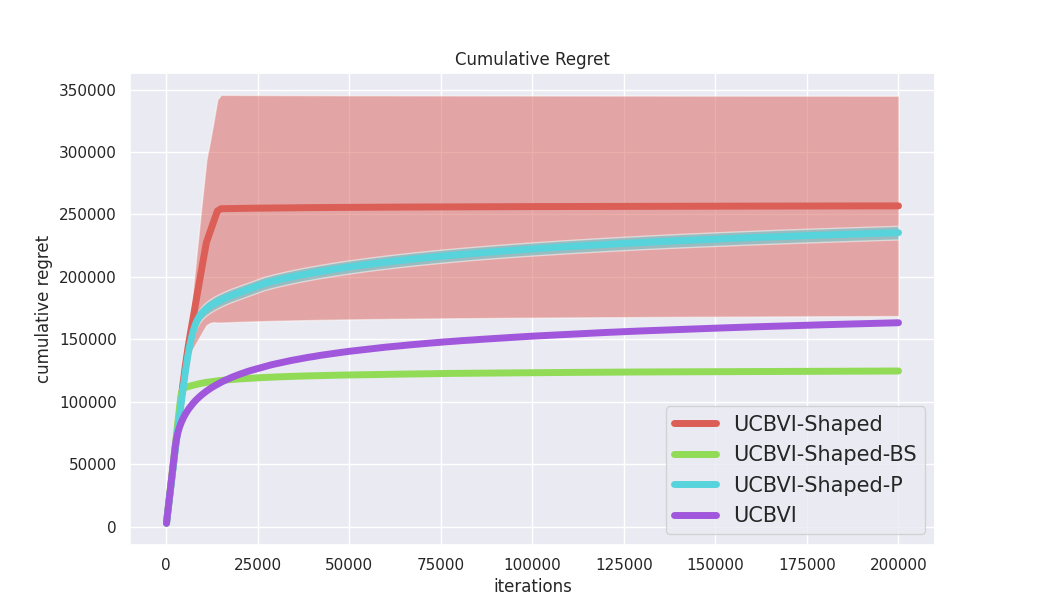}
         \caption{Single Corridor, $\beta = 1.9$}
     \end{subfigure}
     \begin{subfigure}[b]{0.3\textwidth}
         \centering
        \includegraphics[width=\linewidth]{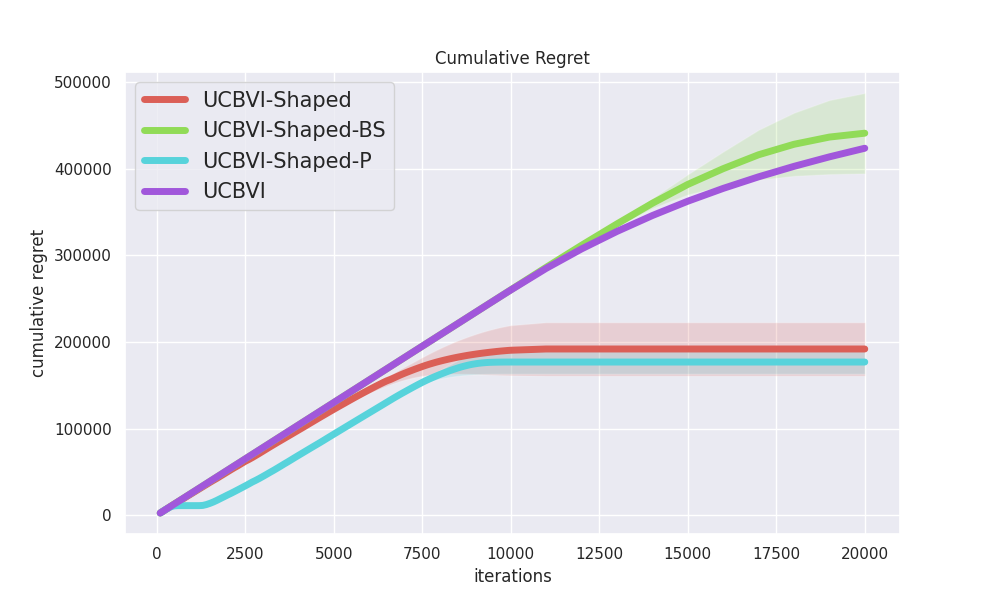}
         \caption{Double Corridor, $\beta = 1.9$}
     \end{subfigure}
    \caption{Cumulative regret for learning in various environments with varying amounts of shaping, as compared with UCBVI, and ablations UCBVI-Shaped-BS (no projection) and UCBVI-Shaped-P (no bonus scaling).}
    \label{fig:comparisons_ablations_ucbvi}
    \vspace{-0.5em}
\end{figure}

\vspace{-0.5em}
We conducted numerical simulations on tabular environments to understand how reward shaping via UCBVI-Shaped provides benefits over standard UCBVI. Fig.~\ref{fig:comparisons_ablations_ucbvi} shows cumulative regret accumulated with different variants of UCBVI-Shaped (with both projection and bonus scaling), UCBVI-Shaped-P (with only projection),  UCBVI-Shaped-BS (with only bonus scaling), UCBVI (standard UCBVI without shaping, as described in ~\cite{azar2017minimax}). This is benchmarked across the three environments described in Fig.~\ref{fig:envs}, with various levels of imperfect shaping applied by varying $\beta = \{1.5, 1.9\}$. As seen from Fig.~\ref{fig:comparisons_ablations_ucbvi}, across all environments UCBVI-shaped with projection and bonus scaling performs most favorably, followed typically by UCBVI-Shaped-P, followed by UCBVI-Shaped-BS and UCBVI, suggesting that reward shaping can significantly help with learning efficiency.

\vspace{-1em}
\subsection{How does the effectiveness of reward shaping vary across environments?}
\vspace{-0.5em}
\begin{wrapfigure}{r}{0.7\linewidth}
    \vspace{-1em}
    \centering
    \begin{subfigure}[b]{0.45\textwidth}
        \includegraphics[width=\linewidth]{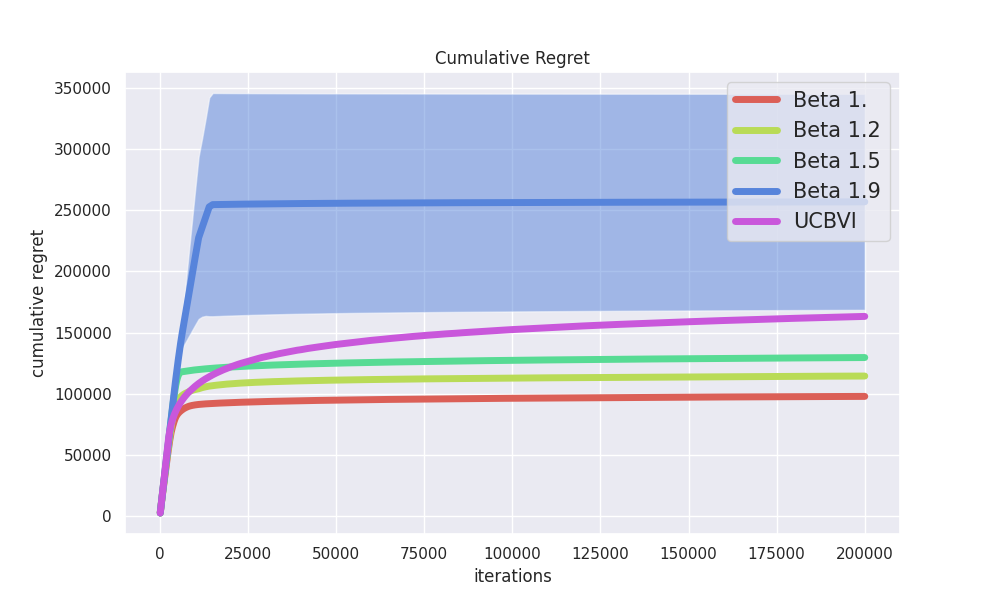}
        \caption{Single Corridor}
    \end{subfigure}
    \begin{subfigure}[b]{0.45\textwidth}
        \includegraphics[width=\linewidth]{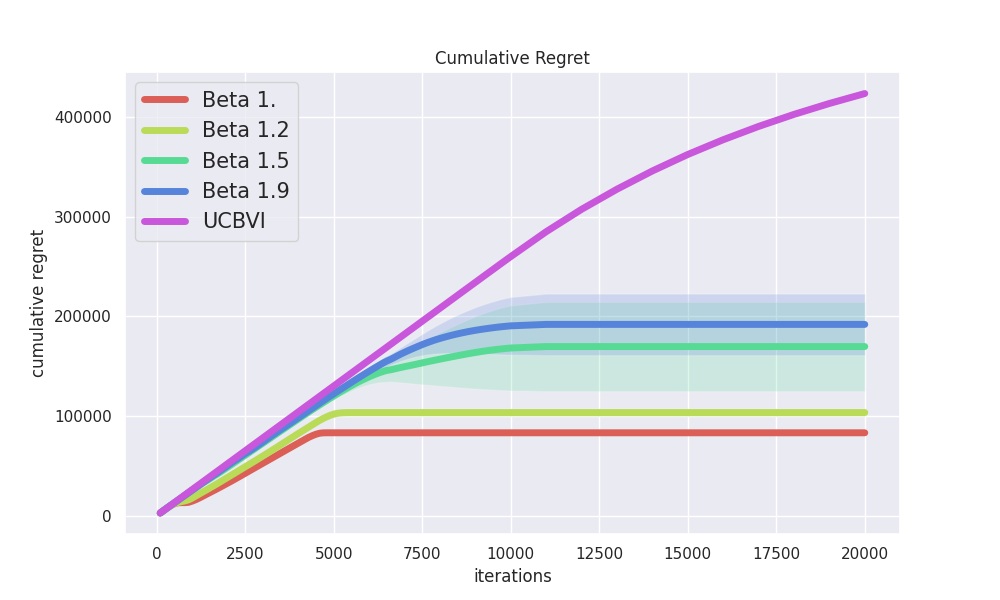}
        \caption{Double Corridor}
    \end{subfigure}
    \caption{Effect of suboptimality of reward shaping on the performance of UCBVI-Shaped. While $\beta=1.2, 1.5$ don't make much of a difference, very large $\beta$ leads to performance degradation}
    \label{fig:suboptimal-shaping}
    \vspace{-0.5em}
\end{wrapfigure}
\begin{figure}
     \centering
     \begin{subfigure}[b]{0.26\textwidth}
         \centering
         \includegraphics[width=0.8\textwidth]{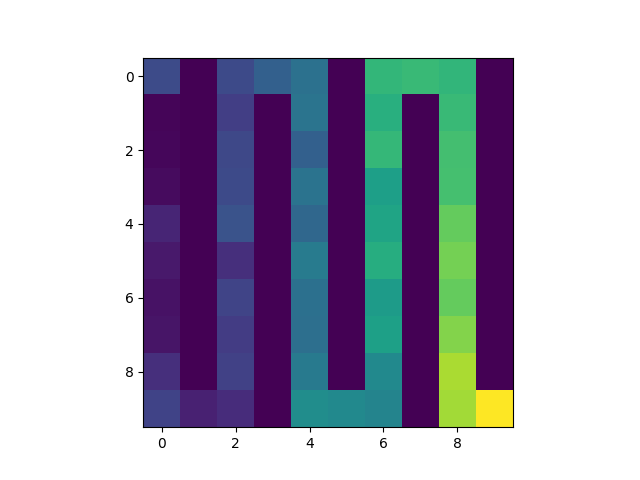}
         \caption{Heatmap of intermediate visitations of UCBVI on single corridor (no shaping)}
     \end{subfigure}
     \begin{subfigure}[b]{0.26\textwidth}
         \centering
         \includegraphics[width=0.8\textwidth]{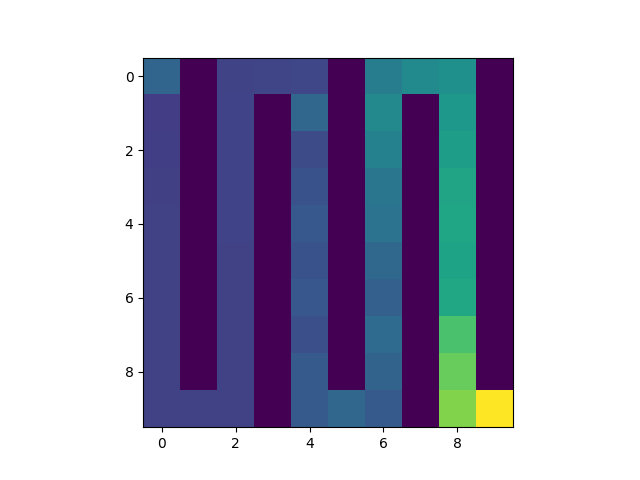}
         \caption{Heatmap of intermediate visitations of UCBVI-Shaped on single corridor}
     \end{subfigure}
      \begin{subfigure}[b]{0.26\textwidth}
         \centering
         \includegraphics[width=\textwidth]{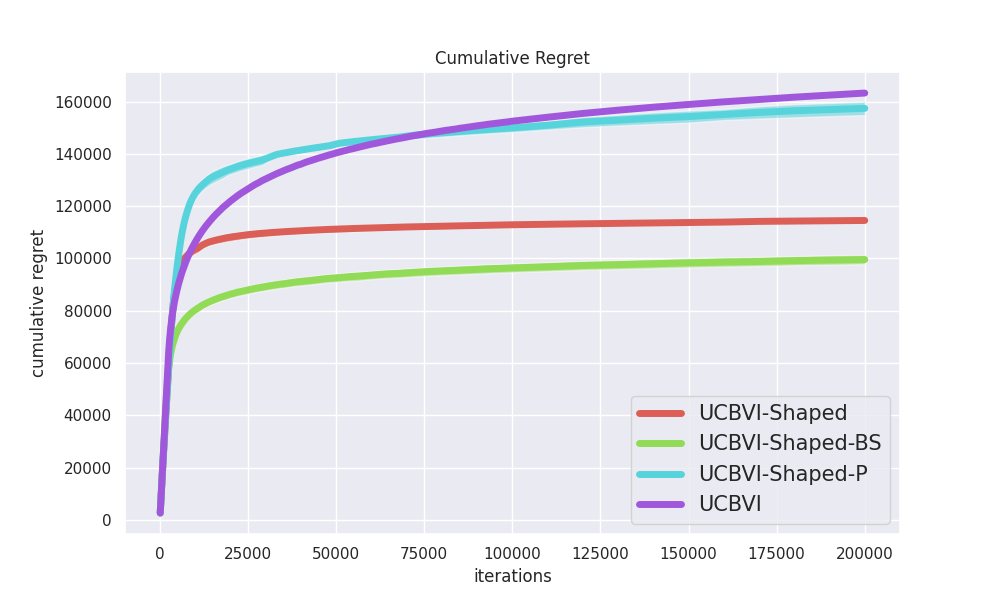}
         \caption{Learning progress of UCBVI-Shaped compared to UCBVI on single corridor}
     \end{subfigure}
     \\
     \begin{subfigure}[b]{0.26\textwidth}
         \centering
         \includegraphics[width=\textwidth]{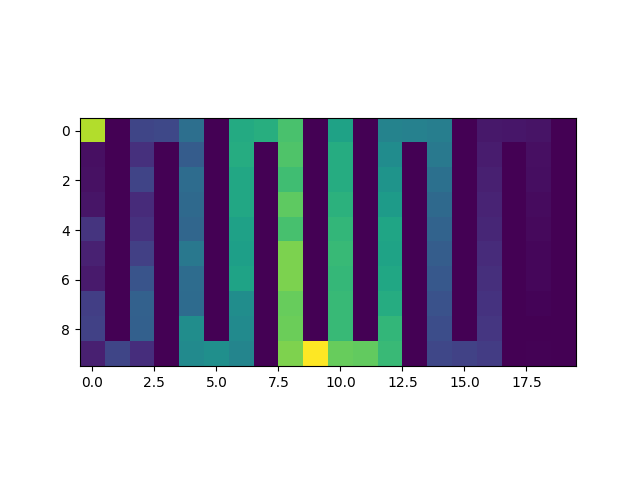}
         \caption{Heatmap of intermediate visitations of UCBVI on double corridor (no shaping)}
     \end{subfigure}
     \begin{subfigure}[b]{0.26\textwidth}
         \centering
         \includegraphics[width=\textwidth]{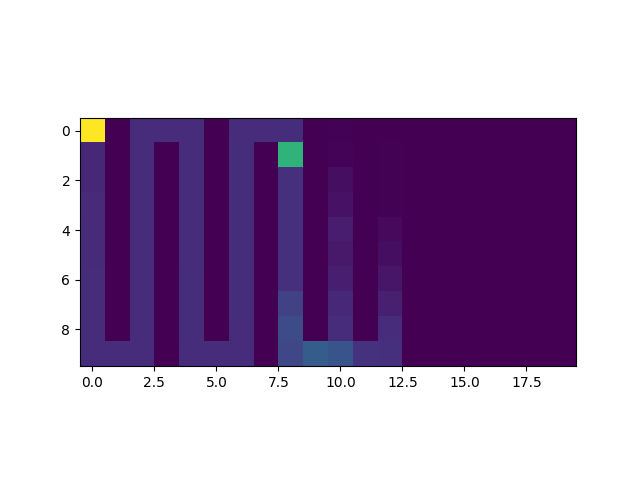}
         \caption{Heatmap of intermediate visitations of UCBVI-Shaped on double corridor}
     \end{subfigure}
    \begin{subfigure}[b]{0.26\textwidth}
         \centering
         \includegraphics[width=\textwidth]{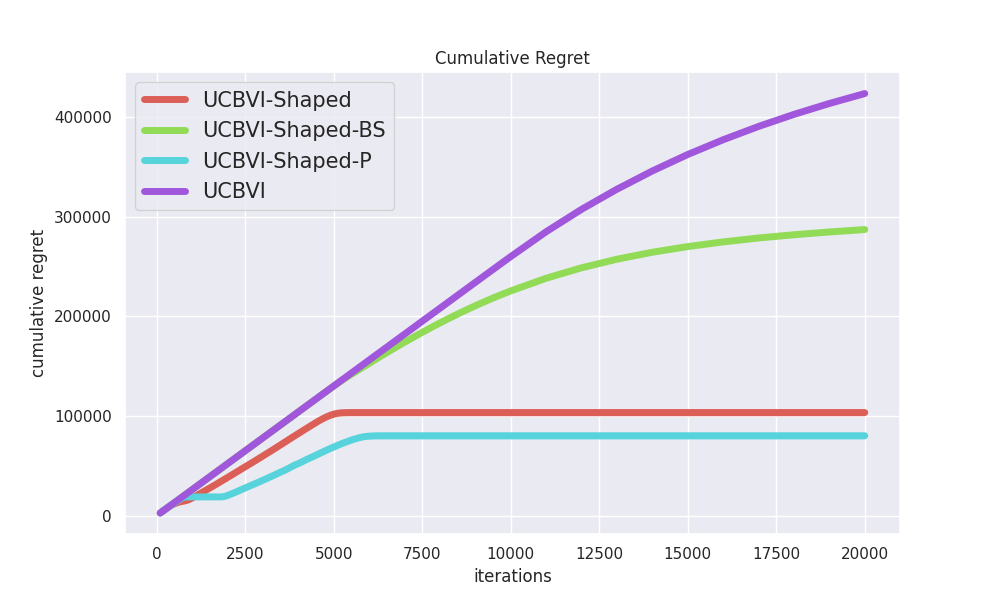}
         \caption{Learning progress of UCBVI-Shaped compared to UCBVI on double corridor}
     \end{subfigure}
        \caption{Visualization of how different environments are affected by reward shaping differently. (Left) intermediate visitations (Right) learning progress of UCBVI vs UCBVI-Shaped. The single corridor environment on the top sees much smaller gains for UCBVI-Shaped compared to double corridor environment.}
        \label{fig:viz_across_envs}
        \vspace{-1em}
\end{figure}

We next conducted some numerical simulations across environments to understand how the nature of the environment itself affects the sample complexity of learning with shaping via UCBVI-Shaped. As shown in Fig.~\ref{fig:viz_across_envs}, we see that UCBVI shaped can be very effective in environments with many irrelevant sub-optimal paths like the double corridor environment in Fig.~\ref{fig:envs}, but is relatively less effective in environments where all exploration is directed the same way such as the single corridor. Even incorrect but optimistic shaping will provide guidance towards the goal, making UCBVI relatively less dominant in the single corridor environment as compared to the double corridor. This suggests that in environments where ruling out an entire part of the exploration space is easy from the shaping, we can expect to see larger benefits.

\subsection{How does the suboptimality of reward shaping affect learning?}

We next compared how different levels of suboptimality of the $\beta$ sandwich term in the reward shaping affect cumulative regret across environments. As shown in Fig.~\ref{fig:comparisons_ablations_ucbvi} (a) and (d), we see that for environments with open paths (like the open gridworld), the shaping degradation has minimal negative effect until it gets very suboptimal. On the other hand, for corridor and double corridor (Fig.~\ref{fig:suboptimal-shaping} (b)), where there are only a few paths to the goal, suboptimal reward shaping along those paths significantly hamper progress. 




\subsection{Is online UCBVI-Shaped able to infer $\beta$ online without prior knowledge?}

\begin{wrapfigure}{r}{0.3\linewidth}
\vspace{-1.5em}
    \centering
    \includegraphics[width=\linewidth]{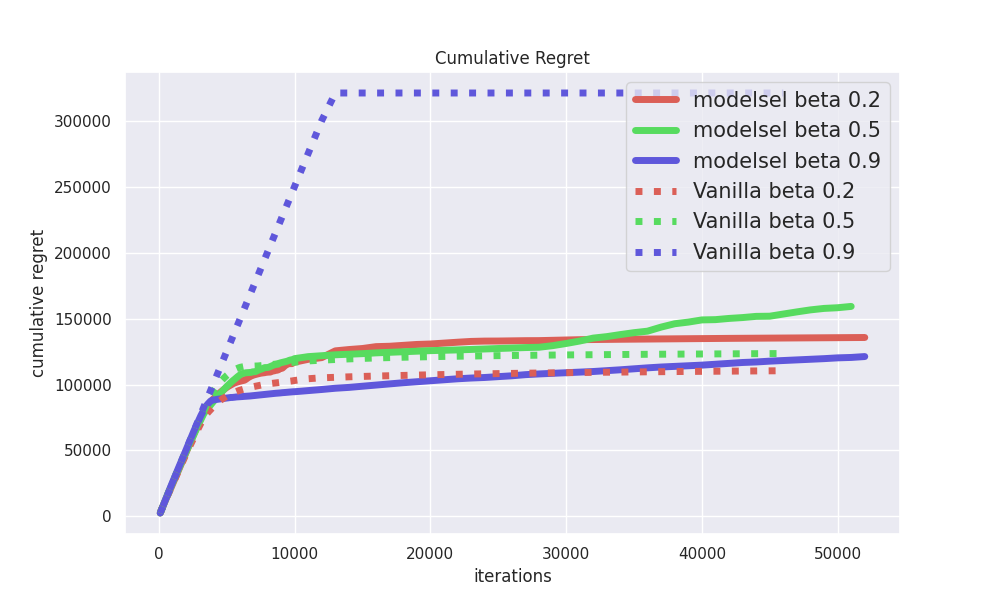}
    \caption{\footnotesize{Understanding performance of online model selection in UCBVI-Shaped}}
    \label{fig:online-exps}
    \vspace{-1em}
\end{wrapfigure}

As described in Section~\ref{sec:practical}, UCBVI-shaped can be freed of the assumption of $\beta$ being known by performing online model selection of $\beta$ and learning values jointly. In particular, we use the Stochastic CORRAL algorithm~\citep{pacchiano2020model}, a variant of the method introduced in~\citep{agarwal2017corralling} to perform online model selection, with the episodic return being the requisite criterion for updating the model selection distribution. As we see in Fig~\ref{fig:online-exps}, this scheme is able to show comparable results to when the actual $\beta$ is known beforehand, only degrading as the value of $\beta$ increases. This suggests that online UCBVI-shaped can be practical in regimes with moderate levels of value corruption.

\vspace{-0.5em}
\section{Discussion} 
\label{sec:discussion}
\vspace{-1em}
In this work, we take a step towards formally analyzing the benefits of reward shaping, proving that effective reward shaping can lead to more efficient learning than uninformed exploration strategies. In our analysis, we study an algorithm that incorporates reward shaping into a modified version of UCBVI, using it to modify bonuses and value function projection. Our analysis shows that incorporating shaped rewards allows for pruning significant parts of the state space and sharpening of optimism in a task directed way. This reduces the dependence of the regret bound on the state space size and on the horizon, depending on the quality of the shaping term and parameters of the MDP. This shows how reward shaping can direct exploration and provide significant sample complexity benefits while retaining asymptotic performance. We hope that this work is a step towards moving sample complexity analysis away from being reward agnostic to actually considering reward shaping more formally in analysis. 

\paragraph{Acknowledgements:} This work was partially supported by the Office of Naval Research. We would like to thank members of the RAIL lab at Berkeley, Improbable AI lab at MIT and Microsoft Research NYC for helpful discussions. 

\normalsize
\bibliographystyle{abbrvnat}
\bibliography{references}

\begin{thebibliography}{57}
\providecommand{\natexlab}[1]{#1}
\providecommand{\url}[1]{\texttt{#1}}
\expandafter\ifx\csname urlstyle\endcsname\relax
  \providecommand{\doi}[1]{doi: #1}\else
  \providecommand{\doi}{doi: \begingroup \urlstyle{rm}\Url}\fi

\bibitem[Abbasi-Yadkori et~al.(2020)Abbasi-Yadkori, Pacchiano, and
  Phan]{abbasi2020regret}
Y.~Abbasi-Yadkori, A.~Pacchiano, and M.~Phan.
\newblock Regret balancing for bandit and rl model selection.
\newblock \emph{arXiv preprint arXiv:2006.05491}, 2020.

\bibitem[Agarwal et~al.(2017)Agarwal, Luo, Neyshabur, and
  Schapire]{agarwal2017corralling}
A.~Agarwal, H.~Luo, B.~Neyshabur, and R.~E. Schapire.
\newblock Corralling a band of bandit algorithms.
\newblock In \emph{Conference on Learning Theory}, pages 12--38. PMLR, 2017.

\bibitem[Agarwal et~al.(2019)Agarwal, Jiang, Kakade, and
  Sun]{agarwal2019reinforcement}
A.~Agarwal, N.~Jiang, S.~M. Kakade, and W.~Sun.
\newblock Reinforcement learning: Theory and algorithms.
\newblock \emph{CS Dept., UW Seattle, Seattle, WA, USA, Tech. Rep}, 2019.

\bibitem[Agarwal et~al.(2020)Agarwal, Henaff, Kakade, and Sun]{agarwal2020pc}
A.~Agarwal, M.~Henaff, S.~Kakade, and W.~Sun.
\newblock Pc-pg: Policy cover directed exploration for provable policy gradient
  learning.
\newblock \emph{Advances in Neural Information Processing Systems},
  33:\penalty0 13399--13412, 2020.

\bibitem[Agrawal and Jia(2017)]{agrawal2017posterior}
S.~Agrawal and R.~Jia.
\newblock Posterior sampling for reinforcement learning: worst-case regret
  bounds.
\newblock \emph{arXiv preprint arXiv:1705.07041}, 2017.

\bibitem[Andrychowicz et~al.(2020)Andrychowicz, Baker, Chociej, Jozefowicz,
  McGrew, Pachocki, Petron, Plappert, Powell, Ray,
  et~al.]{andrychowicz2020learning}
O.~M. Andrychowicz, B.~Baker, M.~Chociej, R.~Jozefowicz, B.~McGrew,
  J.~Pachocki, A.~Petron, M.~Plappert, G.~Powell, A.~Ray, et~al.
\newblock Learning dexterous in-hand manipulation.
\newblock \emph{The International Journal of Robotics Research}, 39\penalty0
  (1):\penalty0 3--20, 2020.

\bibitem[Auer et~al.(2008)Auer, Jaksch, and Ortner]{auer2008near}
P.~Auer, T.~Jaksch, and R.~Ortner.
\newblock Near-optimal regret bounds for reinforcement learning.
\newblock \emph{Advances in neural information processing systems}, 21, 2008.

\bibitem[Ayoub et~al.(2020)Ayoub, Jia, Szepesvari, Wang, and
  Yang]{ayoub2020model}
A.~Ayoub, Z.~Jia, C.~Szepesvari, M.~Wang, and L.~Yang.
\newblock Model-based reinforcement learning with value-targeted regression.
\newblock In \emph{International Conference on Machine Learning}, pages
  463--474. PMLR, 2020.

\bibitem[Azar et~al.(2017)Azar, Osband, and Munos]{azar2017minimax}
M.~G. Azar, I.~Osband, and R.~Munos.
\newblock Minimax regret bounds for reinforcement learning.
\newblock In \emph{International Conference on Machine Learning}, pages
  263--272. PMLR, 2017.

\bibitem[Bai et~al.(2019)Bai, Xie, Jiang, and Wang]{bai2019provably}
Y.~Bai, T.~Xie, N.~Jiang, and Y.-X. Wang.
\newblock Provably efficient q-learning with low switching cost.
\newblock \emph{Advances in Neural Information Processing Systems}, 32, 2019.

\bibitem[Bellemare et~al.(2016)Bellemare, Srinivasan, Ostrovski, Schaul,
  Saxton, and Munos]{bellemare2016unifying}
M.~Bellemare, S.~Srinivasan, G.~Ostrovski, T.~Schaul, D.~Saxton, and R.~Munos.
\newblock Unifying count-based exploration and intrinsic motivation.
\newblock In \emph{Advances in Neural Information Processing Systems}, pages
  1471--1479, 2016.

\bibitem[Berner et~al.(2019)Berner, Brockman, Chan, Cheung, D{\k{e}}biak,
  Dennison, Farhi, Fischer, Hashme, Hesse, et~al.]{berner2019dota}
C.~Berner, G.~Brockman, B.~Chan, V.~Cheung, P.~D{\k{e}}biak, C.~Dennison,
  D.~Farhi, Q.~Fischer, S.~Hashme, C.~Hesse, et~al.
\newblock Dota 2 with large scale deep reinforcement learning.
\newblock \emph{arXiv preprint arXiv:1912.06680}, 2019.

\bibitem[Burda et~al.(2018)Burda, Edwards, Storkey, and
  Klimov]{burda2018exploration}
Y.~Burda, H.~Edwards, A.~Storkey, and O.~Klimov.
\newblock Exploration by random network distillation.
\newblock \emph{arXiv preprint arXiv:1810.12894}, 2018.

\bibitem[Cai et~al.(2020)Cai, Yang, Jin, and Wang]{cai2020provably}
Q.~Cai, Z.~Yang, C.~Jin, and Z.~Wang.
\newblock Provably efficient exploration in policy optimization.
\newblock In \emph{International Conference on Machine Learning}, pages
  1283--1294. PMLR, 2020.

\bibitem[Cheng et~al.(2021)Cheng, Kolobov, and Swaminathan]{cheng21heuristic}
C.~Cheng, A.~Kolobov, and A.~Swaminathan.
\newblock Heuristic-guided reinforcement learning.
\newblock In M.~Ranzato, A.~Beygelzimer, Y.~N. Dauphin, P.~Liang, and J.~W.
  Vaughan, editors, \emph{Advances in Neural Information Processing Systems 34:
  Annual Conference on Neural Information Processing Systems 2021, NeurIPS
  2021, December 6-14, 2021, virtual}, pages 13550--13563, 2021.
\newblock URL
  \url{https://proceedings.neurips.cc/paper/2021/hash/70d31b87bd021441e5e6bf23eb84a306-Abstract.html}.

\bibitem[Cutkosky et~al.(2021)Cutkosky, Dann, Das, Gentile, Pacchiano, and
  Purohit]{cutkosky2021dynamic}
A.~Cutkosky, C.~Dann, A.~Das, C.~Gentile, A.~Pacchiano, and M.~Purohit.
\newblock Dynamic balancing for model selection in bandits and rl.
\newblock In \emph{International Conference on Machine Learning}, pages
  2276--2285. PMLR, 2021.

\bibitem[Dann et~al.(2021)Dann, Marinov, Mohri, and Zimmert]{dann2021beyond}
C.~Dann, T.~V. Marinov, M.~Mohri, and J.~Zimmert.
\newblock Beyond value-function gaps: Improved instance-dependent regret bounds
  for episodic reinforcement learning.
\newblock \emph{Advances in Neural Information Processing Systems},
  34:\penalty0 1--12, 2021.

\bibitem[Efroni et~al.(2019)Efroni, Merlis, Ghavamzadeh, and
  Mannor]{efroni2019tight}
Y.~Efroni, N.~Merlis, M.~Ghavamzadeh, and S.~Mannor.
\newblock Tight regret bounds for model-based reinforcement learning with
  greedy policies.
\newblock \emph{Advances in Neural Information Processing Systems}, 32, 2019.

\bibitem[Faust et~al.(2018)Faust, Oslund, Ramirez, Francis, Tapia, Fiser, and
  Davidson]{faust18prm}
A.~Faust, K.~Oslund, O.~Ramirez, A.~G. Francis, L.~Tapia, M.~Fiser, and
  J.~Davidson.
\newblock {PRM-RL:} long-range robotic navigation tasks by combining
  reinforcement learning and sampling-based planning.
\newblock In \emph{2018 {IEEE} International Conference on Robotics and
  Automation, {ICRA} 2018, Brisbane, Australia, May 21-25, 2018}, pages
  5113--5120. {IEEE}, 2018.
\newblock \doi{10.1109/ICRA.2018.8461096}.
\newblock URL \url{https://doi.org/10.1109/ICRA.2018.8461096}.

\bibitem[Freedman(1975)]{freedman1975tail}
D.~A. Freedman.
\newblock On tail probabilities for martingales.
\newblock \emph{the Annals of Probability}, pages 100--118, 1975.

\bibitem[Fruit et~al.(2018)Fruit, Pirotta, Lazaric, and
  Ortner]{fruit2018efficient}
R.~Fruit, M.~Pirotta, A.~Lazaric, and R.~Ortner.
\newblock Efficient bias-span-constrained exploration-exploitation in
  reinforcement learning.
\newblock In \emph{International Conference on Machine Learning}, pages
  1578--1586. PMLR, 2018.

\bibitem[Golowich and Moitra(2022)]{golowich2022can}
N.~Golowich and A.~Moitra.
\newblock Can q-learning be improved with advice?
\newblock In \emph{Conference on Learning Theory}, pages 4548--4619. PMLR,
  2022.

\bibitem[Houthooft et~al.(2016)Houthooft, Chen, Duan, Schulman, De~Turck, and
  Abbeel]{houthooft2016vime}
R.~Houthooft, X.~Chen, Y.~Duan, J.~Schulman, F.~De~Turck, and P.~Abbeel.
\newblock Vime: Variational information maximizing exploration.
\newblock \emph{Advances in neural information processing systems}, 29, 2016.

\bibitem[Jaksch et~al.(2010)Jaksch, Ortner, and Auer]{jaksch2010near}
T.~Jaksch, R.~Ortner, and P.~Auer.
\newblock Near-optimal regret bounds for reinforcement learning.
\newblock \emph{Journal of Machine Learning Research}, 11:\penalty0 1563--1600,
  2010.

\bibitem[Jin et~al.(2018)Jin, Allen-Zhu, Bubeck, and Jordan]{jin2018q}
C.~Jin, Z.~Allen-Zhu, S.~Bubeck, and M.~I. Jordan.
\newblock Is q-learning provably efficient?
\newblock \emph{Advances in neural information processing systems}, 31, 2018.

\bibitem[Jin et~al.(2020)Jin, Yang, Wang, and Jordan]{jin2020provably}
C.~Jin, Z.~Yang, Z.~Wang, and M.~I. Jordan.
\newblock Provably efficient reinforcement learning with linear function
  approximation.
\newblock In \emph{Conference on Learning Theory}, pages 2137--2143. PMLR,
  2020.

\bibitem[Kakade et~al.(2003)]{kakade2003sample}
S.~M. Kakade et~al.
\newblock \emph{On the sample complexity of reinforcement learning}.
\newblock PhD thesis, University College London, 2003.

\bibitem[Li et~al.(2021{\natexlab{a}})Li, Shi, Chen, Gu, and
  Chi]{li2021breaking}
G.~Li, L.~Shi, Y.~Chen, Y.~Gu, and Y.~Chi.
\newblock Breaking the sample complexity barrier to regret-optimal model-free
  reinforcement learning.
\newblock \emph{Advances in Neural Information Processing Systems}, 34,
  2021{\natexlab{a}}.

\bibitem[Li et~al.(2021{\natexlab{b}})Li, Gupta, Reddy, Pong, Zhou, Yu, and
  Levine]{li2021mural}
K.~Li, A.~Gupta, A.~Reddy, V.~H. Pong, A.~Zhou, J.~Yu, and S.~Levine.
\newblock Mural: Meta-learning uncertainty-aware rewards for outcome-driven
  reinforcement learning.
\newblock In \emph{International Conference on Machine Learning}, pages
  6346--6356. PMLR, 2021{\natexlab{b}}.

\bibitem[Maurer and Pontil(2009)]{maurer2009empirical}
A.~Maurer and M.~Pontil.
\newblock Empirical bernstein bounds and sample variance penalization.
\newblock \emph{arXiv preprint arXiv:0907.3740}, 2009.

\bibitem[M{\'e}nard et~al.(2021)M{\'e}nard, Domingues, Shang, and
  Valko]{menard2021ucb}
P.~M{\'e}nard, O.~D. Domingues, X.~Shang, and M.~Valko.
\newblock Ucb momentum q-learning: Correcting the bias without forgetting.
\newblock In \emph{International Conference on Machine Learning}, pages
  7609--7618. PMLR, 2021.

\bibitem[Ng et~al.(1999)Ng, Harada, and Russell]{ng1999policy}
A.~Y. Ng, D.~Harada, and S.~Russell.
\newblock Policy invariance under reward transformations: Theory and
  application to reward shaping.
\newblock In \emph{ICML}, volume~99, pages 278--287, 1999.

\bibitem[Pacchiano et~al.(2020{\natexlab{a}})Pacchiano, Ball, Parker-Holder,
  Choromanski, and Roberts]{pacchiano2020optimism}
A.~Pacchiano, P.~Ball, J.~Parker-Holder, K.~Choromanski, and S.~Roberts.
\newblock On optimism in model-based reinforcement learning.
\newblock \emph{arXiv preprint arXiv:2006.11911}, 2020{\natexlab{a}}.

\bibitem[Pacchiano et~al.(2020{\natexlab{b}})Pacchiano, Dann, Gentile, and
  Bartlett]{pacchiano2020regret}
A.~Pacchiano, C.~Dann, C.~Gentile, and P.~Bartlett.
\newblock Regret bound balancing and elimination for model selection in bandits
  and rl.
\newblock \emph{arXiv preprint arXiv:2012.13045}, 2020{\natexlab{b}}.

\bibitem[Pacchiano et~al.(2020{\natexlab{c}})Pacchiano, Phan, Abbasi~Yadkori,
  Rao, Zimmert, Lattimore, and Szepesvari]{pacchiano2020model}
A.~Pacchiano, M.~Phan, Y.~Abbasi~Yadkori, A.~Rao, J.~Zimmert, T.~Lattimore, and
  C.~Szepesvari.
\newblock Model selection in contextual stochastic bandit problems.
\newblock \emph{Advances in Neural Information Processing Systems},
  33:\penalty0 10328--10337, 2020{\natexlab{c}}.

\bibitem[Pacchiano et~al.(2021)Pacchiano, Ball, Parker-Holder, Choromanski, and
  Roberts]{pacchiano2021towards}
A.~Pacchiano, P.~Ball, J.~Parker-Holder, K.~Choromanski, and S.~Roberts.
\newblock Towards tractable optimism in model-based reinforcement learning.
\newblock In \emph{Uncertainty in Artificial Intelligence}, pages 1413--1423.
  PMLR, 2021.

\bibitem[Pathak et~al.(2017)Pathak, Agrawal, Efros, and
  Darrell]{pathak2017curiosity}
D.~Pathak, P.~Agrawal, A.~A. Efros, and T.~Darrell.
\newblock Curiosity-driven exploration by self-supervised prediction.
\newblock In \emph{Proceedings of the IEEE Conference on Computer Vision and
  Pattern Recognition Workshops}, pages 16--17, 2017.

\bibitem[Raileanu and Rocktäschel(2020)]{Raileanu2020RIDE:}
R.~Raileanu and T.~Rocktäschel.
\newblock Ride: Rewarding impact-driven exploration for procedurally-generated
  environments.
\newblock In \emph{International Conference on Learning Representations}, 2020.
\newblock URL \url{https://openreview.net/forum?id=rkg-TJBFPB}.

\bibitem[Silver et~al.(2016)Silver, Huang, Maddison, Guez, Sifre, Van
  Den~Driessche, Schrittwieser, Antonoglou, Panneershelvam, Lanctot,
  et~al.]{silver2016mastering}
D.~Silver, A.~Huang, C.~J. Maddison, A.~Guez, L.~Sifre, G.~Van Den~Driessche,
  J.~Schrittwieser, I.~Antonoglou, V.~Panneershelvam, M.~Lanctot, et~al.
\newblock Mastering the game of go with deep neural networks and tree search.
\newblock \emph{nature}, 529\penalty0 (7587):\penalty0 484--489, 2016.

\bibitem[Silver et~al.(2017)Silver, Schrittwieser, Simonyan, Antonoglou, Huang,
  Guez, Hubert, Baker, Lai, Bolton, et~al.]{silver2017mastering}
D.~Silver, J.~Schrittwieser, K.~Simonyan, I.~Antonoglou, A.~Huang, A.~Guez,
  T.~Hubert, L.~Baker, M.~Lai, A.~Bolton, et~al.
\newblock Mastering the game of go without human knowledge.
\newblock \emph{nature}, 550\penalty0 (7676):\penalty0 354--359, 2017.

\bibitem[Stadie et~al.(2015)Stadie, Levine, and
  Abbeel]{stadie2015incentivizing}
B.~C. Stadie, S.~Levine, and P.~Abbeel.
\newblock Incentivizing exploration in reinforcement learning with deep
  predictive models.
\newblock \emph{arXiv preprint arXiv:1507.00814}, 2015.

\bibitem[Tang et~al.(2017)Tang, Houthooft, Foote, Stooke, Chen, Duan, Schulman,
  DeTurck, and Abbeel]{tang2017exploration}
H.~Tang, R.~Houthooft, D.~Foote, A.~Stooke, O.~X. Chen, Y.~Duan, J.~Schulman,
  F.~DeTurck, and P.~Abbeel.
\newblock \# exploration: A study of count-based exploration for deep
  reinforcement learning.
\newblock In \emph{Advances in neural information processing systems}, pages
  2753--2762, 2017.

\bibitem[Van~Seijen et~al.(2017)Van~Seijen, Fatemi, Romoff, Laroche, Barnes,
  and Tsang]{VanSeijen2017Hybrid}
H.~Van~Seijen, M.~Fatemi, J.~Romoff, R.~Laroche, T.~Barnes, and J.~Tsang.
\newblock Hybrid reward architecture for reinforcement learning.
\newblock In I.~Guyon, U.~V. Luxburg, S.~Bengio, H.~Wallach, R.~Fergus,
  S.~Vishwanathan, and R.~Garnett, editors, \emph{Advances in Neural
  Information Processing Systems}, volume~30. Curran Associates, Inc., 2017.
\newblock URL
  \url{https://proceedings.neurips.cc/paper/2017/file/1264a061d82a2edae1574b07249800d6-Paper.pdf}.

\bibitem[Vezhnevets et~al.(2017)Vezhnevets, Osindero, Schaul, Heess, Jaderberg,
  Silver, and Kavukcuoglu]{vezhnevets2017feudal}
A.~S. Vezhnevets, S.~Osindero, T.~Schaul, N.~Heess, M.~Jaderberg, D.~Silver,
  and K.~Kavukcuoglu.
\newblock Feudal networks for hierarchical reinforcement learning.
\newblock In \emph{Proceedings of the 34th International Conference on Machine
  Learning-Volume 70}, pages 3540--3549. JMLR. org, 2017.

\bibitem[Vinyals et~al.(2019)Vinyals, Babuschkin, Czarnecki, Mathieu, Dudzik,
  Chung, Choi, Powell, Ewalds, Georgiev, et~al.]{vinyals2019grandmaster}
O.~Vinyals, I.~Babuschkin, W.~M. Czarnecki, M.~Mathieu, A.~Dudzik, J.~Chung,
  D.~H. Choi, R.~Powell, T.~Ewalds, P.~Georgiev, et~al.
\newblock Grandmaster level in starcraft ii using multi-agent reinforcement
  learning.
\newblock \emph{Nature}, 575\penalty0 (7782):\penalty0 350--354, 2019.

\bibitem[Wang et~al.(2019)Wang, Wang, Du, and Krishnamurthy]{wang2019optimism}
Y.~Wang, R.~Wang, S.~S. Du, and A.~Krishnamurthy.
\newblock Optimism in reinforcement learning with generalized linear function
  approximation.
\newblock \emph{arXiv preprint arXiv:1912.04136}, 2019.

\bibitem[Yang et~al.(2021)Yang, Yang, and Du]{yang2021q}
K.~Yang, L.~Yang, and S.~Du.
\newblock Q-learning with logarithmic regret.
\newblock In \emph{International Conference on Artificial Intelligence and
  Statistics}, pages 1576--1584. PMLR, 2021.

\bibitem[Yang and Wang(2019)]{yang2019sample}
L.~Yang and M.~Wang.
\newblock Sample-optimal parametric q-learning using linearly additive
  features.
\newblock In \emph{International Conference on Machine Learning}, pages
  6995--7004. PMLR, 2019.

\bibitem[Yang and Wang(2020)]{yang2020reinforcement}
L.~Yang and M.~Wang.
\newblock Reinforcement learning in feature space: Matrix bandit, kernels, and
  regret bound.
\newblock In \emph{International Conference on Machine Learning}, pages
  10746--10756. PMLR, 2020.

\bibitem[Yang et~al.(2020)Yang, Jin, Wang, Wang, and Jordan]{yang2020function}
Z.~Yang, C.~Jin, Z.~Wang, M.~Wang, and M.~I. Jordan.
\newblock On function approximation in reinforcement learning: Optimism in the
  face of large state spaces.
\newblock \emph{arXiv preprint arXiv:2011.04622}, 2020.

\bibitem[Ye et~al.(2020)Ye, Chen, Zhang, Chen, Yuan, Liu, Chen, Liu, Qiu, Yu,
  Yin, Shi, Wang, Shi, Fu, Yang, Huang, and Liu]{ye2020towards}
D.~Ye, G.~Chen, W.~Zhang, S.~Chen, B.~Yuan, B.~Liu, J.~Chen, Z.~Liu, F.~Qiu,
  H.~Yu, Y.~Yin, B.~Shi, L.~Wang, T.~Shi, Q.~Fu, W.~Yang, L.~Huang, and W.~Liu.
\newblock Towards playing full moba games with deep reinforcement learning.
\newblock In H.~Larochelle, M.~Ranzato, R.~Hadsell, M.~F. Balcan, and H.~Lin,
  editors, \emph{Advances in Neural Information Processing Systems}, volume~33,
  pages 621--632. Curran Associates, Inc., 2020.
\newblock URL
  \url{https://proceedings.neurips.cc/paper/2020/file/06d5ae105ea1bea4d800bc96491876e9-Paper.pdf}.

\bibitem[Yu et~al.(2019)Yu, Quillen, He, Julian, Hausman, Finn, and
  Levine]{yu2019metaworld}
T.~Yu, D.~Quillen, Z.~He, R.~Julian, K.~Hausman, C.~Finn, and S.~Levine.
\newblock Meta-world: {A} benchmark and evaluation for multi-task and meta
  reinforcement learning.
\newblock In L.~P. Kaelbling, D.~Kragic, and K.~Sugiura, editors, \emph{3rd
  Annual Conference on Robot Learning, CoRL 2019, Osaka, Japan, October 30 -
  November 1, 2019, Proceedings}, volume 100 of \emph{Proceedings of Machine
  Learning Research}, pages 1094--1100. {PMLR}, 2019.
\newblock URL \url{http://proceedings.mlr.press/v100/yu20a.html}.

\bibitem[Zanette and Brunskill(2019)]{zanette2019tighter}
A.~Zanette and E.~Brunskill.
\newblock Tighter problem-dependent regret bounds in reinforcement learning
  without domain knowledge using value function bounds.
\newblock In \emph{International Conference on Machine Learning}, pages
  7304--7312. PMLR, 2019.

\bibitem[Zanette et~al.(2020)Zanette, Brandfonbrener, Brunskill, Pirotta, and
  Lazaric]{zanette2020frequentist}
A.~Zanette, D.~Brandfonbrener, E.~Brunskill, M.~Pirotta, and A.~Lazaric.
\newblock Frequentist regret bounds for randomized least-squares value
  iteration.
\newblock In \emph{International Conference on Artificial Intelligence and
  Statistics}, pages 1954--1964. PMLR, 2020.

\bibitem[Zhai et~al.(2022)Zhai, Baek, Zhou, Jiao, and
  Ma]{zhai2022computational}
Y.~Zhai, C.~Baek, Z.~Zhou, J.~Jiao, and Y.~Ma.
\newblock Computational benefits of intermediate rewards for goal-reaching
  policy learning.
\newblock \emph{Journal of Artificial Intelligence Research}, 73:\penalty0
  847--896, 2022.

\bibitem[Zhang et~al.(2020)Zhang, Zhou, and Ji]{zhang2020almost}
Z.~Zhang, Y.~Zhou, and X.~Ji.
\newblock Almost optimal model-free reinforcement learningvia
  reference-advantage decomposition.
\newblock \emph{Advances in Neural Information Processing Systems},
  33:\penalty0 15198--15207, 2020.

\bibitem[Zhou et~al.(2021)Zhou, He, and Gu]{zhou2021provably}
D.~Zhou, J.~He, and Q.~Gu.
\newblock Provably efficient reinforcement learning for discounted mdps with
  feature mapping.
\newblock In \emph{International Conference on Machine Learning}, pages
  12793--12802. PMLR, 2021.

\end{thebibliography}

\appendix

\newpage
\section{Full Proofs}
\label{sec:supporting_lemmas}

Let $f_h : \Scal \rightarrow \mathbb{R}$ be an arbitrary family of horizon indexed functions satisfying $\| f_h \|_\infty \leq B$ for some $B > 0$. We assume $f_h$ is not a random variable as a function of $\Scal$ . The following upper bounds hold,

\begin{lemma}\label{lemma::fixed_f_bound}
Fix $\delta \in (0,1)$, $\forall t \in \mathbb{N}$, $s \in \Scal$, $a \in \Acal$ and $h \in [H]$, with probability at least $1-\delta$, we have
\begin{align*}
       \abs{\Phat_t(\cdot|s,a)^\top f_{h+1} - \Pstar(\cdot|s,a)^\top f_{h+1}} 
      &\leq 16 \sqrt{ \frac{ \mathrm{Var}(f_{h+1}(s') | s,a) \ln \frac{2|\Scal||\Acal|t}{\delta} }{ N_h^t(s,a)} } + \frac{12B\frac{2|\Scal||\Acal|t}{\delta}}{N_h^t(s,a) } \\
      &\leq 16B \sqrt{ \frac{ \ln \frac{2|\Scal||\Acal|t}{\delta} }{ N_h^t(s,a)} } + \frac{12B\frac{2|\Scal||\Acal|t}{\delta}}{N_h^t(s,a) } 
\end{align*}
 for all $t \in \mathbb{N}$. Similarly with probability at least $1-\delta$, we have
\begin{equation*}
       \abs{\Phat_t(\cdot|s,a)^\top f_{h+1} - \Pstar(\cdot|s,a)^\top f_{h+1}} \leq 16 \sqrt{ \frac{ \widehat{\mathrm{Var}}(f_{h+1}(s') | s,a) \ln \frac{2|\Scal||\Acal|t}{\delta} }{ N_h^t(s,a)} } + \frac{12B\frac{2|\Scal||\Acal|t}{\delta}}{N_h^t(s,a) }
\end{equation*}
 for all $t \in \mathbb{N}$. Where $\widehat{\mathrm{Var}}(f_{h+1}(s') | s,a)  = \frac{1}{N_h^k(s,a)( N_h^k(s,a)-1)} \sum_{1 \leq i < j < N_h^k(s,a)} (f_{h+1}(s_{h+1}^i) - f_{h+1}(s_{h+1}^j))^2$.
\end{lemma}

\begin{proof}
Consider a fixed tuple $s,a,t, h \in \Scal \times \Acal \times \mathbb{N} \times [H]$. By the definition of $\Phat_t$, we have
\begin{equation*}
     \Phat_t(\cdot|s,a)^\top f_{h+1} = \frac{1}{N_h^t(s,a)}\sum_{i=1}^{t-1} 
     \indicator{(s_{h'}^i,a_{h'}^i) = (s,a)}f_{h+1}(s_{h'+1}^i).
\end{equation*}

Now denote $\mathcal{H}_{h,i}$ as the entire history from $t=0$ to iteration $t=i$ where in iteration $i$, the history $\mathcal{H}_{h,i}$ includes all interactions from step $0$ up to and including time step $h$. Next, $\forall i = 0,1,\dots,t-1$, define the random variables:

\begin{equation*}
    X_{i, h'}(s,a) = \indicator{(s^i_{h'},a^i_{h'})=(s,a)}f_{h+1}(s^i_{h'+1}) - \bb E\brac{\indicator{(s^i_{h'},a^i_{h'})=(s,a)}f_{h+1}(s^i_{h'+1})|\mc H_{h,i}}.
\end{equation*}

Notice that $\abs{X_{i, h'}}\leq B$, if $\indicator{(s_{h'}^i,a_{h'}^i)=(s,a)}=1$, else $\abs{X_{i,h'}} = 0$ so that 

\begin{equation*}
    \mathrm{Var}(  X_{t, h}(s,a)  | H_{h-1,t}) \leq \begin{cases}
    0 &\text{if } \abs{X_{i,h'}} = 0 \\
    B^2 &\text{o.w.}
    \end{cases}
\end{equation*}

An anytime Bernstein bound (see Lemma~\ref{lemma:simplified_freedman}) implies,

\begin{align*}
  \abs{\sum_{i=1}^{t-1} 
      X_{i, h'}(s,a) } &\leq 16 \sqrt{ N_h^t(s,a) \mathrm{Var}( f_{h+1}(s')  | s,a)  \ln \frac{2|\Scal||\Acal|t}{\delta}  } + 12B\frac{2|\Scal||\Acal|t}{\delta} \\
      &\leq 16 B\sqrt{ N_h^t(s,a) \ln \frac{2|\Scal||\Acal|t}{\delta}  } + 12B\frac{2|\Scal||\Acal|t}{\delta}.
\end{align*}
With probability at least $1-\delta$ for all $t \in \mathbb{N}$. Thus, we have

\begin{equation*}
   \abs{\Phat_t(\cdot|s,a)^\top f_{h+1} - \Pstar(\cdot|s,a)^\top f_{h+1}} \leq 16B \sqrt{ \frac{ \ln \frac{2|\Scal||\Acal|t}{\delta} }{ N_h^t(s,a)} } + \frac{12B\frac{2|\Scal||\Acal|t}{\delta}}{N_h^t(s,a) }
\end{equation*}
with probability at least $1-\delta$ for all $t \in \mathbb{N}$. The second inequality can be proven following the exact same proof process as described above, but instead making use of the empirical Bernstein bound of~\ref{lemma::bernstein_anytime} instead.

\end{proof}

The following bound relates the error of $\Phat_t$ and $\Pstar$. 

\begin{corollary}[State-action wise model error under $\Vstar$ (Lemma 7.3 of~\cite{agarwal2019reinforcement})]
\label{corollary:ModelErrorEmpiricalBernsteinV}
Fix $\delta\in (0,1)$, $\forall t\in [1,2,\dots T],s\in \Scal,a\in \Acal,h\in [H]$, consider $\forall \Vstar_h:\Scal \rightarrow [0,H]$. With probability at least $1-\delta$, we have

\begin{align*}
    \abs{\Phat_t(\cdot|s,a)^\top \Vstar_{h+1} - \Pstar(\cdot|s,a)^\top \Vstar_{h+1}} \leq  \qquad \qquad \qquad \qquad \qquad \qquad \qquad \qquad \qquad \qquad \qquad  \qquad \qquad\\
    \qquad   \qquad \qquad\min\left(  16\sqrt{ \frac{ \widehat{\mathrm{Var}}_{s' \sim \Phat_t(\cdot | s,a) }(\Vstar_{h+1}(s') | s,a)  \ln\frac{ 2 |\Scal| |\Acal| t}{\delta} }{ N_h^t(s,a) }} + \frac{12\Vstarmax }{N_h^t(s,a)}\ln \frac{ 2 |\Scal| |\Acal|t}{\delta}, 2\Vstarmax\right),
\end{align*}
for all $t \in \mathbb{N}$. Where $\widehat{\mathrm{Var}}(\Vstar_{h+1}(s') | s,a)  = \frac{1}{N_h^k(s,a)( N_h^k(s,a)-1)} \sum_{1 \leq i < j < N_h^k(s,a)} (\Vstar_{h+1}(s_{h+1}^i) - \Vstar_{h+1}(s_{h+1}^j))^2$ and $\Vstarmax = \max_{s } \Vstar(s)$. 


\end{corollary}

\begin{proof}
A direct application of Lemma~\ref{lemma::fixed_f_bound} using $B = \Vstarmax$ yields the desired result. 
\end{proof}

A direct consequence of Lemma~\ref{corollary:ModelErrorEmpiricalBernsteinV} and Assumption~\ref{assump::sandwich} is that, 

\begin{corollary}\label{corollary::variance_bonus_domination}
The $\Vstar$ ``variance bonus'' is upper bounded by the empirical $\Vtilde$ ``second moment bonus'':

\begin{equation*}
    16\sqrt{ \frac{ \widehat{\mathrm{Var}}_{s' \sim \Phat_t(\cdot | s,a) }(\Vstar_{h+1}(s') | s,a)  \ln\frac{ 2 |\Scal||\Acal|t}{\delta} }{ N_h^t(s,a) }}  \leq 16 \beta \cdot \sqrt{ \frac{ \widehat{\mathbb{E}}_{s' \sim \Phat_t(\cdot | s,a) }[\Vtilde^2_{h+1}(s') | s,a ]  \ln\frac{ 2 |\Scal||\Acal|}{\delta} }{ N_h^t(s,a) }} 
\end{equation*}
and 

\begin{equation*}
    \frac{12\Vstarmax }{N_h^t(s,a)}\ln \frac{ 2 |\Scal| |\Acal|t}{\delta} \leq \frac{12\beta \Vtildemax }{N_h^t(s,a)}\ln \frac{ 2 |\Scal| |\Acal|t}{\delta}.
\end{equation*}
\end{corollary}

\begin{proof}
For any random variable $X \sim \mathbb{P}$, 
\begin{equation*}
    \mathrm{Var}(X) \leq \mathbb{E}_{X \sim \mathbb{P}}\left[ X^2 \right]. 
\end{equation*}
Therefore, for all $s,a$ and any empirical distribution $\Phat_t(\cdot | s,a)$,

\begin{equation*}
    \widehat{\mathrm{Var}}_{s' \sim \Phat_t(\cdot | s,a) }(\Vstar_{h+1}(s') | s,a) \leq \widehat{\mathbb{E}}_{s' \sim \Phat_t(\cdot | s,a) }\left[\left(\Vstar_{h+1}(s') \right)^2| s,a\right].
\end{equation*}

Finally, by Assumption~\ref{assump::sandwich},

\begin{equation*}
    \widehat{\mathbb{E}}_{s' \sim \Phat_t(\cdot | s,a) }\left[\left(\Vstar_{h+1}(s') \right)^2| s,a\right] \leq \beta^2 \widehat{\mathbb{E}}_{s' \sim \Phat_t(\cdot | s,a) }\left[\Vtilde^2_{h+1}(s')| s,a\right].
\end{equation*}
The result follows.
\end{proof}

Corollary~\ref{corollary::variance_bonus_domination} and Lemma~\ref{lemma::fixed_f_bound} implies that with probability $1-\delta$ for all $t \in \mathbb{N}, s \in \Scal, a\in \Acal$ and $h \in [1, \cdots, H]$ the bonuses $b_h^t$ from Equation~\ref{eq::shaped_bonus} satisfy,

\begin{align}
   \left| \left(\Phat_t(\cdot |s,a)  - \Pstar(\cdot | s,a) \right)\cdot \Vstar_{h+1}  \right| &\stackrel{(i)}{\leq} \min\left( 16 \sqrt{ \frac{ \widehat{\mathrm{Var}}(\Vstar_{h+1}(s') | s,a) \ln \frac{2|\Scal||\Acal|t}{\delta} }{ N_h^t(s,a)} } + \frac{12\Vstarmax \ln \frac{2|\Scal||\Acal|t}{\delta}}{N_h^t(s,a) }, 2 \Vstarmax\right)\notag\\
   &\stackrel{(ii)}{\leq} b_h^t(s,a).\label{equation::bonus_justification}
\end{align}

Where inequality $(i)$ follows by Corollary~\ref{corollary:ModelErrorEmpiricalBernsteinV} and $(ii)$ by Corollary~\ref{corollary::variance_bonus_domination}.

Define as $\Vhat_{h}^t$ to the stage $h$ value function of $\pi_t$ as computed in the learned model and using bonus augmented rewards (with bonuses defined as in~\ref{eq::shaped_bonus}) with the data collected up to round $t-1$. We start by showing $\Vhat_h^t$ is always larger than $\Vstar_h(s)$.

We show the following supporting Lemma, a generalization of Lemma~\ref{lemma::fixed_f_bound} where the functions $f_h: \Scal \rightarrow \mathbb{R} $ satisfying $\| f_h\|_\infty$ are allowed to be random. This lemma is a slight refinement from its corresponding result in the literature, where the dependence on the size of the support of $| \Pstar(\cdot | s,a) | $ is upper bounded by $|\Scal$. Having a more refined control on the size of the support of $\Pstar(\cdot | s,a)$ is what allows our final bounds to only depend on the size of the effective state space. 

\begin{lemma}
\label{lemma:ModelErrorEmpiricalBernsteinGeneralized}
Fix $\delta\in (0,1)$, $\forall t\in \mathbb{N},s\in \Scal,a\in \Acal,h\in [H]$,  with probability at least $1-\delta$, we have

\begin{align*}
      &\abs{\Phat_t(\cdot|s,a)^\top f_{h+1} - \Pstar(\cdot|s,a)^\top f_{h+1}}\\
      \leq &\;16 \beta \cdot \sqrt{ \frac{| \mathrm{support}( \Pstar(\cdot |s,a) )| \widehat{\bb E}_{s' \sim \Phat_t(\cdot | s,a) }[\Vtilde^2_{h+1}(s') | s,a]  \ln\frac{ 8B |\Scal| |\Acal| t^3}{\delta} }{ N_h^t(s,a) }} \\
      &+\quad\frac{12B| \mathrm{support}( \Pstar(\cdot |s,a) )|}{N_h^t(s,a)}\ln \frac{ 8B |\Scal| |\Acal|t^3}{ \delta} + \frac{1}{t^2}
\end{align*}
simultaneously for all $\{f_h: \Scal \rightarrow \mathbb{R}\}_{h = 0}^{H-1}$ such that $f_h(s) \in \left[ \Vtilde_h, \beta \Vtilde_h\right]$ and $\| f_h\|_\infty \leq B$. Where $\mathrm{support}( \Pstar(\cdot |s,a) ) $ corresponds to the size of the support of distribution $\Pstar(\cdot |s,a)$. 

\end{lemma}

\begin{proof}

The same proof template as in Lemma~\ref{lemma::fixed_f_bound} plus a union bound over a covering of the space of $f$ functions yields the desired result. Note that for all $s, a \in \Scal \times \Acal$,
\begin{equation*}
    \widehat{\mathrm{Var}}(f_{h+1}(s') | s,a) \leq \widehat{\mathbb{E}}_{s' \sim \Phat_t(\cdot | s,a) }[f^2_{h+1}(s') | s,a] \leq \beta^2 \widehat{\mathbb{E}}_{s' \sim \Phat_t(\cdot | s,a) }[\Vtilde^2_{h+1}(s') | s,a]
\end{equation*}

Therefore Lemma~\ref{lemma::fixed_f_bound} implies that for any fixed $f_{h+1}$, with probability at least $1-\delta'$

\begin{align}
    \label{eq::single-point-convergence}    
        &\abs{\Phat_t(\cdot|s,a)^\top f_{h+1} - \Pstar(\cdot|s,a)^\top f_{h+1}}\\
        \leq&\; 16 \beta \cdot \sqrt{ \frac{ \widehat{\mathrm{Var}}_{s' \sim \Phat_t(\cdot | s,a) }(f_{h+1}(s') | s,a)  \ln\frac{ 2 |\Scal| |\Acal| t}{\delta'} }{ N_h^t(s,a) }} + \frac{12B}{N_h^t(s,a)}\ln \frac{ 2 |\Scal| |\Acal|t}{\delta'} \notag\\
        \leq &\;16 \beta \cdot \sqrt{ \frac{ \widehat{\mathbb{E}}_{s' \sim \Phat_t(\cdot | s,a) }[\tilde{V}^2_{h+1}(s') | s,a]  \ln\frac{ 2 |\Scal| |\Acal| t}{\delta'} }{ N_h^t(s,a) }} +  \frac{12B}{N_h^t(s,a)}\ln \frac{ 2 |\Scal| |\Acal|t}{\delta'} 
\end{align}

for all (fixed) $f \in \{f_h: \Scal \rightarrow \mathbb{R}\}_{h = 0}^{H-1}$ and for all $s, a, h \in \Scal \times \Acal \times H$ simultaneously.  We now apply a standard $\epsilon-$net covering argument to the inequality we have proven above. Notice that regardless of the number of samples, $\mathrm{support}( \Phat_t(\cdot |s,a) ) \subseteq \mathrm{support}( \Pstar(\cdot |s,a) )$. Thus, the only ``entries'' that matter in $f_{h+1}$ are those corresponding to states in $\mathrm{support}( \Pstar(\cdot |s,a) )$. 

Let's consider an $\epsilon-$net of $f_{h+1}$ restricted to $\mathrm{support}( \Pstar(\cdot |s,a) )$. Since for any $s \in\Scal$, we assume $f_{h+1} \in [\Vtilde_{h+1}, \beta \Vtilde_{h+1}]$, there exists an epsilon net $\mathcal{N}_\epsilon$ under the infinity norm satisfying $|\mathcal{N}_\epsilon| \leq (\frac{2B}{\epsilon})^{| \mathrm{support}( \Pstar(\cdot |s,a) )|}$. For any $f_{h+1}$ we denote by $f_{h+1}^\epsilon$ its closest element in $\mathcal{N}_\epsilon$ (in the infinity norm over $\mathrm{support}( \Pstar(\cdot |s,a) )$). The following holds, 

\begin{align*}
     \left|\abs{\Phat_t(\cdot|s,a)^\top f_{h+1} - \Pstar(\cdot|s,a)^\top f_{h+1}}  -  \abs{\Phat_t(\cdot|s,a)^\top f^\epsilon_{h+1} - \Pstar(\cdot|s,a)^\top f^\epsilon_{h+1}} \right| \leq  \\
     \abs{\Phat_t(\cdot|s,a)^\top f_{h+1}- \Phat_t(\cdot|s,a)^\top f_{h+1}} + \abs{\Pstar(\cdot|s,a)^\top f^\epsilon_{h+1} - \Pstar(\cdot|s,a)^\top f^\epsilon_{h+1}} \leq      2\epsilon.
\end{align*}

And therefore, setting $\delta' = \frac{\delta}{(\frac{2B}{\epsilon})^{| \mathrm{support}( \Pstar(\cdot |s,a) )|}} \leq \frac{\delta}{|\mathcal{N}_\epsilon|} $, a union bound over all elements of $\mathcal{N}_\epsilon$ implies that with probability at least $1-\delta$, we have

\begin{align*}
      &\abs{\Phat_t(\cdot|s,a)^\top f_{h+1} - \Pstar(\cdot|s,a)^\top f_{h+1}}\\
      \leq &\;16 \beta \cdot \sqrt{ \frac{| \mathrm{support}( \Pstar(\cdot |s,a) )| \widehat{\bb E}_{s' \sim \Phat_t(\cdot | s,a) }[\tilde{V}^2_{h+1}(s') | s,a]  \ln\frac{ 4B |\Scal| |\Acal| t}{\epsilon\delta} }{ N_h^t(s,a) }}\\
      &+\frac{12B| \mathrm{support}( \Pstar(\cdot |s,a) )|}{N_h^t(s,a)}\ln \frac{ 4B |\Scal| |\Acal|t}{\epsilon\delta} + 2\epsilon
\end{align*}

for all $f_{h+1}$ simultaneously. Setting $\epsilon = \frac{1}{2t^2}$ we get that with probability at least $1-\delta$,

\begin{align*}
      &\abs{\Phat_t(\cdot|s,a)^\top f_{h+1} - \Pstar(\cdot|s,a)^\top f_{h+1}}\\
      \leq &\; 16 \beta \cdot \sqrt{ \frac{| \mathrm{support}( \Pstar(\cdot |s,a) )| \widehat{\bb E}_{s' \sim \Phat_t(\cdot | s,a) }[\tilde{V}^2_{h+1}(s') | s,a]  \ln\frac{ 8B |\Scal| |\Acal| t^3}{\delta} }{ N_h^t(s,a) }} \\
      &+\frac{12B| \mathrm{support}( \Pstar(\cdot |s,a) )|}{N_h^t(s,a)}\ln \frac{ 8B |\Scal| |\Acal|t^3}{ \delta} + \frac{1}{t^2}
\end{align*}

for all $f_{h+1}$ simultaneously and for all $t \in \mathbb{N}$. This completes the result.

\end{proof}

\subsection{Proof of Lemma~\ref{lemma::optimism}}\label{appendix::lemma_optimism_proof}
In this section we show that optimism holds for all state-action pairs. We restate Lemma~\ref{lemma::optimism} for the reader's convenience.

\lemmaoptimism*

\begin{proof}
We prove via induction. At the additional time step $H$ we have $\Vhat_H^t(s) = \Vstar_H(s) = 0$ for all $s$. 

Starting at $h+1$, and assuming we have $\Vhat_{h+1}^t (s) \geq \Vstar_{h+1}(s)$ for all $s$, we move to $h$ below.

Consider any $s, a \in \Scal \times \Acal$. First if $\Qhat_h^t(s,a) = H$ then we have $\Qhat_h^t(s,a) \geq \Qstar_h(s,a)$. The following inequalities hold,

\begin{align*}
    \Qhat_h^t(s,a) - \Qstar_h(s,a) &= b_h^t(s,a) + \Phat_t(\cdot |s,a) \cdot \Vhat_{h+1}^t - \Pstar(\cdot | s,a) \cdot \Vstar_{h+1}\\
    &\stackrel{(i)}{\geq} b_h^t(s,a) + \Phat_t(\cdot |s,a) \cdot \Vstar_{h+1} - \Pstar(\cdot | s,a) \cdot \Vstar_{h+1}\\
    &=  b_h^t(s,a) + \left(\Phat_t(\cdot |s,a)  - \Pstar(\cdot | s,a) \right)\cdot \Vstar_{h+1} \\
    &\stackrel{(ii)}{\geq} b_h^t(s,a) - 16\sqrt{ \frac{ \widehat{\mathbb{E}}_{s' \sim \Phat_t(\cdot | s,a) }[\Vtilde^2_{h+1}(s') | s,a ]  \ln\frac{ 2 |\Scal| |\Acal| t}{\delta} }{ N_h^t(s,a) }} - \frac{12\Vtildemax }{N_h^t(s,a)}\ln \frac{ 2 |\Scal||\Acal|t}{\delta} \\
    &\stackrel{(iii)}{\geq} 0
\end{align*}

Where $(i)$ holds because of the inductive optimism assumption. Inequality $(ii)$ holds because by Corollary~\ref{corollary:ModelErrorEmpiricalBernsteinV} we have,

\begin{equation*}
    \left(\Phat_t(\cdot |s,a)  - \Pstar(\cdot | s,a) \right)\cdot \Vstar_{h+1}  \leq 16 \sqrt{ \frac{ \widehat{\mathrm{Var}}(\Vstar_{h+1}(s') | s,a) \ln \frac{2|\Scal||\Acal|t}{\delta} }{ N_h^t(s,a)} } + \frac{12\Vstarmax\frac{2|\Scal||\Acal|t}{\delta}}{N_h^t(s,a) }
\end{equation*}

and $(iii)$ by Corollary~\ref{corollary::variance_bonus_domination} applied to the definition of $b_h^t$.

\end{proof}


\section{Supporting Results for Section~\ref{section::prunning_state_space}}

A consequence of Lemma~\ref{lemma::fixed_f_bound}, and Corollary~\ref{corollary::variance_bonus_domination},

\begin{corollary}\label{corollary::empirical_model_vs_v_tilde_bond}

Fix $\delta\in (0,1)$, $\forall t\in \mathbb{N} ,s\in \Scal,a\in \Acal,h\in [H]$, consider $\forall~ \Vtilde_h:\Scal \mapsto [0,H]$ with probability at least $1-\delta$, we have

\begin{equation*}
    \abs{\Phat_t(\cdot|s,a)^\top \Vtilde_{h+1} - \Pstar(\cdot|s,a)^\top \Vtilde_{h+1}} \leq 16\sqrt{ \frac{ \widehat{\mathbb{E}}_{s' \sim \Phat_t(\cdot | s,a) }\left[\Vtilde^2_{h+1}(s') | s,a\right]  \ln\frac{ 2 |\Scal| |\Acal| t}{\delta} }{ N_h^t(s,a) }} + \frac{12 \Vtildemax}{N_h^t(s,a)}\ln \frac{ 2 |\Scal| |\Acal|t}{\delta}.
\end{equation*}

\end{corollary}

We will now show the empirical $\Qtilde$ values can be approximately upper bounded by the $\Qtilde^u$ values. Let's define the ``tilde bonus'' as $\tilde{b}_h^t(s,a) = 16\beta\sqrt{ \frac{ \widehat{\mathbb{E}}_{s' \sim \Phat_t(\cdot | s,a) }\left[\Vtilde^2_{h+1}(s') | s,a\right]  \ln\frac{ 2 |\Scal| |\Acal| t}{\delta} }{ N_h^t(s,a) }} + \frac{12\beta \Vtildemax}{N_h^t(s,a)}\ln \frac{ 2 |\Scal| |\Acal|t}{\delta}$.

\begin{corollary}\label{corollary::upper_bound_empirical_Q_with_Qtildeu}
With probability at least $1-2\delta$ the empirical $Q$ function of state action pair $(s,a)$ satisfies,
\begin{equation*}
      \Qhat_h^t(s,a)  \leq r(s,a) + b_h^t(s,a) + \tilde{b}_h^t(s,a) + \beta \Pstar(\cdot|s,a)^\top \Vtilde_{h+1} := \Qtilde^u(s,a) +  b_h^t(s,a) +  \tilde{b}_h^t(s,a).
\end{equation*}
for all $s,a \in \Scal \times \Acal$ and for all $h \in [0, \cdots, H-1]$ and all $t \in \mathbb{N}$. 
\end{corollary}

\begin{proof}
By definition $\widehat Q_h^t(s,a) = r(s,a) + b_h^t(s,a) + \Phat^t(\cdot | s,a) \Vhat^t_{h+1}$, since $\Vhat^t_{h+1}$ is clipped above by $\beta \Vtilde_{h+1}$,
\begin{align*}
\widehat Q_h^t(s,a) &= r(s,a) + b_h^t(s,a) + \Phat^t(\cdot | s,a) \Vhat^t_{h+1} &\\
&\leq r(s,a) + b_h^t(s,a) + \beta \Phat^t(\cdot | s,a) \Vtilde_{h+1}&\\
&\leq r(s,a) + b_h^t(s,a) +  \tilde b^t_h(s,a) + \beta \Pstar(\cdot | s,a)^\top  \Vtilde_{h+1}\\
&= \Qtilde^u(s,a) +  b_h^t(s,a) +  \tilde{b}_h^t(s,a).
\end{align*}
The last inequality is a consequence of Corollary~\ref{corollary::empirical_model_vs_v_tilde_bond}. The result follows.

\end{proof}

Corollary~\ref{corollary::upper_bound_empirical_Q_with_Qtildeu} implies that once $b_h^t(s,a) + \tilde{b}_h^t(s,a) <  \Vstar(s) - \Qtilde^u(s,a)$ optimism guarantees that from that point on, action $a$ will never again be taken at state $s$. Indeed, since optimism (see Lemma~\ref{lemma::optimism}) $\widehat{Q}_h^t(s,\pi_\star(s))$ is at least $\Vstar(s)$, action $a$ will be dominated by action $\pi_\star(s)$ from that point onward. We make this intuition precise in the following Lemma by upper bounding the number of times action $a$ is taken at state $s$ when $(s,a) \in \pseudosub_\Delta$. 


\begin{lemma}\label{lemma::bounding_visitation_pseudosub}
With probability at least $1-2\delta$ and for all $\Delta \in [0,1]$ the state action pairs $(s,a) \in \pseudosub_\Delta $ satisfy the bound,
\begin{equation*}
    N_{h(s)}^t(s,a) \leq \frac{8192 \beta^2 \times \left( \Vtildemax \right)^2 \cdot \ln \frac{2 |\Scal||\Acal| t}{\delta} }{\Delta^2}
 \end{equation*}
For all $t \in \mathbb{N}$. Where $h(s)$ corresponds to horizon index of the state partitions that contains\footnote{Recall that by Assumption~\ref{assump::h_indexing} the states are assumed to be $h-$indexed and therefore the state space can be written as $\Scal = \cup_{h \in [H]} \Scal_h$. This also implies that $N_h^t(s) = 0$ for all $h \neq h(s)$.  } state $s$. 
\end{lemma}

\begin{proof}

We start by assuming the events of Lemma~\ref{lemma::optimism} (optimism) and Corollary~\ref{corollary::upper_bound_empirical_Q_with_Qtildeu} ($\Qhat_h^t(s,a) \leq \Qtilde^u(s,a) + b_h^t(s,a) + \tilde b_h^t(s,a)$) hold. We do not need any $\Delta-$dependent high probability event to hold for our results to be valid.

Once $b_{h(s)}^t(s,a) + \tilde{b}_{h(s)}^t(s,a) <  \Vtilde(s) - \Qtilde^u(s,a)$, action $a$ will never be taken again at state $s$. The following bound holds for $b_{h(s)}^t(s,a) + \tilde{b}_{h(s)}^t(s,a)$,

\begin{align*}
    b_{h(s)}^t(s,a) + \tilde{b}_{h(s)}^t(s,a) &\leq 32 \beta \sqrt{ \frac{ \widehat{\mathbb{E}}_{s' \sim \Phat_t(\cdot | s,a) }[\Vtilde^2_{h(s)+1}(s') | s,a ]  \ln\frac{ 2 |\Scal||\Acal|t}{\delta} }{ N_{h(s)}^t(s,a) }}  +  \frac{24 \beta\max_{s'} \Vtilde_{h(s)+1}(s')}{N_{h(s)}^t(s,a)}\ln \frac{ 2 |\Scal||\Acal|t}{\delta} \\
    &\leq 32 \beta \max_{s} \Vtilde_{h(s)+1}(s) \sqrt{\frac{\ln\frac{ 2 |\Scal||\Acal|t}{\delta}}{N_{h(s)}^t(s,a)} } + \frac{24\beta\max_{s'} \Vtilde_{h(s)+1}(s')}{N_{h(s)}^t(s,a)}\ln \frac{ 2 |\Scal||\Acal|t}{\delta}.
\end{align*}

Assume $(s,a) \in \pseudosub_\Delta$. When 
\begin{equation*}
    N_{h(s)}^t(s,a) > \frac{8192 \beta^2 \times \max_{s', h'} \Vtilde^2_{h'}(s') \cdot \ln \frac{2 |\Scal||\Acal| t}{\delta} }{\Delta^2}
\end{equation*}
we get
\begin{equation}\label{equation::bounding_bonus_terms}
    b_{h(s)}^t(s,a) + \tilde{b}_{h(s)}^t(s,a) < \Delta \leq \Vstar(s) - \Qtilde^u(s,a) .
\end{equation}

The result follows because as soon as $b_{h(s)}^t(s,a) + \tilde{b}_{h(s)}^t(s,a) < \Delta$ holds, the optimism guarantees that from that point on action $a$ will never again be taken at state $s$. Indeed, let $\pi^\star(s)$ be the optimal action at state $s$,
\begin{equation*}
    \Qhat_{h(s)}^t(s,\pi^\star(s) ) \geq \Qstar(s,\pi^\star(s)) = \Vstar(s) \geq \Qtilde^u(s,a) + \Delta 
\end{equation*}
Where the last inequality holds because by definition $(s,a) \in \pseudosub_\Delta$.

Plugging in the bound of Corollary~\ref{corollary::upper_bound_empirical_Q_with_Qtildeu}, and invoking inequality~\ref{equation::bounding_bonus_terms},

\begin{equation*}
     \Qhat_{h(s)}^t(s,\pi^\star(s) ) \geq   \Qtilde^u(s,a) + \Delta  > \Qtilde^u(s,a) +  b_{h(s)}^t(s,a) + \tilde{b}_{h(s)}^t(s,a) \geq    \Qhat_{h(s)}^t(s,a) .
\end{equation*}

implying action $a$ will not be selected by the greedy policy induced by the empirical $Q$ function $\Qhat_{h(s)}^t$. This finalizes the result. 
\end{proof}



\begin{restatable}[Maximum Visitation of PseudoSuboptimal Pairs]{lemma}{maxvisitationpseudosub}\label{lemma::upper_bounding_sum_visitation_pseudosub-appendix}

With probability at least $1-2\delta$, and for all $\Delta \in [0,1]$ and $T \in \mathbb{N}$ the number of episodes whose sample trajectories contain a state from $\pathpseudosub_\Delta$ is upper bounded by,
\begin{align*}
   &\sum_{t=1}^T \mathbf{1}( \tau_t \cap \left(\pathpseudosub_\Delta \cup \pseudosub_\Delta\right) \neq \emptyset)\\ \leq&\;
   |\boundarypseudosub_\Delta| \times \frac{8192 \beta^2 \times \left(\Vtildemax\right)^2 \cdot \ln \frac{2 |\Scal||\Acal| T}{\delta} }{\Delta^2} 
\end{align*}
and
\begin{align*}
    &\sum_{(s,a) \in  \left (\left\{\pathpseudosub_\Delta \times \Acal \right \}\cup \pseudosub_\Delta\right)} N_{h(s)}^T(s,a)\\
    \leq&\; H \times |\boundarypseudosub_\Delta |\times \frac{8192 \beta^2 \times \left(\Vtildemax\right)^2 \cdot \ln \frac{2 |\Scal||\Acal| T}{\delta} }{\Delta^2}
\end{align*}
Where $h(s)$ corresponds to horizon index of the state partitions that contains state $s$.
\end{restatable}




\begin{proof}

Let $\tau_t$ be the trajectory sampled by our algorithm at time $t$. Notice that whenever $\tau_t \cap \pathpseudosub_\Delta \neq \emptyset$, there must be a state action pair in $\tau_t$ (occurring previous to the state in $\tau_t$ that lies in  $\pathpseudosub_\Delta$ ) that is in $\boundarypseudosub_\Delta$, namely the first state-action pair belonging to $\pseudosub_\Delta$ in $\tau_t$. We obtain,

\begin{equation*}
    \sum_{t=1}^T \mathbf{1}( \tau_t \cap \left(\pathpseudosub_\Delta \cup \pseudosub_\Delta\right) \neq \emptyset ) \leq \sum_{t=1}^T \mathbf{1}( \tau_t \cap \boundarypseudosub_\Delta \neq \emptyset ).
\end{equation*}

Conditioning on the event from Lemma~\ref{lemma::bounding_visitation_pseudosub} implies that for all $T \in \mathbb{N}$
\begin{align*}
  \sum_{t=1}^T \mathbf{1}( \tau_t \cap \boundarypseudosub_\Delta \neq \emptyset )
  \leq&\;  \sum_{(s,a) \in \boundarypseudosub_\Delta} N_{h(s)}^T(s,a)  \\
  \leq&\;|\boundarypseudosub_\Delta|\times \frac{8192 \beta^2 \times \max_{s', h'} \Vtilde^2_{h'}(s') \cdot \ln \frac{2 |\Scal||\Acal| t}{\delta} }{\Delta^2}.
\end{align*}

Thus,
\begin{align*}
    &\sum_{t=1}^T \mathbf{1}( \tau_t \cap \left(\pathpseudosub_\Delta \cup \pseudosub_\Delta\right) \neq \emptyset )\qquad\\
    \leq &\; |\boundarypseudosub_\Delta|\times \frac{8192 \beta^2 \times \max_{s', h'} \Vtilde^2_{h'}(s') \cdot \ln \frac{2 |\Scal||\Acal| t}{\delta} }{\Delta^2}.
\end{align*}

Finalizing the proof of the first bullet. To prove the second bullet, for any $t \in \mathbb{N}$ the following inequalities hold
\begin{align*}
    &\sum_{(s,a) \in \left(\pathpseudosub_\Delta \times \Acal \cup \pseudosub_\Delta\right)} N_{h(s)}^t(s,a)\\
    \leq &\; H  \sum_{i=1}^t \mathbf{1}( \tau_t \cap \left(\pathpseudosub_\Delta \times \Acal \cup \pseudosub_\Delta\right) \neq \emptyset )  \\
    \leq &\; H \sum_{i=1}^t \mathbf{1}( \tau_t \cap \boundarypseudosub_\Delta \neq \emptyset ) \\
    \leq &\; H \sum_{(s,a) \in \boundarypseudosub_\Delta} N_{h(s)}^t(s,a)\\
    \leq &\; H \times |\boundarypseudosub_\Delta|\times \frac{8192 \beta^2 \times \max_{s', h'} \Vtilde^2_{h'}(s') \cdot \ln \frac{2 |\Scal||\Acal| t}{\delta} }{\Delta^2}.
\end{align*}
The result follows.
\end{proof}



Let's now consider a refinement to the upper bound of Equation~\ref{equation::simulation_optimism}. For any\footnote{From now on we'll use $T$ to index the final timestep of the regret sequence we are bounding.} $T \in \mathbb{N}$,

\begin{align}
    \sum_{t=1}^T \Vstar(s_0) - V^{\pi_t}(s_0) &\leq \sum_{t=1}^T \mathbb{E}_{\tau \sim \pi_t }\left[ \sum_{h=1}^H b_h^t(s_h, a_h) + \left( \Phat_t(\cdot | s_h, a_h) - \Pstar( \cdot | s_h, a_h) \right) \cdot \Vhat_{h+1}^{\pi_t} \right] \notag\\
    &\leq \underbrace{\sum_{t=1}^T \mathbb{E}_{\tau \sim \pi_t }\left[ \sum_{h=1}^H b_h^t(s_h, a_h) + \left( \Phat_t(\cdot | s_h, a_h) - \Pstar( \cdot | s_h, a_h) \right) \cdot \left( \Vhat_{h+1}^{\pi_t} -\Vstar_{h+1} \right)\right] }_{\mathrm{I}} + \notag\\
    &\quad \sum_{t=1}^T \mathbb{E}_{\tau \sim \pi_t }\left[ \sum_{h=1}^H  \left( \Phat_t(\cdot | s_h, a_h) - \Pstar( \cdot | s_h, a_h) \right) \cdot \Vstar_{h+1} \right]  \label{equation::simulation_optimism_refined} .
\end{align}

By Equation~\ref{equation::bonus_justification} it follows that with probability at least $1-\delta$ for all $s,a,h,T \in \Scal \times \Acal \times [H]\times \mathbb{N}$,

\begin{equation*}
\sum_{t=1}^T \mathbb{E}_{\tau \sim \pi_t }\left[ \sum_{h=1}^H  \left( \Phat_t(\cdot | s_h, a_h) - \Pstar( \cdot | s_h, a_h) \right) \cdot \Vstar_{h+1} \right] \leq \sum_{t=1}^T \mathbb{E}_{\tau \sim \pi_t }\left[ \sum_{h=1}^H  b_h^t(s_h, a_h) \right]
\end{equation*}

We will now focus on bounding $\mathrm{I}$. We'll make use of Lemma 7.8 from~\cite{agarwal2019reinforcement}

\begin{lemma}[Lemma 7.8 from~\cite{agarwal2019reinforcement}]\label{lemma::lemma_7_8_from_theory_book}
With probability at least $1-\delta$ for all $t\in\mathbb{N},s \in \Scal,a \in \Acal$, we have,
\begin{equation*}
    \left| \Phat_t(s' | s,a) - \Pstar(s'|s,a)   \right| \leq \sqrt{ \frac{2\Pstar(s'|s,a) \ln\left(\frac{|\Scal| |\Acal| t H}{\delta}\right) }{N_{h(s)}^t(s,a)}  } + \frac{2\ln\left(\frac{|\Scal| |\Acal| t H}{\delta}\right)}{N_{h(s)}^t(s,a)}
\end{equation*}

Where $h(s)$ corresponds to horizon index of the state partitions that contains state $s$. 
\end{lemma}

A slight modification of the proof\footnote{Instead of using $|\Scal|$ to uniformly bound the support of $\Pstar(\cdot |s,a)$ we explicitly write the bound in terms of its support} of Lemma 7.7 in~\cite{agarwal2019reinforcement} yields,

\begin{lemma}[Support Aware version of Lemma 7.7 in~\cite{agarwal2019reinforcement}]\label{lemma::modified_lemma_7_7_theory_book}

With probability at least $1-\delta$ for all $t \in \mathbb{N},  s\in \Scal, a \in \Acal$ and all $f:\Scal \rightarrow \mathbb{R}$ satisfying $f : \Scal \rightarrow  [0, B]$ we have,
\begin{equation*}
    \left| \left( \Phat_t(\cdot | s,a) - \Pstar(\cdot | s,a) \right) f \right| \leq \frac{\mathbb{E}_{s' \sim \Pstar(\cdot | s,a) } \left[f(s') \right]}{H} + B\min\left(\frac{3\left| \mathrm{support}( \Pstar(\cdot | s,a))  \right| H\ln\left(\frac{|\Scal| |\Acal| t H}{\delta}\right)}{N_{h(s)}^t(s,a)} , 1\right)
\end{equation*}
Where $h(s)$ corresponds to horizon index of the state partitions that contains state $s$. 
\end{lemma}

\begin{proof}
We start by conditioning on the event that Lemma~\ref{lemma::lemma_7_8_from_theory_book} holds. Take any function $f : \Scal \rightarrow [0, B]$. We have,
\begin{align*}
        \left| \left( \Phat_t(\cdot | s,a) - \Pstar(\cdot | s,a) \right) f \right| &\leq \sum_{s' \in \Scal} \left| \left( \Phat_t(s' | s,a) - \Pstar(s' | s,a) \right)  \right| f(s') \\
        &\leq \sum_{s' \in \Scal} \sqrt{ \frac{2\Pstar(s'|s,a) \ln\left(\frac{|\Scal| |\Acal| t H}{\delta}\right) f^2(s') }{N_{h(s)}^t(s,a)}  } + \sum_{s' \in \Scal} \frac{2\ln\left(\frac{|\Scal| |\Acal| t H}{\delta}\right) f(s')}{N_{h(s)}^t(s,a)}\\
        &\stackrel{(i)}{\leq} \sum_{s' \in \Scal} \sqrt{ \frac{2\Pstar(s'|s,a) \ln\left(\frac{|\Scal| |\Acal| t H}{\delta}\right) f^2(s') }{N_{h(s)}^t(s,a)}  } +  \frac{2 B\left| \mathrm{support}( \Pstar(\cdot | s,a))  \right| \ln\left(\frac{|\Scal| |\Acal| t H}{\delta}\right) }{N_{h(s)}^t(s,a)} \\
        &\stackrel{(ii)}{\leq} \sqrt{ \left| \mathrm{support}( \Pstar(\cdot | s,a))  \right|  } \sqrt{ \frac{\sum_{s' \in \Scal} 2\Pstar(s'|s,a) \ln\left(\frac{|\Scal| |\Acal| t H}{\delta}\right) f^2(s') }{N_{h(s)}^t(s,a)}  } +  \\
        &\quad \frac{2B \left| \mathrm{support}( \Pstar(\cdot | s,a))  \right| \ln\left(\frac{|\Scal| |\Acal| t H}{\delta}\right) }{N_{h(s)}^t(s,a)}\\
        &=    \sqrt{ \frac{2\left| \mathrm{support}( \Pstar(\cdot | s,a))  \right| BH\ln\left(\frac{|\Scal| |\Acal| t H}{\delta}\right)}{N_{h(s)}^t(s,a)} \frac{\sum_{s' \in \Scal} \Pstar(s'|s,a)  f^2(s') }{BH}  } +  \\
        &\quad \frac{2B \left| \mathrm{support}( \Pstar(\cdot | s,a))  \right| \ln\left(\frac{|\Scal| |\Acal| t H}{\delta}\right) }{N_{h(s)}^t(s,a)} \\
        &\stackrel{(iii)}{\leq} \frac{\left| \mathrm{support}( \Pstar(\cdot | s,a))  \right| BH\ln\left(\frac{|\Scal| |\Acal| t H}{\delta}\right)}{N_{h(s)}^t(s,a)} + \frac{\sum_{s' \in \Scal} \Pstar(s'|s,a)  f^2(s') }{BH} + \\
        &\quad \frac{2B \left| \mathrm{support}( \Pstar(\cdot | s,a))  \right| \ln\left(\frac{|\Scal| |\Acal| t H}{\delta}\right) }{N_{h(s)}^t(s,a)} \\
        &\stackrel{(iv)}{\leq} \frac{\left| \mathrm{support}( \Pstar(\cdot | s,a))  \right| BH\ln\left(\frac{|\Scal| |\Acal| t H}{\delta}\right)}{N_{h(s)}^t(s,a)} + \frac{\sum_{s' \in \Scal} \Pstar(s'|s,a)  f(s') }{H} + \\
        &\quad \frac{2B \left| \mathrm{support}( \Pstar(\cdot | s,a))  \right| \ln\left(\frac{|\Scal| |\Acal| t H}{\delta}\right) }{N_{h(s)}^t(s,a)}
\end{align*}
 Inequality $(i)$ follows because $\| f\|_\infty \leq B$. Inequality~$(ii)$ is a result of Cauchy-Schwarz, $(iii)$ uses the inequality $ab \leq \frac{a^2 + b^2}{2}$ and $(iv)$ uses the condition $f \in [0, B]$ again. The above display implies, 
 
 \begin{equation*}
     \left| \left( \Phat_t(\cdot | s,a) - \Pstar(\cdot | s,a) \right) f \right|  \leq \frac{3\left| \mathrm{support}( \Pstar(\cdot | s,a))  \right| BH\ln\left(\frac{|\Scal| |\Acal| t H}{\delta}\right)}{N_{h(s)}^t(s,a)} + \frac{\sum_{s' \in \Scal} \Pstar(s'|s,a)  f(s') }{H} .
 \end{equation*}
Finally, since $ f (s) \in [0, B]$,
\begin{equation*}
 \left| \left( \Phat_t(\cdot | s,a) - \Pstar(\cdot | s,a) \right) f \right| \leq B.
\end{equation*}
Combining these last two equations we conclude,

\begin{equation*}
     \left| \left( \Phat_t(\cdot | s,a) - \Pstar(\cdot | s,a) \right) f \right| \leq \min\left(B,\frac{3\left| \mathrm{support}( \Pstar(\cdot | s,a))  \right| BH\ln\left(\frac{|\Scal| |\Acal| t H}{\delta}\right)}{N_{h(s)}^t(s,a)} \right) + \frac{\sum_{s' \in \Scal} \Pstar(s'|s,a)  f(s') }{H}.
\end{equation*}
The result follows.
\end{proof}

From now on we'll use the notation 
$$\xi_h^t(s,a) := (\beta-1) \Vtildemax \min\left(\frac{3\left| \mathrm{support}( \Pstar(\cdot | s,a))  \right|  H\ln\left(\frac{|\Scal| |\Acal| t H}{\delta}\right)}{N_h^t(s,a)} , 1\right).$$ 

Recall that by Assumption~\ref{assump::h_indexing}, if we define $h(s)$ as to horizon index of the state partitions that contains state $s$, we have $N_{h}(s,a) = 0 $ for all $h \neq h(s)$. 

The following corollary holds, 
\begin{corollary}
With probability at least $1-2\delta$ for all $t \in \mathbb{N}, s \in \Scal, a \in \Acal, h \in [H]$ we have,
\begin{align*}
     \left( \Phat_t(\cdot | s,a) - \Pstar(\cdot | s,a) \right) \left(  \Vhat_{h+1}^{\pi_t} -\Vstar_{h+1}  \right)   &\leq \frac{\mathbb{E}_{s' \sim \Pstar(\cdot | s,a) } \left[   \Vhat_{h+1}^{\pi_t}(s') -\Vstar_{h+1}(s')   \right]}{H} +\xi_h^t(s,a)
\end{align*}
\end{corollary}

\begin{proof}
Under the event of Lemma~\ref{lemma::optimism} (optimism), it follows that $\Vhat_{h+1}^{\pi_t}(s') \geq \Vstar_{h+1}(s')$ for all $t \in \mathbb{N}, s \in \Scal $. Moreover, $\Vhat_{h+1}^{\pi_t}(s') - \Vstar_{h+1}(s') \leq \beta \Vtilde_{h+1}(s')  - \Vstar_{h+1}(s') \leq (\beta-1) \Vtilde_{h+1}(s')$.

Thus we can set $B = (\beta-1) \Vtildemax$ in Lemma~\ref{lemma::modified_lemma_7_7_theory_book}. 
\end{proof}

We will now restate a modified version of Lemma 7.10 from~\cite{agarwal2019reinforcement} that provides us a bound for term $\mathrm{I}$ in terms of expectations over sums of $b_h^t$ and $\xi_h^t$ terms.

\begin{lemma}[Notation Adapted version of Lemma 7.10 from~\cite{agarwal2019reinforcement}]\label{lemma::adapted_lemma_7_10}
For all $T \in \mathbb{N}$ with probability at least $1-3\delta$,
\begin{equation*}
    \mathrm{I} \leq e \sum_{t=1}^T \mathbb{E}_{\tau \sim \pi_t }\left[ \sum_{h=1}^H  b_h^t(s_h, a_h) + \xi_h^t(s_h,a_h)\right].
\end{equation*}
\end{lemma}

Thus, with probability at least $1-3\delta$,

\begin{align*}
    \sum_{t=1}^T \Vstar(s_0) - V^{\pi_t}(s_0) &\leq  e \sum_{t=1}^T \mathbb{E}_{\tau \sim \pi_t }\left[ \sum_{h=1}^H  2b_h^t(s_h, a_h) + \xi_h^t(s_h,a_h)\right].
\end{align*}

The following Lemma will allow us to change from a sum of expectations over the played policies to a sum over the sample trajectories. 

\begin{lemma}\label{lemma::U_decomposition_refined}
Simultaneously for all $\mathcal{U} \subset \Scal \times \Acal$ and all $T \in \mathbb{N}$. 
\begin{align*}
       \sum_{t=1}^T \Vstar(s_0) - V^{\pi_t}(s_0) &\leq e\sum_{t=1}^T \sum_{h=1}^H  \mathbf{1}( (s^t_h, a^t_h) \in \mathcal{U} ) \left(2b_h^t(s^t_h, a^t_h) + \xi_h^t(s^t_h,a^t_h)\right)  + \\
       &\quad e\sum_{t=1}^T \sum_{h=1}^H  \mathbf{1}( (s^t_h, a^t_h) \not\in \mathcal{U} ) \left(2b_h^t(s^t_h, a^t_h) + \xi_h^t(s^t_h,a^t_h)\right)  + \\
       &\quad \Ocal\left(\beta \Vtildemax \sqrt{ HT \ln\left(\frac{T}{\delta} \right) }  \right)
\end{align*}
and
\begin{align*}
       \sum_{t=1}^T \Vstar(s_0) - V^{\pi_t}(s_0) &\leq e\sum_{t=1}^T  \mathbf{1}( \tau_t \cap \mathcal{U} = \emptyset ) \sum_{h=1}^H  \left(2b_h^t(s^t_h, a^t_h) + \xi_h^t(s^t_h,a^t_h)\right) + \\
       &\quad \Ocal\left(\beta \Vtildemax \sqrt{ HT \ln\left(\frac{T}{\delta} \right) }+  \beta H \Vtildemax\sum_{t=1}^T  \mathbf{1}( \tau_t \cap \mathcal{U}\neq \emptyset) \right)
\end{align*}
With probability at least $1-4\delta$. 

\end{lemma}

\begin{proof}
By Lemma~\ref{lemma::adapted_lemma_7_10}, for all $T \in \mathbb{N}$ with probability at least $1-3\delta$,

\begin{align}\label{equation::initial_upper_bound_regret_refined}
    \sum_{t=1}^T \Vstar(s_0) - V^{\pi_t}(s_0) &\leq  e \sum_{t=1}^T \mathbb{E}_{\tau \sim \pi_t }\left[ \sum_{h=1}^H  2b_h^t(s_h, a_h) + \xi_h^t(s_h,a_h)\right].
\end{align}

Since $|2b_h^t(s_h, a_h) + \xi_h^t(s_h,a_h) | \leq  \Ocal\left( \beta \Vtildemax \right)$, anytime Hoeffding lemma~\ref{lemma::anytime_hoeffding} and the union bound implies that for all $T \in \mathbb{N}$ with probability at least $1-4\delta$,

\begin{align}\label{equation::apply_hoeffding_refined_first}
    \sum_{t=1}^T \Vstar(s_0) - V^{\pi_t}(s_0) &\leq  e \sum_{t=1}^T \sum_{h=1}^H  2b_h^t(s^t_h, a^t_h) + \xi_h^t(s^t_h,a^t_h) + \Ocal\left( \beta \Vtildemax \sqrt{ HT \ln\left(\frac{T}{\delta} \right) } \right)
\end{align}

Observe that for any $\mathcal{U} \in \Scal \times \Acal$,



\begin{align}
      \sum_{t=1}^T \sum_{h=1}^H  2b_h^t(s^t_h, a^t_h) + \xi_h^t(s^t_h,a^t_h)
    &=  \sum_{t=1}^T  \mathbf{1}( \tau_t \cap \mathcal{U} = \emptyset ) \sum_{h=1}^H  \left(2b_h^t(s^t_h, a^t_h) + \xi_h^t(s^t_h,a^t_h)\right)  +\notag\\
    &\quad \sum_{t=1}^T  \mathbf{1}( \tau_t \cap \mathcal{U} \neq \emptyset )  \sum_{h=1}^H  \left(2b_h^t(s^t_h, a^t_h) + \xi_h^t(s^t_h,a^t_h)\right) \notag\\
    &\leq \sum_{t=1}^T  \mathbf{1}( \tau_t \cap \mathcal{U} = \emptyset ) \sum_{h=1}^H  \left(2b_h^t(s^t_h, a^t_h) + \xi_h^t(s^t_h,a^t_h)\right)  + \notag\\
    &\quad \Ocal\left(\beta H \Vtildemax\sum_{t=1}^T  \mathbf{1}( \tau_t \cap \mathcal{U}  \neq \emptyset)  \right) \label{equation::refined_upper_bound_traj}. 
\end{align}

The result follows by combining inequalities~\ref{equation::initial_upper_bound_regret_refined},~\ref{equation::apply_hoeffding_refined_first} and~\ref{equation::refined_upper_bound_traj}.

\end{proof}
\subsection{Full Proof of Theorem~\ref{theorem::main_result}} \label{section::full_proof_main_result}

We are ready to prove our main supporting Lemma, 

\begin{lemma}

With probability at east $1-6\delta$ for all $\Delta > 0$ and $T\in \mathbb{N}$ simultaneously,

\begin{align*}
     \sum_{t=1}^T \Vstar(s_0) - V^{\pi_t}(s_0)  \leq &\; \beta H \Vtildemax \sqrt{|\mathcal{U}^{\mathrm{good}}|T\ln \frac{|\Scal||\Acal|T}{\delta} } +\\
     &\;\Ocal\Bigg( \min\Bigg(  (\beta-1)\Vtildemax \left|\Scal \right|^2 |\Acal| H^2 \ln\left(\frac{|\Scal| |\Acal| T H}{\delta}\right)\log(T+1) + \\
     &\; \beta H \Vtildemax \sqrt{|\mathcal{U}^{\mathrm{good}}|T\ln \frac{2|\Scal||\Acal|T}{\delta} }  +\\   
     &\;\beta  \left(\Vtildemax\right)^2 \left(\ln \frac{|\Scal||\Acal|T}{\delta}\right) \frac{\sqrt{H|\Scal||\Acal| |\boundarypseudosub_\Delta|  }}{\Delta}  + \\
     &\;\beta \Vtildemax |\Scal||\Acal| \ln \frac{|\Scal||\Acal|T}{\delta} \ln(T+1), \\
     &\;(\beta-1)\Vtildemax \left|\Scal \backslash \pathpseudosub_\Delta  \right|  | \mathcal{U}^{\mathrm{good}}|H^2 \ln\left(\frac{|\Scal| |\Acal| T H}{\delta}\right) \log(T+1)+\\
     &\;\beta \Vtildemax |\mathcal{U}^{\mathrm{good}}| \ln \frac{|\Scal||\Acal|T}{\delta} \ln(T+1)+\\
     &\;  \beta H \Vtildemax \sqrt{|\mathcal{U}^{\mathrm{good}}|T\ln \frac{|\Scal||\Acal|T}{\delta} } + \\
     &\;\beta^3 H \left(\Vtildemax \right)^3|\boundarypseudosub_\Delta| \times \frac{ \ln \frac{ |\Scal||\Acal| T}{\delta} }{\Delta^2}    \Bigg)\Bigg)
\end{align*}

Where $\mathcal{U}^{\mathrm{good}} = (\Scal \times \Acal )\backslash\left( \pathpseudosub_\Delta \times \Acal \cup \pseudosub_\Delta\right)$. 

\end{lemma}

\begin{proof}
We will condition on the events from Lemmas~\ref{lemma::upper_bounding_sum_visitation_pseudosub-appendix} and~\ref{lemma::U_decomposition_refined}, something that happens with probability at least $1-6\delta$.

We'll start with the decomposition from Lemma~\ref{lemma::U_decomposition_refined}. 
Set $\mathcal{U}^{\mathrm{good}} = (\Scal \times \Acal)\backslash \left[\left(\pathpseudosub_\Delta \times \Acal \right) \cup \pseudosub_\Delta \right]$. In what follows we'll use $\mathcal{U}^{\mathrm{good}}$ to denote this set when convenient. We will also use the notation $\mathcal{U}^{\mathrm{bad}} = (\Scal \times \Acal) \backslash \mathcal{U}^{\mathrm{good}}  $ to denote the complement of $\mathcal{U}_{\mathrm{good}}$.

Recall that as a consequence of Equation~\ref{equation::neighbors_restricted} for all state action pairs $(s,a) \in  \mathcal{U}^{\mathrm{good}}$, we have that  $\neighbor(s,a) \subseteq \Scal \backslash \pathpseudosub_\Delta$. 

Therefore, the support of $\Pstar(\cdot | s,a )$ and $\Phat_t(\cdot | s,a) $ is contained in $\Scal \backslash \pathpseudosub_\Delta$ for $(s,a) \in  \Scal \backslash ( \pathpseudosub_\Delta \times \Acal \cup \pseudosub_\Delta)$. This implies that for all $(s,a) \in (\Scal \times \Acal )\backslash ( \pathpseudosub_\Delta \times \Acal \cup \pseudosub_\Delta)$, we have

\begin{align}\label{equation::upper_bound_xi}
 \xi_h^t(s,a) \leq (\beta-1) \Vtildemax \min\left(\frac{3\left|\Scal \backslash \pathpseudosub_\Delta  \right|  H\ln\left(\frac{|\Scal| |\Acal| t H}{\delta}\right)}{N_h^t(s,a)} , 1\right)~\forall~(s,a) \in \mathcal{U}^{\mathrm{good}}. \end{align}
 
 Similarly,

 \begin{align}\label{equation::upper_bound_xi_bad}
 \xi_h^t(s,a) \leq (\beta-1) \Vtildemax \min\left(\frac{3\left|\Scal \right|  H\ln\left(\frac{|\Scal| |\Acal| t H}{\delta}\right)}{N_h^t(s,a)} , 1\right)~\forall~(s,a) \in \mathcal{U}^{\mathrm{bad}}. \end{align}

Let $\mathcal{U}^{\mathrm{good}} = (\Scal \times \Acal )\backslash\left( \pathpseudosub_\Delta \times \Acal \cup \pseudosub_\Delta\right)$. Inequality~\ref{equation::upper_bound_xi} implies,
\begin{align}
     &\sum_{t=1}^T \sum_{h=1}^H   \mathbf{1}( (s^t_h, a^t_h) \in \mathcal{U}^{\mathrm{good}} ) \xi_{h}^t(s_h^t, a_h^t)  \label{equation::bound_xi_U_sum} \\
     \leq\;&  \sum_{t=1}^T \sum_{h=1}^H   \mathbf{1}( (s^t_h, a^t_h) \in \mathcal{U}^{\mathrm{good}} ) (\beta-1) \Vtildemax \min\left(\frac{3\left|\Scal \backslash \pathpseudosub_\Delta  \right|  H\ln\left(\frac{|\Scal| |\Acal| t H}{\delta}\right)}{N_h^t(s_h^t,a_h^t)} , 1\right) \notag\\
     \leq\;&  3(\beta-1)\Vtildemax \left|\Scal \backslash \pathpseudosub_\Delta  \right|  H \ln\left(\frac{|\Scal| |\Acal| T H}{\delta}\right)\sum_{t=1}^T \sum_{h=1}^H     \frac{\mathbf{1}( (s^t_h, a^t_h) \in \mathcal{U}^{\mathrm{good}} )}{N_h^t(s_h^t,a_h^t)} \notag \\
     \stackrel{(i)}{\leq}\;& 6(\beta-1)\Vtildemax \left|\Scal \backslash \pathpseudosub_\Delta  \right|  | \mathcal{U}^{\mathrm{good}}|H^2 \ln\left(\frac{|\Scal| |\Acal| T H}{\delta}\right) \log(T+1). \notag
\end{align}

Where inequality $(i)$ holds because of Lemma~\ref{lemma:inverse_counts_bounds}. The exact same argument applied to the upper bound of Equation~\ref{equation::bound_xi_U_sum} implies,
\begin{align}
     \sum_{t=1}^T \sum_{h=1}^H   \mathbf{1}( (s^t_h, a^t_h) \in \mathcal{U}^{\mathrm{bad}} ) \xi_{h}^t(s_h^t, a_h^t)  
     &\leq  6(\beta-1)\Vtildemax \left|\Scal \right| | \mathcal{U}^{\mathrm{bad}}|  H^2 \ln\left(\frac{|\Scal| |\Acal| T H}{\delta}\right)\log(T+1) \notag\\
     &\leq  6(\beta-1)\Vtildemax \left|\Scal \right|^2 |\Acal| H^2 \ln\left(\frac{|\Scal| |\Acal| T H}{\delta}\right)\log(T+1).\label{equation::bound_xi_U_sum_bad}
\end{align}

Let's now bound the sum $ \sum_{t=1}^T \sum_{h=1}^H   \mathbf{1}( (s^t_h, a^t_h) \in \mathcal{U}^{\mathrm{good}} ) b_h^t(s_h^t,a_h^t) $.

\begin{align}
    &\sum_{t=1}^T \sum_{h=1}^H   \mathbf{1}( (s^t_h, a^t_h) \in \mathcal{U}^{\mathrm{good}} ) b_h^t(s_h^t,a_h^t)  \label{equation::bound_b_h_U_sum}\\
    =&\; \sum_{t=1}^T \sum_{h=1}^H \mathbf{1}( (s^t_h, a^t_h) \in \mathcal{U}^{\mathrm{good}} ) \min\left( 16 \beta \sqrt { \frac{ \widehat{\mathbb{E}}_{s' \sim \Phat_t(\cdot | s_h^t,a_h^t) }[\Vtilde^2_{h+1}(s') | s_h^t,a_h^t ]  \ln\frac{ 2 |\Scal||\Acal|}{\delta} }{ N_h^t(s_h^t,a_h^t) }}  +  \frac{12\beta \Vtildemax }{N_h^t(s_h^t,a_h^t)}\ln \frac{ 2 |\Scal||\Acal|t}{\delta}, 2\beta \Vtildemax\right) \notag \\
    \leq&\; \sum_{t=1}^T \sum_{h=1}^H \mathbf{1}( (s^t_h, a^t_h) \in \mathcal{U}^{\mathrm{good}} ) \left( 16 \beta \sqrt { \frac{ \widehat{\mathbb{E}}_{s' \sim \Phat_t(\cdot | s_h^t,a_h^t) }[\Vtilde^2_{h+1}(s') | s_h^t,a_h^t ]  \ln\frac{ 2 |\Scal||\Acal|}{\delta} }{ N_h^t(s_h^t,a_h^t) }}  +  \frac{12\beta \Vtildemax }{N_h^t(s_h^t,a_h^t)}\ln \frac{ 2 |\Scal||\Acal|t}{\delta}\right)\notag\\
    \stackrel{(a)}{\leq}&\; 16\beta \Vtildemax \sqrt{\ln \frac{2|\Scal||\Acal|T}{\delta} } \sum_{t=1}^T \sum_{h=1}^H    \frac{\mathbf{1}( (s^t_h, a^t_h) \in \mathcal{U}^{\mathrm{good}} ) }{ \sqrt{N_h^t(s_h^t,a_h^t) }}  +
    12\beta \Vtildemax \ln \frac{2|\Scal||\Acal|T}{\delta} \sum_{t=1}^T \sum_{h=1}^H    \frac{\mathbf{1}( (s^t_h, a^t_h) \in \mathcal{U}^{\mathrm{good}} ) }{N_h^t(s_h^t,a_h^t)} \notag\\
        \stackrel{(i)}{\leq} &\; 32\beta H \Vtildemax \sqrt{|\mathcal{U}^{\mathrm{good}}|T\ln \frac{2|\Scal||\Acal|T}{\delta} }  +
    24\beta \Vtildemax |\mathcal{U}^{\mathrm{good}}| \ln \frac{2|\Scal||\Acal|T}{\delta} \ln(T+1) . \notag
\end{align}
Where inequality $(i)$ holds because of Lemmas~\ref{lemma:inverse_squareroot_bounds} and~\ref{lemma:inverse_counts_bounds}. The exact same proof argument as in the proof of inequality~\ref{equation::bound_b_h_U_sum} that leads to inequality $(a)$ yields,

\begin{align}
    &\sum_{t=1}^T \sum_{h=1}^H   \mathbf{1}( (s^t_h, a^t_h) \in \mathcal{U}^{\mathrm{bad}} ) b_h^t(s_h^t,a_h^t)  \label{equation::bound_b_h_U_sum_bad} \\
    \leq&\; 16\beta \Vtildemax \sqrt{\ln \frac{2|\Scal||\Acal|T}{\delta} } \sum_{t=1}^T \sum_{h=1}^H    \frac{\mathbf{1}( (s^t_h, a^t_h) \in \mathcal{U}^{\mathrm{bad}} ) }{ \sqrt{N_h^t(s_h^t,a_h^t) }}  +
    12\beta \Vtildemax \ln \frac{2|\Scal||\Acal|T}{\delta} \sum_{t=1}^T \sum_{h=1}^H    \frac{\mathbf{1}( (s^t_h, a^t_h) \in \mathcal{U}^{\mathrm{bad}} ) }{N_h^t(s_h^t,a_h^t)}  \notag\\
    \stackrel{(i)}{\leq}&\; 32\beta  \Vtildemax \sqrt{H|\mathcal{U}^{\mathrm{bad}}|\ln \frac{2|\Scal||\Acal|T}{\delta}\sum_{(s,a) \in (\Scal \times \Acal) \backslash\mathcal{U}^{\mathrm{good}}} \sum_{h=1}^{H} N_h^T(s,a)  }  +
    24\beta \Vtildemax |\mathcal{U}^{\mathrm{bad}}| \ln \frac{2|\Scal||\Acal|T}{\delta} \ln(T+1) \notag\\
    \stackrel{(ii)}{\leq}&\; 32 \times 8192\beta  \left(\Vtildemax\right)^2 \left(\ln \frac{2|\Scal||\Acal|T}{\delta}\right) \frac{\sqrt{H|\mathcal{U}^{\mathrm{bad}}| |\boundarypseudosub_\Delta|  }}{\Delta}  +\notag\\
    &\;24\beta \Vtildemax |\mathcal{U}^{\mathrm{bad}}| \ln \frac{2|\Scal||\Acal|T}{\delta} \ln(T+1)\notag \\
    = &\;\Ocal\left(\beta  \left(\Vtildemax\right)^2 \left(\ln \frac{2|\Scal||\Acal|T}{\delta}\right) \frac{\sqrt{H|\Scal||\Acal| |\boundarypseudosub_\Delta|  }}{\Delta} \right) +\notag\\
    &\; \Ocal\left(\beta \Vtildemax |\Scal||\Acal| \ln \frac{2|\Scal||\Acal|T}{\delta} \ln(T+1) \right) \notag,
\end{align}
where inequality $(i)$ follows because of Lemmas~\ref{lemma:inverse_squareroot_bounds} and~\ref{lemma:inverse_counts_bounds}.  And inequality $(ii)$ follows from Lemma~\ref{lemma::upper_bounding_sum_visitation_pseudosub-appendix}.

Combining inequalities~\ref{equation::bound_xi_U_sum},~\ref{equation::bound_xi_U_sum_bad},~\ref{equation::bound_b_h_U_sum} and~\ref{equation::bound_b_h_U_sum_bad} with Lemma~\ref{lemma::U_decomposition_refined} we get,

\begin{align*}
       \sum_{t=1}^T \Vstar(s_0) - V^{\pi_t}(s_0) \leq&\; e\sum_{t=1}^T \sum_{h=1}^H  \mathbf{1}( (s^t_h, a^t_h) \in \mathcal{U}^{\mathrm{good}} ) \left(2b_h^t(s^t_h, a^t_h) + \xi_h^t(s^t_h,a^t_h)\right) + \\
       &\; e\sum_{t=1}^T \sum_{h=1}^H  \mathbf{1}( (s^t_h, a^t_h) \in \mathcal{U}^{\mathrm{bad}} ) \left(2b_h^t(s^t_h, a^t_h) + \xi_h^t(s^t_h,a^t_h)\right)+ \Ocal\left(\beta \Vtildemax \sqrt{ HT \ln\left(\frac{T}{\delta} \right) }  \right) \\
       \leq&\; \Ocal\left((\beta-1)\Vtildemax \left|\Scal \right|^2 |\Acal| H^2 \ln\left(\frac{|\Scal| |\Acal| T H}{\delta}\right)\log(T+1) \right)+\\
       &\;\Ocal\left( \beta H \Vtildemax \sqrt{|\mathcal{U}^{\mathrm{good}}|T\ln \frac{2|\Scal||\Acal|T}{\delta} }   \right) + \\
        &\;\Ocal\left(\beta  \left(\Vtildemax\right)^2 \left(\ln \frac{|\Scal||\Acal|T}{\delta}\right) \frac{\sqrt{H|\Scal||\Acal| |\boundarypseudosub_\Delta|  }}{\Delta} \right) +\\
    &\;\Ocal\left(\beta \Vtildemax |\Scal||\Acal| \ln \frac{|\Scal||\Acal|T}{\delta} \ln(T+1) \right) 
\end{align*}
We will now derive a bound that has only logarithmic dependence on $\Scal$ at the cost of a quadratic dependence on $\Delta^2$.

By Lemma~\ref{lemma::U_decomposition_refined}, inequalities~\ref{equation::bound_xi_U_sum} and~\ref{equation::bound_b_h_U_sum} and Lemma~\ref{lemma::upper_bounding_sum_visitation_pseudosub-appendix}. 

\begin{align*}
       \sum_{t=1}^T \Vstar(s_0) - V^{\pi_t}(s_0) &\leq e\sum_{t=1}^T  \mathbf{1}( \tau_t \cap \mathcal{U}^{\mathrm{bad}} = \emptyset ) \sum_{h=1}^H  \left(2b_h^t(s^t_h, a^t_h) + \xi_h^t(s^t_h,a^t_h)\right) + \\
       &\quad \Ocal\left(\beta \Vtildemax \sqrt{ HT \ln\left(\frac{T}{\delta} \right) }+  \beta H \Vtildemax\sum_{t=1}^T  \mathbf{1}( \tau_t \cap \mathcal{U}\neq \emptyset) \right) \\
       &\leq  e\sum_{t=1}^T   \sum_{h=1}^H   \mathbf{1}( (s^t_h, a^t_h) \in \mathcal{U}^{\mathrm{good}} ) \left(2b_h^t(s^t_h, a^t_h) + \xi_h^t(s^t_h,a^t_h)\right) + \\
       &\quad  \Ocal\left(\beta \Vtildemax \sqrt{ HT \ln\left(\frac{T}{\delta} \right) }+  \beta H \Vtildemax\sum_{t=1}^T  \mathbf{1}( \tau_t \cap \mathcal{U}\neq \emptyset) \right) \\
       &\stackrel{(i)}{=} \Ocal\left((\beta-1)\Vtildemax \left|\Scal \backslash \pathpseudosub_\Delta  \right|  | \mathcal{U}^{\mathrm{good}}|H^2 \ln\left(\frac{|\Scal| |\Acal| T H}{\delta}\right) \log(T+1) \right) + \\
       &\quad \Ocal\left( \beta \Vtildemax |\mathcal{U}^{\mathrm{good}}| \ln \frac{|\Scal||\Acal|T}{\delta} \ln(T+1)\right) +\\
       &\quad \Ocal\left( \beta H \Vtildemax \sqrt{|\mathcal{U}^{\mathrm{good}}|T\ln \frac{|\Scal||\Acal|T}{\delta} }  \right)+\\
        &\quad \Ocal\left( \beta^3 H \left(\Vtildemax \right)^3    |\boundarypseudosub_\Delta| \times \frac{ \ln \frac{ |\Scal||\Acal| T}{\delta} }{\Delta^2}  \right)
\end{align*}
This holds for all $\Delta \in [0,1]$ simultaneously because Lemma~\ref{lemma::upper_bounding_sum_visitation_pseudosub-appendix} holds simultaneously for all $\Delta \in [0,1]$ and for all $T \in \mathbb{N}$ because all the events we have conditioned refer to properties that hold for all $T \in \mathbb{N}$. This finalizes the proof of the desired result. 
\end{proof}

Using the bound $|\mathcal{U}^{\mathrm{good}}| \leq \left|\Scal \backslash \pathpseudosub_\Delta  \right| |\Acal|$, we have the following version of our results

\begin{theorem}[Full version of Theorem~\ref{theorem::main_result}]\label{theorem::main_results_refined}
The regret of UCBVI - Shaped satisfies,

\begin{align*}
\sum_{t=1}^T   \Vstar(s_0) - V^{\pi_t}(s_0) &= \Ocal\Bigg( \min_{\Delta}\Bigg(  H\beta \Vtildemax  \sqrt{| \Scal \backslash \pathpseudosub_\Delta| |\Acal| T \ln \frac{\Vtildemax |\Scal| |\Acal| T}{\delta} } +  \\
&\quad   \beta \Vtildemax    \ln \frac{\Vtildemax |\Scal| |\Acal| T}{\delta}\times \min \left( \bar{A}(\Delta), \bar{B}(\Delta) \right) \Bigg)   \Bigg).
\end{align*}
For all $T \in \mathbb{N}$ with probability at least $1-6\delta$. Where 
\begin{align*}
  \bar{A}(\Delta) &= \frac{  \beta \Vtildemax H^{1/2}|\Scal|^{1/2} |\Acal|^{1/2}  |\boundarypseudosub_\Delta|^{1/2} }{\Delta} +  \frac{(\beta-1)}{\beta} \left|\Scal \right|^2 |\Acal| H^2 \log(T+1)   \\
  &\quad + |\Scal||\Acal|  \ln(T+1),
\end{align*}
and
\begin{align*}
\bar{B}(\Delta) &= \frac{\beta^2 \left(\Vtildemax\right)^2H|\boundarypseudosub_\Delta| }{\Delta^2}   +  \frac{(\beta-1)}{\beta} \left|\Scal \backslash \pathpseudosub_\Delta  \right|^2|\Acal| H^2  \log(T+1) \\
&\quad + \left|\Scal \backslash \pathpseudosub_\Delta  \right||\Acal| \ln(T+1).
\end{align*}
\end{theorem}

\subsection{Generalizations}\label{section::generalizations}

In this section we briefly discuss what can we say in the case the shaping functions $\{ \Qtilde_h\}_{h \in [H]}$ are of the form $\Qtilde_h : \Scal \times \Acal \rightarrow \mathbb{R} $ such that $\Qstar_h(s,a) \leq \beta \Qtilde_h(s,a) $ instead of our state-only assumption. By doing so we recover some of the results of~\cite{golowich2022can} for which we provide simple proofs. We will use the notation $\Qtildemax = \max_{s,a,h} \Qtilde_h(s,a)$.

Let's consider a generic optimistic RL algorithm $\mathbb{A}$ that computes an \emph{optimistic} $Q$ function estimator $\{\widehat Q_h^t\}_{h \in [H]}$ at the beginning of time-step $t$ satisfying $\widehat{Q}_h^t(s,\pi_\star(s)) \geq Q_\star(s,\pi_\star(s))$. Notice that we only consider optimism at the optimal action $\pi_\star(s)$. Let's also assume the policy played by $\mathbb{A}$ at time $t$ equals $\pi_t(|s,a,h) = \argmax_{a \in \Acal} \widehat Q_h^t(s,a)$. We will introduce the notion of a least upper bound stability for optimistic algorithms. 

\begin{definition}
We say that an optimistic algorithm $\mathbb{A}$ is stable w.r.t. to clipping if substituting the $\{ \Qtilde_h^t\}_{h\in [H]}$ values by their clipped versions
\begin{align*}
    \Qhat^{t, \mathrm{clipped}}_h(s, a) &\leftarrow\textcolor{blue}{\min \Big(}\wh{Q}_h^t(s,a)\textcolor{blue}{, \beta\Qtilde(s, a)\Big)}
\end{align*}
does not affect its regret guarantees. 
\end{definition}

 Clipping is done \emph{after} the $\wh{Q}_h^t(s,a)$ values have been computed. 


\textbf{Optimism is preserved.} When the un-clipped $\{\widehat Q_h^t\}_{h\in [H]}$ are optimistic, then  as long as $\beta \widetilde Q_h(s,\pi_\star(s)) \geq Q_\star(s, \pi_\star(s))$, the maximum among all the clipped values is also optimistic.
\begin{equation*}
\max_{a \in \Acal} \wh{Q}^{t, \mathrm{clipped}}_h(s, a) = \max_{a \in \Acal} \left[ \min \Big(\wh{Q}_h^t(s,a), \beta\Qtilde(s, a)\Big) \right]\geq \min \Big(\wh{Q}_h^t(s,\pi_\star(s)), \beta\Qtilde(s, \pi_\star(s))\Big)\geq \Qstar(s, \pi_\star(s))
\end{equation*}

If $\mathbb{A}$'s policy at time $t$ equals $\pi_t(|s,h) = \argmax_{a \in \Acal}  \Qhat^{t, \mathrm{clipped}}_h(s, a)$, as long as the un-clipped $\{\Qhat_h^t\}_{h\in [H]}$ are optimistic the support of policy $\pi_t$ satisfies
\begin{equation*}
    \mathrm{Support}( \pi_t( \cdot | s,a,h) ) = \{ a' \text{ s.t. } \beta \Qtilde_h( s,a' ) \geq \Qstar_h(s,\pi_\star(s,h))  \} 
\end{equation*}
This is simply because for all $a \in \Acal \backslash \{ a' \text{ s.t. } \beta \Qtilde_h( s,a' ) \geq Q_h(s,\pi_\star(s,h)  \} $, the clipped $\{\Qhat^{t, \mathrm{clipped}}_h(s, a)\}_{h\in [H]}$ satisfy
\begin{equation*}
    \Qhat^{t, \mathrm{clipped}}_h(s, a)< \Qstar_h(s,\pi_\star(s,h)) 
\end{equation*}
In other words, for all $t \in \mathbb{N}$ the clipped algorithm will only visit state-action pairs in $ \cup_{s \in \Scal, h \in [H]}\{ a \text{ s.t. } \beta \Qtilde_h( s,a ) \geq \Qstar_h(s,\pi_\star(s,h))  \} $. Alternatively this can be thought of as if the learner were to interact only with a prunned MDP with state space $\Scal$ and state (and horizon) dependent action sets $\Acal(s)$ of the form $\Acal(s) =  \{ a \text{ s.t. } \beta \Qtilde_{h(s)}( s,a ) \geq \Qstar_{h(s)}(s,\pi_\star(s,h(s)))  \} $.

It is easy to see the corresponding version of UCBVI-Shaped satisfies the following regret guarantee that recovers the min-max rates of Theorem 1.1 in~\citep{golowich2022can},

\begin{lemma}
The regret of UCBVI-Shaped with $\Qtilde$-shaping satisfies,
\begin{align*}
     \sum_{t=1}^T \Vstar(s_0) - V^{\pi_t}(s_0) &\leq \Ocal\left( \beta  \Qtildemax \sqrt{|\mathrm{Pair}_{\mathrm{eff}}|T\ln \frac{|\Scal||\Acal|T}{\delta} }  \right) + \\
     &\quad \Ocal\left( \beta \Qtildemax H \left|\Scal  \right||\mathrm{Pair}_{\mathrm{eff}}| \ln \frac{|\Scal||\Acal|T}{\delta} \ln(T+1)  \right) 
\end{align*}
For all $T \in \mathbb{N}$ with probability at least $1-\delta$, where\footnote{Recall we have assumed that $\mathcal{S}$ is $h$ indexed so that any state $s \in \mathcal{S}$ can be accessed only during in-episode time-step $h(s)$.} $\mathrm{Pair}_{\mathrm{eff}} = \{ (s,a) \in \Scal \times \Acal\text{ s.t. }  a \in \Acal(s)\}$. 
\end{lemma}

\begin{proof}
We borrow the bound from Theorem 7.6 in~\cite{agarwal2019reinforcement}. Observe that as long as optimism holds (Lemma~\ref{lemma::optimism}), the regret decomposition of Lemma~\ref{lemma::U_decomposition_refined} is satisfied. 

\begin{align*}
     \sum_{t=1}^T \Vstar(s_0) - V^{\pi_t}(s_0) \leq&\; e\sum_{t=1}^T \sum_{h=1}^H  \left(2b_h^t(s^t_h, a^t_h) + \xi_h^t(s^t_h,a^t_h)\right) +  \Ocal\left(\beta \Qtildemax \sqrt{ HT \ln\left(\frac{T}{\delta} \right) }  \right) 
\end{align*}

Notice that as long as optimism holds, for all $s_h^t$ the action $a_h^t$ satisfies $a_h^t \in \Acal(s_h^t)$. Thus, 

\begin{align*}
     \sum_{t=1}^T \sum_{h=1}^H   \xi_{h}^t(s_h^t, a_h^t)& \leq  \sum_{t=1}^T \sum_{h=1}^H   \beta \Qtildemax \min\left(\frac{3\left|\Scal  \right|  H\ln\left(\frac{|\Scal| |\Acal| t H}{\delta}\right)}{N_h^t(s_h^t,a_h^t)} , 1\right) \notag\\
     \leq\;&  3\beta\Qtildemax \left|\Scal  \right|  H \ln\left(\frac{|\Scal| |\Acal| T H}{\delta}\right)\sum_{t=1}^T \sum_{h=1}^H     \frac{1}{N_h^t(s_h^t,a_h^t)} \notag \\
     \stackrel{(i)}{\leq}\;& 6\beta\Qtildemax \left|\Scal  \right|  \mathrm{Pair}_{\mathrm{eff}} \cdot H \ln\left(\frac{|\Scal| |\Acal| T H}{\delta}\right) \log(T+1). \notag
\end{align*}

Where inequality $(i)$ follows from Lemma~\ref{lemma:inverse_counts_bounds}. Similarly,

\begin{align*}
    \sum_{t=1}^T \sum_{h=1}^H  b_h^t(s_h^t,a_h^t)    &\leq 16\beta \Qtildemax \sqrt{\ln \frac{2|\Scal||\Acal|T}{\delta} } \sum_{t=1}^T \sum_{h=1}^H    \frac{1 }{ \sqrt{N_h^t(s_h^t,a_h^t) }}  \\
    &\quad +
    12\beta \Qtildemax \ln \frac{2|\Scal||\Acal|T}{\delta} \sum_{t=1}^T \sum_{h=1}^H    \frac{1 }{N_h^t(s_h^t,a_h^t)}  \notag\\
    \stackrel{(i)}{\leq}&\; 32\beta  \Qtildemax \sqrt{|\mathrm{Pair}_{\mathrm{eff}}|T\ln \frac{2|\Scal||\Acal|T}{\delta} }  \\
    &\quad +
    24\beta \Qtildemax |\mathrm{Pair}_{\mathrm{eff}}| \ln \frac{2|\Scal||\Acal|T}{\delta} \ln(T+1) 
\end{align*}

Where $(i)$ holds by Lemmas~\ref{lemma:inverse_squareroot_bounds} and~\ref{lemma:inverse_counts_bounds}. Combining these last two inequalities yields

\begin{align*}
     \sum_{t=1}^T \Vstar(s_0) - V^{\pi_t}(s_0) &\leq \Ocal\left( \beta  \Qtildemax \sqrt{|\mathrm{Pair}_{\mathrm{eff}}|T\ln \frac{|\Scal||\Acal|T}{\delta} }  \right) + \\
     &\quad \Ocal\left( \beta \Qtildemax H \left|\Scal  \right||\mathrm{Pair}_{\mathrm{eff}}| \ln \frac{|\Scal||\Acal|T}{\delta} \ln(T+1)  \right) 
\end{align*}
The result follows.

\end{proof}

Similarly, we can obtain a result for instance dependent bounds for a clipped version of the StrongEuler algorithm. We consider the exact same clipping mechanism as in the previous discussion. Optimism guarantees (just as we have described above) that $\mathrm{Support}( \pi_t( \cdot | s,a,h) ) = \{ a' \text{ s.t. } \beta \Qtilde_h( s,a' ) \geq \Qstar_h(s,\pi_\star(s,h))  \} $. Therefore the data generated by the clipped StrongEuler algorithm will be produced from the `prunned' MDP with state space $\mathcal{S}$ and state-horizon dependent action sets $\mathcal{A}(s)$ (i.e. with state-action space $\mathrm{Pair}_{\mathrm{eff}}$). Strong optimism holds for the estimates $\Qhat_h^t(s,a)$ for all $(s,a) \in \mathrm{Pair}_{\mathrm{eff}}$ and all time steps $t$ (notice that $\pi_*$ is an optimal policy both in the original as well as in the pruned MDP). The exact same proofs as in~\cite{dann2021beyond} (see proof of Theorem 3.2) hold in this case with the only modification being to adapt the argument for MDPs where the number of actions is state-dependent thus yielding the following result,

\begin{lemma}\label{lemma::generalization_golowich}
The regret of clipped-StrongEuler satisfies with high probability,
\begin{equation*}
\sum_{t=1}^T \Vstar(s_0) - V^{\pi_t}(s_0) =\mathcal{O}\left( \sum_{(s,a) \in \mathrm{Pair}_{\mathrm{eff}}} \frac{\Qstar_{h(s)}(s,a)}{\mathrm{gap}_{h(s)}(s,a)} \log(T) \right)
\end{equation*}
where $\mathrm{gap}_h(s,a) = \Vstar_h(s) - \Qstar_h(s,a)$
\end{lemma}

Stating this result in terms of the return gaps (see Definition 3.1 in~\cite{dann2021beyond}) over $\mathrm{Pair}_{\mathrm{eff}}$ is possible. Lemma~\ref{lemma::generalization_golowich} recovers the instance dependent results of~\cite{golowich2022can}.

\subsection{Connections with instance dependent rates for reinforcement learning}

Although our main results are not instance dependent, our rates can be much better if the gaps over the effective state space are very small and the effective state space is also very small. Moreover, in contrast with instance dependent rates our results depend on an effective state space size that may be much smaller than the original state space. Our main innovation lies in these state pruning results.

\subsection{Model Selection}\label{section::model_selection}

Although UCBVI-Shaped requires knowledge of $\beta$, Algorithm~\ref{algo:UCBVI_shaped} can be used to design a parameter free version. As explained in Section~\ref{sec:practical} we simply require to initialize algorithm~\ref{algo:UCBVI_shaped} with a set of $N$ different $\beta$ parameter guesses $[\beta_1, \cdots, \beta_N]$. A good choice for these is an exponential parameter grid of the form $\beta_1 = 1, \cdots, \beta_N = 2^N$.

In the case we are using the Corral algorithm of~\citep{pacchiano2020model} (and~\citep{agarwal2017corralling}), it is enough to set the learning rates based on a putative regret bound value of $\sqrt{T}$ (see the proof of Theorem 5.3 in~\cite{pacchiano2020model} where the learning rate can be set to $\eta = \sqrt{\frac{N}{T}}$). The main term in the regret guarantee of Theorem~\ref{theorem::main_result} scales with $\sqrt{T}$. Thus, for any choice of $\beta_i$ if this value of $\beta_i$ was valid, the regret rate would scale as $C_i\sqrt{T\log(1/\delta)}$ for some parameter $C_i \in [1, H \beta_i \Vtildemax\sqrt{|\Scal||\Acal|}]$. If using regret Balancing as an online model selection mechanism (see~\citep{pacchiano2020regret} and~\citep{cutkosky2021dynamic,abbasi2020regret}) we can use an exponential parameter grid of the form $\left\{\beta_i \times [1, 2, \cdots, H \beta_i \Vtildemax\sqrt{|\Scal||\Acal|}\right\}_{i=1}^N$. The number of base algorithms necessary in this case is at most $N\log(H \beta_N \Vtildemax\sqrt{|\Scal||\Acal|})$.

As a consequence of Theorem 5.3 and as a simple corollary of Theorem 5.3 in~\cite{pacchiano2020model} or Theorem 5.1 in~\cite{pacchiano2020regret} we conclude that,

\begin{lemma}
Provided that $\beta_N \Vtilde_h(s) \geq \Vstar_h(s) $ for all $s \in \Scal$ and $h \in [H]$ (although it could be that $\beta_i \Vtilde_h(s) \geq \Vstar_h(s) $ for all $s \in \Scal$ and $h \in [H]$ for $i \ll N$ ), the expected regret of Algorithm~\ref{algo:UCBVI_shaped} satisfies,
\begin{equation*}
     \mathbb{E}\left[ \sum_{t=1}^T   \Vstar(s_0) - V^{\pi_t}(s_0)  \right] = \widetilde{\mathcal{O}}\left(  C^2\sqrt{N T }  \right)
\end{equation*}
when using Stochastic CORRAL as a model selection strategy. Similarly the expected regret of Algorithm~\ref{algo:UCBVI_shaped} satisfies,
\begin{equation*}
    \sum_{t=1}^T   \Vstar(s_0) - V^{\pi_t}(s_0)   = \widetilde{\mathcal{O}}\left(  C^2\sqrt{N \log(H \beta_N \Vtildemax\sqrt{|\Scal||\Acal|}) T \log(1/\delta)}  \right)
\end{equation*}
with probability at least $1-\mathcal{O}(\delta)$ when using Regret Balancing as a model selection strategy.
\end{lemma}

\begin{proof}
This result is (almost) an immediate corollary of Theorems 5.3 and 5.1 in~\citep{pacchiano2020model} and~\citep{pacchiano2020model} respectively. In order to apply these results to our setting we only need to note that for the optimal value of $\beta$ (among the choices in $[\beta_1, \cdots, \beta_N]$) there exists a constant $C \in [1, 2, \cdots, H \beta_N \Vtildemax\sqrt{|\Scal||\Acal|}]$ such that regret rate of UCBVI-Shaped (Theorem~\ref{theorem::main_result}) can be upper bounded as,

\begin{align*}
    \sum_{t=1}^T   \Vstar(s_0) - V^{\pi_t}(s_0) = \widetilde{\mathcal{O}}\left(  C\sqrt{T \log(1/\delta)}  \right).
\end{align*}

with probability at least $1-\mathcal{O}(\delta)$ and where $\widetilde{\mathcal{O}}$ may hide polynomial factors in $|\mathcal{S}|$ and $|\mathcal{A}|$.

Applying the in-expectation results of Theorem 5.3 in~\cite{pacchiano2020model} for the Stochastic CORRAL model selection algorithm yields (setting $\delta \ll \frac{1}{T}$),

\begin{align*}
    \mathbb{E}\left[ \sum_{t=1}^T   \Vstar(s_0) - V^{\pi_t}(s_0)  \right] = \widetilde{\mathcal{O}}\left(  C^2\sqrt{N T \log(1/\delta)}  \right)
\end{align*}
Where $\widetilde{\mathcal{O}}$ may hide polynomial factors in $|\mathcal{S}|$ and $|\mathcal{A}|$ and $T$. Similarly, Regret Balancing yields a bound,
\begin{align*}
    \sum_{t=1}^T   \Vstar(s_0) - V^{\pi_t}(s_0)   = \widetilde{\mathcal{O}}\left(  C^2\sqrt{N \log(H \beta_N \Vtildemax\sqrt{|\Scal||\Acal|}) T \log(1/\delta)}  \right)
\end{align*}
with probability at least $1-\mathcal{O}(\delta)$. 

\end{proof}

The pseudocode for this algoritihm is provided below:

\begin{small}
\begin{algorithm}[!h]
  \caption{Online UCBVI - Shaped}\label{algo:UCBVI_shaped_model_selection}
  \begin{algorithmic}[1]
    \State \textbf{Input} reward function $r$ (assumed to be known), set of $\beta$ values - $\left[ \beta_1, \beta_2, \dots, \beta_N\right]$
    \State Initialize model selection probability $p(\beta)$ as a uniform distribution over $\left[ \beta_1, \beta_2, \dots, \beta_N\right]$
      \State \textbf{for} {$t=1,...,T$}
        \State \qquad Sample a value of $\beta \sim p(\beta)$
        \State \qquad Run UCBVI-Shaped ($\beta$) with this sampled beta
        \State \qquad Update $p(\beta)$ using samples from UCBVI-Shaped using an online model selection algorithm.
      \State \textbf{End for }
  \end{algorithmic}
\end{algorithm}
\end{small}

\section{Supporting Technical Lemmas}

\begin{lemma}[Anytime Hoeffding Inequality~\citep{pacchiano2020regret}]\label{lemma::anytime_hoeffding} Let $\{ Y_\ell\}_{\ell=1}^\infty$ be a martingale difference sequence such that $Y_\ell$ is $Y_\ell \in [a_\ell, b_\ell]$ almost surely for some constants $a_\ell, b_\ell$ almost surely for all $\ell = 1, \cdots, t$. then 

\begin{equation*}
      \sum_{\ell=1}^t Y_\ell \leq  2\sqrt{\sum_{\ell=1}^t (b_\ell-a_\ell)^2 \ln\left(\frac{12t^2}{\widetilde{\delta}}\right)} 
\end{equation*}
with probability at least $1-\widetilde{\delta}$ for all $t \in \mathbb{N}$ simultaneously.
\end{lemma}

\begin{lemma}[Freedman Bounded RVs - unsimplified~\citep{freedman1975tail,azar2017minimax}]\label{lemma:simplified_freedman}
Suppose $\{ X_t \}_{t=1}^\infty$ is a martingale difference sequence with $| X_t | \leq b$. Let 
\begin{equation*}
    \mathrm{Var}_\ell(X_\ell) = \mathrm{Var}( X_\ell | X_1, \cdots, X_{\ell-1})
\end{equation*}
Let $V_t = \sum_{\ell=1}^t \mathrm{Var}_\ell(X_\ell)$ be the sum of conditional variances of $X_t$.  Then we have that for any $\widetilde{\delta} \in (0,1)$ and $t \in \mathbb{N}$
\begin{equation*}
    \mathbb{P}\left(  \sum_{\ell=1}^t X_\ell >    2\sqrt{V_t}A_t + 3b A_t^2 \right) \leq \widetilde{\delta}
\end{equation*}
Where $A_t = \sqrt{2 \ln \ln \left(2 \left(\max\left(\frac{V_t}{b^2} , 1\right)\right)\right) + \ln \frac{6}{\widetilde{\delta}}}$ and $h(s)$ corresponds to horizon index of the state partitions that contains state $s$.
\end{lemma}

\begin{lemma}[Empirical Bernstein Anytime, Theorem 4 of~\cite{maurer2009empirical}]\label{lemma::bernstein_anytime}
Let $\ell \geq 2$ and $\left\{Z_t\right\}_{t=1}^\ell$ be i.i.d. random variables with distribution $Z$ satisfying $|Z_t| \leq b$ for all $t \in [\ell]$ 
Let the sample variance be defined as,
\begin{equation*}
    \mathrm{Var}_\ell(Z) = \frac{1}{\ell(\ell-1)} \sum_{1 \leq i < j \leq \ell} (Z_i - Z_j)^2
\end{equation*}
With probability at least $1-\delta$,
\begin{equation*}
    \mathbb{E}[ Z] - \frac{1}{\ell} \sum_{i=1}^\ell Z_i \leq \sqrt{\frac{4  \mathrm{Var}_\ell(Z) \ln \frac{4\ell^2}{\delta} }{ \ell} } + \frac{7 b \ln \frac{4\ell^2}{\delta}}{3(\ell-1)}
\end{equation*}
for all $\ell \in [\mathbb{N}]$.
\end{lemma}

\begin{lemma}[Lemma 7.5 of \cite{agarwal2019reinforcement}] 
\label{lemma:inverse_squareroot_bounds}
Consider arbitrary $T$ sequence of trajectories $\tau_t = \{s_h^t,a_h^t\}_{h=1}^{H}$, for $t = 0,1,\dots, T$. We have
\begin{equation}
    \sum_{t=1}^{T}\sum_{h=1}^{H}\frac{\mathbf{1}((s_h^t, a_h^t) \in \mathcal{U})  }{\sqrt{N_h^t(s_h^t,a_h^t)}}\leq 2\sqrt{|\mathcal{U}|\sum_{(s,a) \in \mathcal{U}}  N_{h(s)}^T(s,a)}  \leq 2\sqrt{|\mathcal{U}|T}.
\end{equation}
Where $\mathcal{U} \subseteq \Scal \times \Acal$ is an arbitrary set of state action pairs and  $h(s)$ corresponds to horizon index of the state partitions that contains state $s$.
\end{lemma}
\begin{proof}
Consider swapping the order of summation
\begin{equation}
    \begin{split}
        &\sum_{t=1}^{T}\sum_{h=1}^{H}\frac{1}{\sqrt{N_h^t(s_h^t,a_h^t)}} = \sum_{h=1}^{H}\sum_{t=1}^{T}\frac{1}{\sqrt{N_h^t(s_h^t,a_h^t)}} = \sum_{(s,a)\in \mathcal{U}}\sum_{i=1}^{N_{h(s)}^T(s,a)}\frac{1}{\sqrt{i}}\\
        &\leq 2\sum_{(s,a)\in \mathcal{U}}\sqrt{N_{h(s)}^T(s,a)}\leq 2\sqrt{|\mathcal{U}| \sum_{(s,a) \in \mathcal{U}} N_{h(s)}^T(s,a)} \leq 2\sqrt{|\mathcal{U}|T},
    \end{split}
\end{equation}

where the first inequality use the fact that $\sum_{i=1}^N1/\sqrt{i}\leq 2\sqrt{N}$ and the second inequality holds due to Cauchy-Schwarz inequality. 
\end{proof}

\begin{lemma}[Sum of inverse counts]\label{lemma:inverse_counts_bounds}
Consider arbitrary $T$ sequence of trajectories $\tau_t = \{s_h^t,a_h^t\}_{h=1}^{H}$, for $t = 0,1,\dots, T$. We have
\begin{equation}
    \sum_{t=1}^{T}\sum_{h=1}^{H}\frac{\mathbf{1}((s_h^t, a_h^t) \in \mathcal{U})  }{N_h^t(s_h^t,a_h^t)}\leq 2|\mathcal{U}| \log(T+1).
\end{equation}
Where $\mathcal{U} \subseteq \Scal \times \Acal$ is an arbitrary set of state action pairs. 
\end{lemma}
\begin{proof}
Consider swapping the order of summation
\begin{equation}
    \begin{split}
        &\sum_{t=1}^{T}\sum_{h=1}^{H}\frac{1}{{N_h^t(s_h^t,a_h^t)}} = \sum_{h=1}^{H}\sum_{t=1}^{T}\frac{1}{{N_h^t(s_h^t,a_h^t)}} = \sum_{(s,a)\in \mathcal{U}}\sum_{i=1}^{N_{h(s)}^T(s,a)}\frac{1}{{i}}\\
        &\leq 2\sum_{(s,a)\in \mathcal{U}}\log\left(N_{h(s)}^T(s,a)+1\right)\leq 2|\mathcal{U}|  \log(T+1)
    \end{split}
\end{equation}

\end{proof}

\section{Supporting Algorithms}
The UCBVI algorithm~\cite{azar2017minimax,agarwal2019reinforcement} is described below.
\begin{small}
\begin{algorithm}[!h]
  \caption{UCBVI}\label{algo:UCBVI}
  \begin{algorithmic}[1]
    \State \textbf{Input} reward function $r$ (assumed to be known), confidence parameters 
      \State \textbf{for} {$t=1,...,T$}
        \State \qquad Compute $\wh{P}_t$ using all previous empirical transition data as $\wh{\bb P}_t(s^\prime|s,a) := \frac{N_h^t(s,a,s^\prime)}{N_h^t(s,a)},\;\forall h,s,a,s^\prime$.
        \State \qquad Compute reward bonus $b_h^t(s,a)=2H\sqrt{\ln(SAHK/\delta)/N_h^t(s,a)}$
        \State \qquad Run Value-Iteration on $\{\wh{\bb P}_t,r+b_h^t\}_{h=0}^{H-1}$.  
        \State \qquad Set $\pi_t$ as the returned policy of VI.
      \State \textbf{End for }
  \end{algorithmic}
\end{algorithm}
\end{small}

UCBVI works by adding a count based bonus to the rewards and then running value iteration over the empirical model based on these bonus augmented rewards. This encourages optimism of the empirical value functions which ensures an appropriate trade-off between exploration and exploitation. In contrast with previous approaches with provable guarantees for tabular RL problems such as UCRL~\cite{auer2008near}.

As described in~\cite{azar2017minimax}, the sample complexity of UCBVI is (up to low order terms) $\widetilde\Ocal\left( H\sqrt{|\Scal||\Acal| }\right)$ with variance aware bonuses and $\widetilde\Ocal\left( H^{3/2} \sqrt{|\Scal||\Acal|T }\right)$ when the bonus terms are computed using simple count based scores as above. Our results are based on a modification of the rates derived in~\cite{agarwal2019reinforcement} that are of order $\widetilde \Ocal( H^{2} |\Scal| \sqrt{|\Acal|T}  ) $. Despite this base rate having a suboptimal dependence in $H$ and $|\Scal|$ in contrast with the minimax rate of~\cite{azar2017minimax} our results still represent an big improvement w.r.t. the un-shaped rates of~\cite{azar2017minimax} when the effective state space is small.  

\section{Experimental Details}
\label{sec:exp_details}
We conducted experiments building on the UCBVI framework outlined in ~\cite{azar2017minimax}. Specifically, we used the bonus 1 formulation from the UCBVI Algorithm~\ref{algo:UCBVI} in ~\cite{azar2017minimax}, with the bonus specifically being $\frac{C}{\sqrt{N(s,a)}}$ as the base UCBVI implementation. Our implementation is based on the open source implementation by Ian Osband ~\url{https://github.com/iosband/TabulaRL}, and open source code is attached. 

The implementations of UCBVI-BS replaces this bonus by $\frac{C.\Vtilde}{\sqrt{N(s,a)}}$.  In our implementation we chose the scaling to be $0.1$ with a hyperparameter sweep. The $\Vtilde$ was chosen by scaling the optimal value function $\Vstar$ by factors chosen between $1 + \beta$ and $1 - \beta$,  with $beta$ as $0.2, 0.5, 0.9$ (as described in the experimental section). 

The environments themselves are open gridworlds of size $8$, a corridor of size $10x10$ and the double corridor has size $10x20$. The reward is $0$ everywhere except the goal location.  Every experiment is averaged over 3 random seeds.

\section{Intuition for $\pseudosub_\Delta$ and $\pathpseudosub_\Delta$}
\begin{figure}
    \centering
    \includegraphics[width=0.6\textwidth]{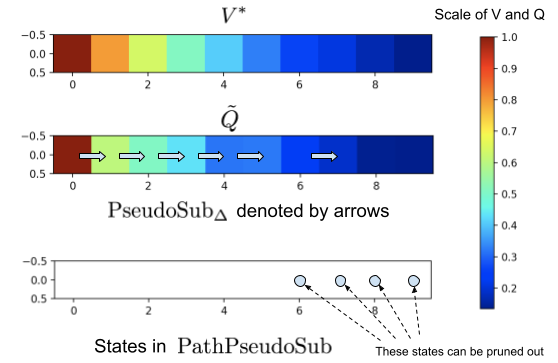}
    \caption{Illustration of the concepts of $\pseudosub_\Delta$ and $\pathpseudosub_\Delta$. Top row shows the optimal value function for this chain MDP (colorbar shows scale). Middle row shows the $\tilde{Q}$ obtained from the shaping and the $(s, a)$ pairs which are in $\pseudosub_\Delta$ (denoted by arrows). From $\pseudosub_\Delta$, the bottom row shows states in $\pathpseudosub_\Delta$ which can be pruned out via shaping.}
    \label{fig:pseudosub-illustration}
\end{figure}

Here we provide some intuition for the concepts of $\pseudosub_\Delta$ and $\pathpseudosub_\Delta$ in a deterministic chain environment (actions being left and right with deterministic transitions). The environment has a reward of $1$ at state $0$ and $0$ elsewhere. Discount factor is $0.8$ here, and the $\Vtilde$ has a sandwich factor $\beta=0.3$. The true optimal value function is shown at the top of Fig.~\ref{fig:pseudosub-illustration}. The reward shaping is shown in the middle row of Fig.~\ref{fig:pseudosub-illustration}. The concept of pseudosub is shown via the arrows at $(s, a)$ pairs that lie in $\pseudosub_\Delta$ for this particular choice of $\Vtilde$ and $\Delta$. They basically denote states which are suboptimal with $\Delta$ confidence according to the shaping function $\Vtilde$. Now given this definition of $\pseudosub_\Delta$, the corresponding $\pathpseudosub_\Delta$ is shown in the last row. These are states that can be eliminated via reward shaping. As we can see, around half of the states can be eliminated this way, making exploration much more directed. 

\section{Results for Corrupted $\Vtilde$}
\begin{figure}
    \centering
    \includegraphics[width=0.6\textwidth]{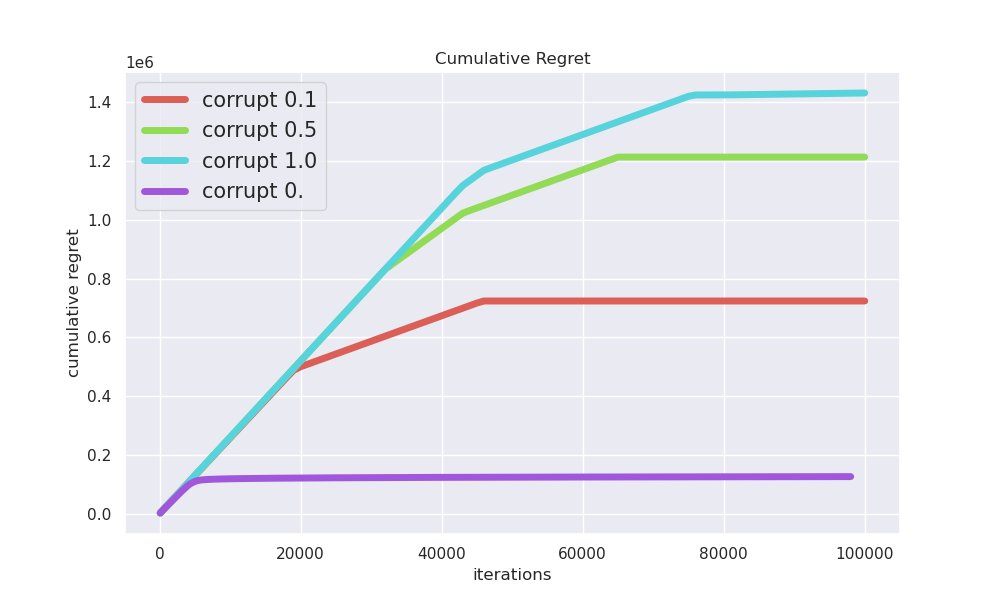}
    \caption{Illustration of behavior with corrupted shaping functions $\Vtilde$ with varying levels of corruption}
    \label{fig:corrupt}
\end{figure}

We evaluate the behavior of UCBVI-Shaped as we use a corrupted version of $\Vtilde$ for the shaping in Fig.~\ref{fig:corrupt}. In particular, we do this by constructing the sandwiched value function $\Vtilde$ as described in our experimental evaluation (in this case with $\beta=1.5$ in the corridor environment) and then adding gaussian noise to corrupt the value function. We experiment with corruptions of $0.$, $0.1$, $0.5$, $1.0$ variance, zero-centered gaussian noise. The resulting regret is seen in Fig.~\ref{fig:corrupt}. As expected, the results show that corruption of $\Vtilde$ hurts the results as more and more corruption happens, but still performs better than without any shaping at all. 

\section{Results for RND}

\begin{figure}
    \centering
    \includegraphics[width=0.5\textwidth]{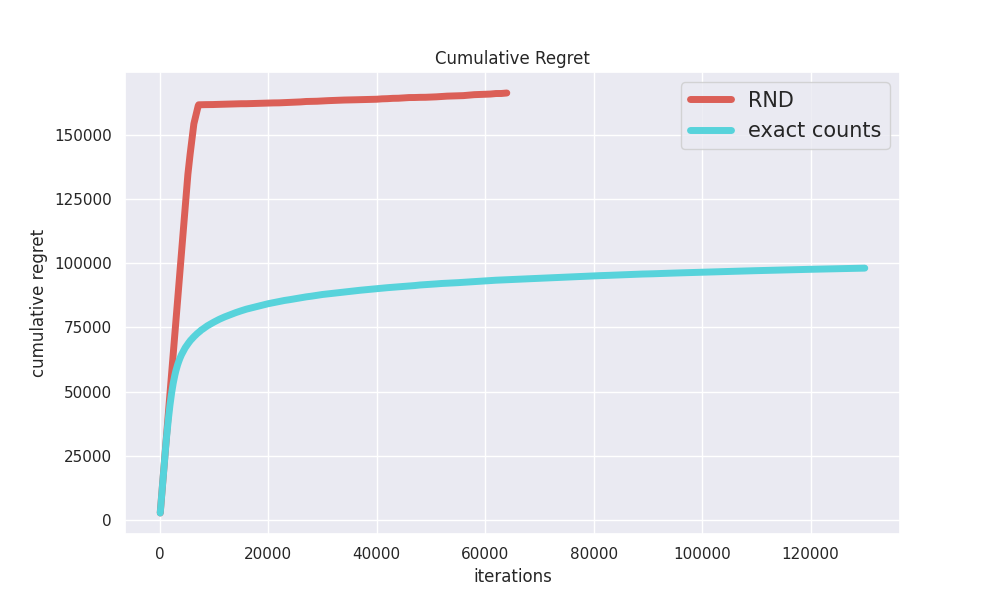}
    \caption{Illustration of behavior of UCBVI-Shaped-BS with RND based pseudocounts instead of exact counts.}
    \label{fig:rnd}
\end{figure}

We also ran numerical simulations (in Fig.~\ref{fig:rnd}) with a neural network based ``pseudo-count" method - random network distillation (RND) ~\cite{burda2018exploration}. This trains a neural network against a random pre-initialized target and uses projection error as a ``pseudocount". We ran UCBVI-Shaped-BS with the inverse counts typical in UCBVI-shaped replaced with model error. The results in Fig.~\ref{fig:rnd} are considerably slower than with exact counts likely due to noise in counts estimation, although the results are likely to be improved with environment specific tuning. 

\section{Discussion of Applicability of Reward Shaping Assumption}

There can’t be any way to obtain an improvement to the UCBVI regret bounds by using a reward shaping mechanism unless there is some information about the optimal policy / optimal value function encoded in the shaping function. In other words, there cannot be any free lunch in this setting. Prior knowledge is fundamental for reward shaping to be successful. 

In this work we have chosen to make the assumption that the side information available to us $\Vtilde$ can be used to sandwich the value of $\Vstar$. There are many examples where this assumption may hold. 

\begin{enumerate}

\item Maze like environments. In any goal metrized goal oriented environment (for example a maze) with a stay-in-place action, if the position of the goal state is (roughly) known and the reward function equals -1 at every step from a state to a neighboring state and 0 for the stay-in-place action at the goal state, using the metric distance from the current state to the goal state may be a good estimator of the number of steps needed to reach the goal ($-V^\star$). The accuracy of this approximation will depend on how off this heuristic is. 

\item Sim to Real. We may think of training a robot in a simulation environment and use the learned optimal value function in the simulation as our input $\Vtilde$. Having this information at hand should make training in real much easier than from scratch depending on how accurately the $\Vtilde$ estimator from the simulation tracks $V^*$. Under the assumption that dynamics are lipschitz bounded, and there is bounded state estimation error, dynamics estimation error, the resulting value function also shares the sandwich property of the true value function. 

\item Object manipulation. In this setting even if the objective is to manipulate a huge combination of objects, an appropriate combination of euclidean distances of objects can serve as a good proxy for $\Vstar$. While this may underestimate the distance to the goal especially in the presence of obstacles, in most cases a sandwich term $\beta$ can be found that bounds the true value function (under the assumption that the number of obstacles is bounded). This becomes particularly pronounced as the number of objects increases and there is a combinatorially large state space, all of which cannot possibly be explored. 

\item Motion planning: Recent methods have attempted to combine motion planning and RL ~\cite{faust18prm}. One such instantation would involve using an anytime motion planning to generate a reward function to guide RL. An anytime motion planner may start by generating suboptimal paths, and gets better over time. If the motion planning is run only for some time so as to get a beta-sandwiched suboptimal path (and hence reward shaping as a sandwich on $\Vstar$) to guide RL, this would then provide suboptimal shaping that can then be improved with subsequent reinforcement learning as described in our work. 
\item First person RPG: For games like first person RPGs, reward shaping terms may look like a sum of euclidean distances to different objects of interest. This will underestimate the true value function will sandwich the optimal value function depending on how many obstacles are in the environment. This environment will also likely benefit from pruning off large parts of the state space. 
    
\end{enumerate}

In all of these situations, it is possible to construct a multiplicative approximation to $\Vstar$ to be used as $\Vtilde$. 

\section{Limitations}

A limitation of our work is that the reward shaping has to be provided before hand rather than learned or inferred, which would be very interesting to explore in future work. Additionally, the shaping bound we have is a point-wise one rather than a correlation based bound in expectation, which can be susceptible to be outlier rewards which have a large sandwich term $\beta$. This should be investigated in future work in more detail. Additionally, it is likely that reward shaping can directly allow for horizon reduction in value computation, which should be considered in future work more directly. The behavior of these types of methods under function approximation also may be quite different, which should be studied more systematically.

\subsection{Does decaying suboptimal reward shaping help over standard shaped rewards?}
\vspace{-0.5em}
\begin{figure}[!ht]
    \vspace{-1em}
    \centering
    \begin{subfigure}[b]{0.3\textwidth}
        \includegraphics[width=\linewidth]{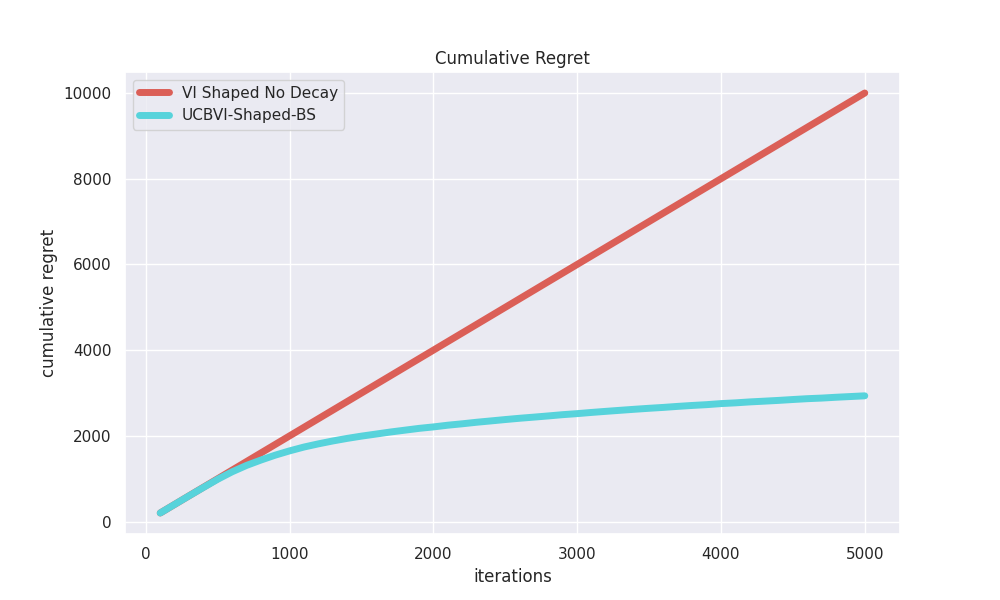}
        \caption{Quantitative comparison of effect of bonus decay}
    \end{subfigure}
    \begin{subfigure}[b]{0.3\textwidth}
        \includegraphics[width=0.8\linewidth]{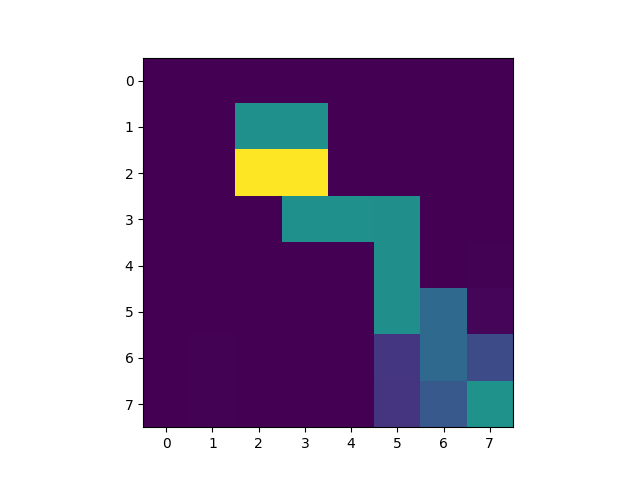}
        \caption{Visualization of non decaying agent getting stuck}
    \end{subfigure}
    \begin{subfigure}[b]{0.3\textwidth}
        \includegraphics[width=0.8\linewidth]{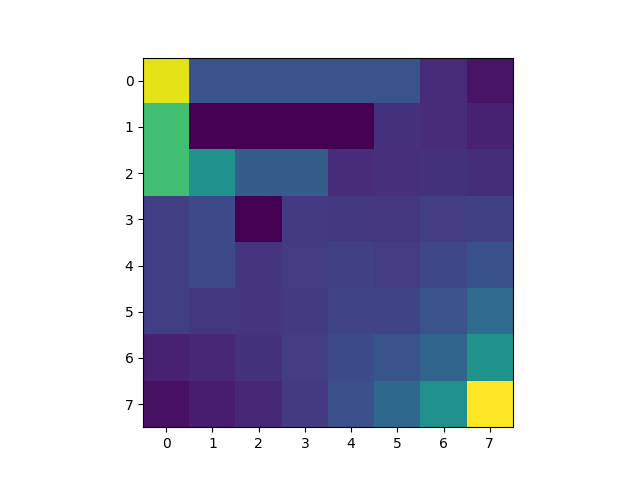}
        \caption{Visualization of decaying agent (UCBVI-Shaped) reaching goal}
    \end{subfigure}
    \caption{\footnotesize{Effect of bonus decay on the performance of UCBVI-Shaped vs standard reward shaping under shaping misspecification. As we can see, without decay the agent can stuck in arbitrarily sub-optimal points, whereas with decay the agent easily converges to an optimal solution.}}
    \label{fig:decay}
    \vspace{-1em}
\end{figure}
We ran simulations to understand whether the adaptive decay of $\Vtilde$ by $\frac{1}{\sqrt{N(s, a)}}$ actually provides tangible benefits, specifically when the shaping is suboptimal. We compared UCBVI-Shaped with a variant where $\Vtilde$ is added to the reward linearly, as done in many practical RL implementations. As seen in Fig.~\ref{fig:decay}, with suboptimal shaping, UCBVI-shaped dampens the effect of the suboptimal shaping and succeed, whereas simple addition of shaping leads to the agent becoming trapped. This suggests that not only can reward shaping improve sample efficiency, but incorporating reward shaping via UCBVI-Shaped can mitigate the potential bias from suboptimal shaping. \vspace{-0.5em}


\end{document}